\documentclass[10pt,twocolumn,letterpaper]{article}

\usepackage[pagenumbers]{cvpr} 

\usepackage[accsupp]{axessibility}
\usepackage{arydshln}
\usepackage{multirow}
\usepackage{multicol}
\usepackage{svg}
\usepackage{graphicx}
\usepackage{overpic}
\usepackage{xcolor}
\usepackage{tikz}

\usepackage{silence}
\makeatletter
\robustify\@latex@warning@no@line
\makeatother
\usepackage{authblk}
\newcommand{\email}[1]{\texttt{\small{#1@ucsb.edu}}}

\usepackage{color}

\definecolor{darkgreen}{RGB}{30,150,30}
\definecolor{darkblue}{RGB}{0,0,127}
\definecolor{darkyellow}{RGB}{171,133,0}
\definecolor{darkred}{RGB}{180,20,20}
\definecolor{darkmagenta}{RGB}{127,0,127}
\definecolor{darkcyan}{RGB}{0,127,127}

\newcommand{\ourwork}{\text{Instruct-CLIP}}
\newcommand{\ourworkabbr}{\text{I-CLIP}}
\newcommand{\myvis}{\text{vis}}
\newcommand{\mytxt}{\text{txt}}
\newcommand{\iclipvis}{\ourworkabbr{}_\myvis{}}
\newcommand{\icliptxt}{\ourworkabbr{}_\mytxt{}}
\newcommand{\dinofunc}{\ensuremath{\text{DINO}_\text{v2}}}
\newcommand{\lddinofunc}{\ensuremath{\text{LD-DINO}_\text{v2}}}
\newcommand{\dinotext}{\ensuremath{\text{DINOv2}}}
\newcommand{\lddinotext}{\ensuremath{\text{LD-DINOv2}}}

\definecolor{cvprblue}{rgb}{0.21,0.49,0.74}
\usepackage[pagebackref,breaklinks,colorlinks,allcolors=cvprblue]{hyperref}

\def\paperID{3380} 
\def\confName{CVPR}
\def\confYear{2025}

\title{\ourwork{}: Improving Instruction-Guided Image Editing\\with Automated Data Refinement Using Contrastive Learning}

\author[]{Sherry X. Chen\hspace{15mm}Misha Sra\hspace{15mm}Pradeep Sen}

\affil[]{University of California, Santa Barbara\\\email{\{xchen774, sra, psen\}}}

\begin{document}
\maketitle
\begin{abstract}
Although natural language instructions offer an intuitive way to guide automated image editing, deep-learning models often struggle to achieve high-quality results, largely due to the difficulty of creating large, high-quality training datasets. To do this, previous approaches have typically relied on text-to-image (T2I) generative models to produce pairs of original and edited images that simulate the input/output of an instruction-guided image-editing model. However, these image pairs often fail to align with the specified edit instructions due to the limitations of T2I models, which negatively impacts models trained on such datasets. To address this, we present \ourwork{} (\ourworkabbr{}), a self-supervised method that learns the semantic changes between original and edited images to refine and better align the instructions in existing datasets. Furthermore, we adapt \ourwork{} to handle noisy latent images and diffusion timesteps so that it can be used to train latent diffusion models (LDMs) and efficiently enforce alignment between the edit instruction and the image changes in latent space at any step of the diffusion pipeline. We use \ourwork{} to correct the InstructPix2Pix dataset and get over 120K refined samples we then use to fine-tune their model, guided by our novel \ourworkabbr{}-based loss function. The resulting model can produce edits that are more aligned with the given instructions. Our code and dataset are available at \href{https://github.com/SherryXTChen/Instruct-CLIP.git}{https://github.com/SherryXTChen/Instruct-CLIP.git}.
\end{abstract}
    
\section{Introduction}

\begin{figure}[ht]
\centering

\resizebox{0.95\linewidth}{!}{ 
\begin{minipage}{\linewidth}

\begin{small}
\begin{tabbing}
    \hspace{2.7em}\= \hspace{9em} \= \hspace{6.8em} \= \kill
    \> Original \> IP2P \> \ourworkabbr{} (Ours)  \\
\end{tabbing}    
\end{small}
\vspace{-0.3in}
    \begin{subfigure}{\linewidth}
        \includegraphics[width=\linewidth, trim=0 85px 0 0, clip]{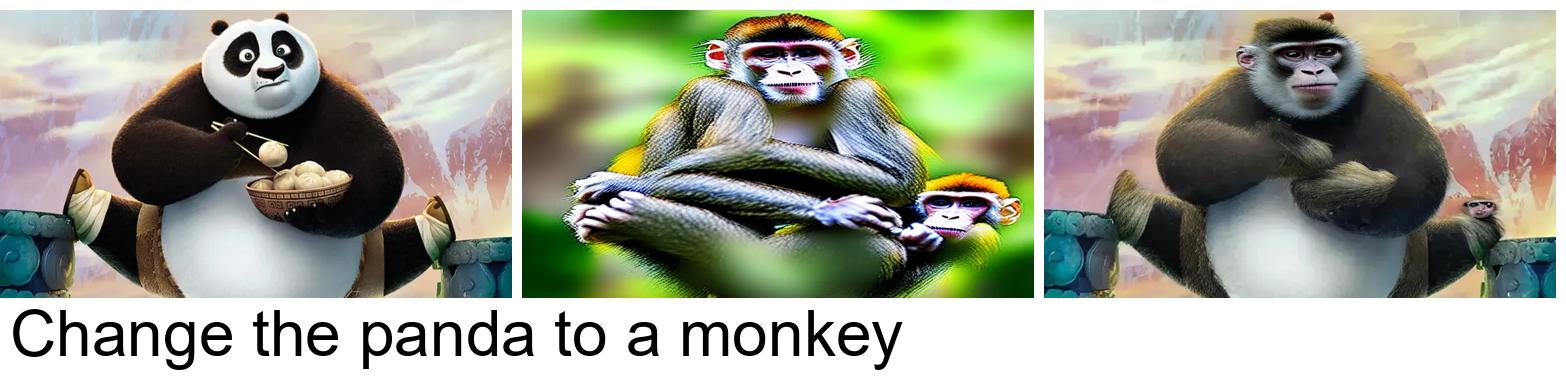}
        \parbox{\linewidth}{\vspace{-0.03in}
        ``Change the panda to a monkey''
        \vspace{0.02in}}
    \end{subfigure}

    \begin{subfigure}{\linewidth}
        \includegraphics[width=\linewidth, trim=0 135px 0 0, clip]{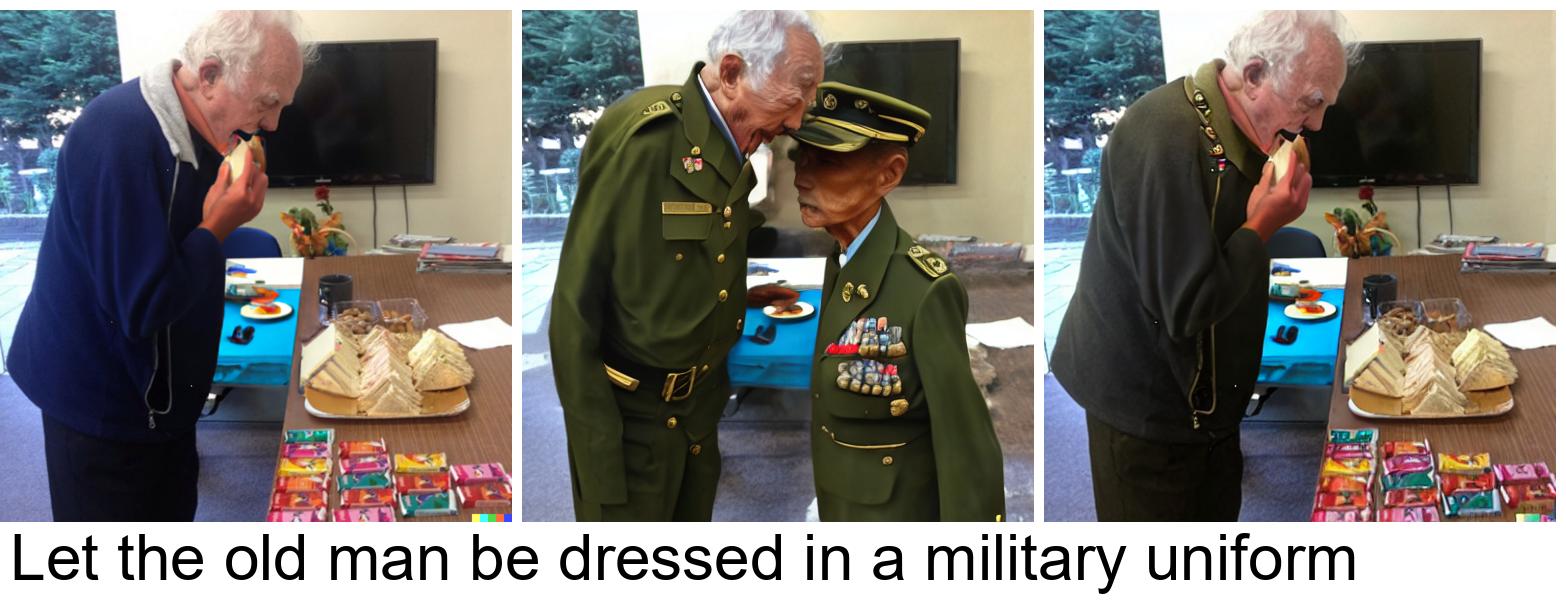}
        \parbox{\linewidth}{\vspace{-0.03in}
        ``Let the old man be dressed in a military uniform''
        \vspace{0.02in}}
    \end{subfigure}

    \begin{subfigure}{\linewidth}
        \includegraphics[width=\linewidth, trim=0 85px 0 0, clip]{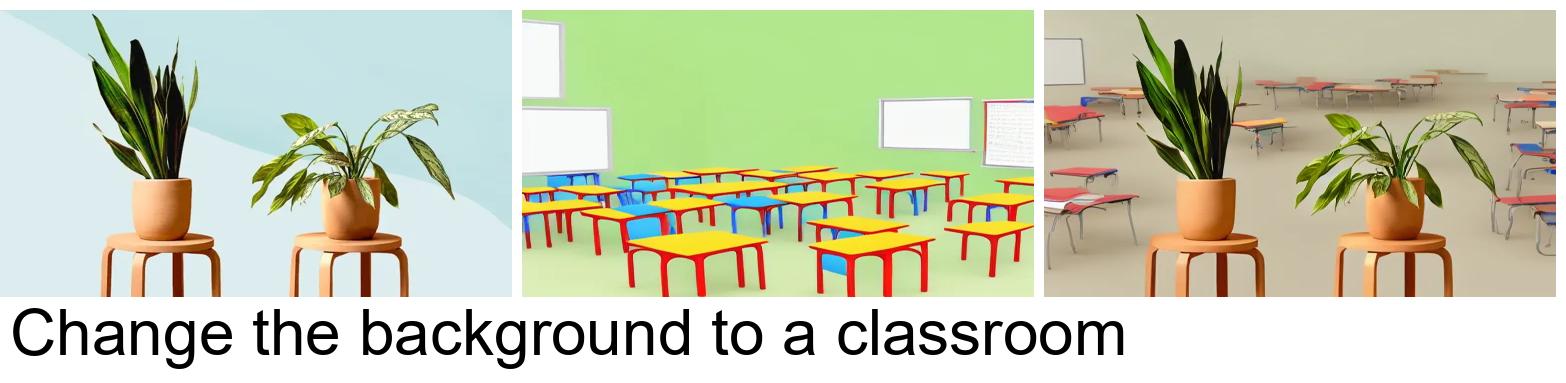}
        \parbox{\linewidth}{\vspace{-0.03in}
        ``Change the background to a classroom''
        \vspace{0.02in}}
    \end{subfigure}

     \begin{subfigure}{\linewidth}
        \includegraphics[width=\linewidth, trim=0 90px 0 0, clip]{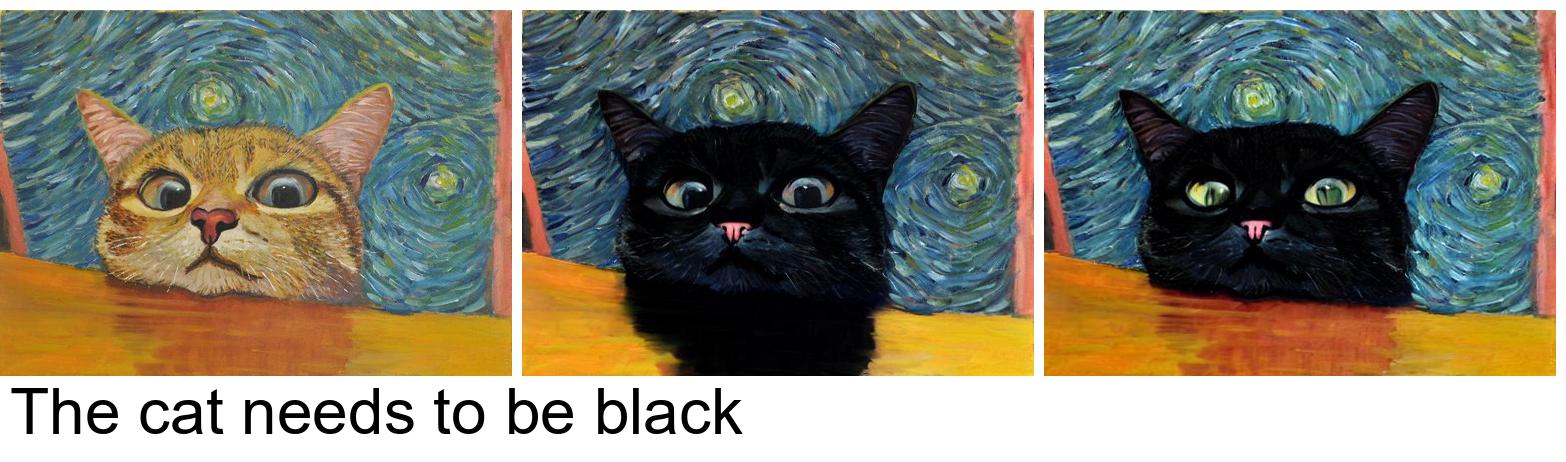}
        \parbox{\linewidth}{
        ``The cat needs to be black''
        \vspace{0.02in}}
    \end{subfigure}

     \begin{subfigure}{\linewidth}
        \includegraphics[width=\linewidth, trim=0 85px 0 0, clip]{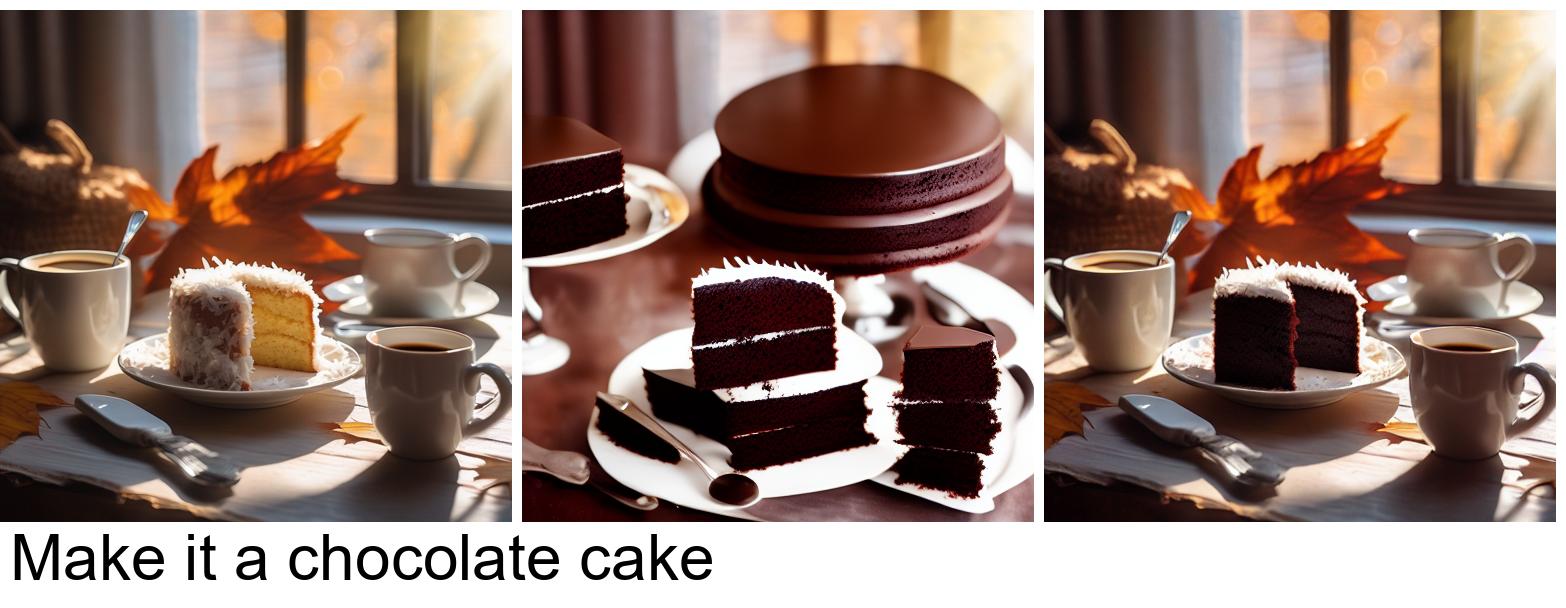}
        \parbox{\linewidth}{\vspace{-0.03in}
        ``Make it a chocolate cake''
        \vspace{0.02in}}
    \end{subfigure}
    
\end{minipage}
} 
\vspace{-0.07in}
\caption{Results showcasing the strength of \ourwork{}  (\ourworkabbr{}) compared to state-of-the-art InstructPix2Pix (IP2P)~\cite{brooks2023instructpix2pix}.\vspace{-0.15in}}
\label{fig:teaser}
\end{figure}

Recent advances in \emph{instruction-guided image editing} have introduced intuitive and powerful tools that require only a single text instruction for image editing~\cite{brooks2023instructpix2pix,zhao2024ultraedit,zhang2023hive,zhang2024magicbrush,yildirim2023inst}. In particular, InstructPix2Pix (IP2P)~\cite{brooks2023instructpix2pix} pioneered diffusion-based instruction-driven image editing, serving as the foundation for numerous subsequent works~\cite{zhang2024magicbrush,mirzaei2025watch,li2024zone}. These methods leverage pre-trained text-to-image (T2I) models~\cite{song2020denoising,dhariwal2021diffusion,rombach2022high,podell2023sdxl,luo2023latent} and condition the generation process on edit instructions instead of the usual prompts. However, these models still need to be fine-tuned on appropriate datasets to learn how to make instruction-guided edits.

\begin{figure}
    \begin{tiny}
    \centering
        \begin{subfigure}[t]{0.475\linewidth}
        \begin{small}
        \begin{tabbing}
            \hspace{1.6em}\= \hspace{6.3em} \= \kill
            \> Original \> Edited  \\
        \end{tabbing}    
        \end{small}
        \vspace{-0.3in}

        \includegraphics[width=\linewidth, trim=0 100px 0 0, clip]{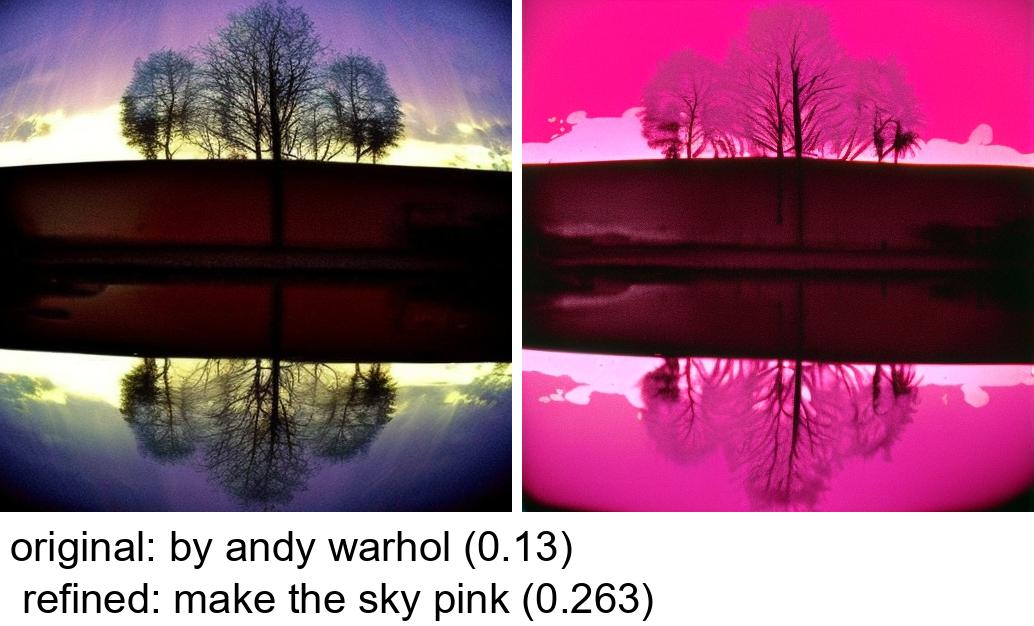}
        \parbox{\linewidth}{
        {\sc Original:} ``by andy warhol'' (0.130)\\
        {\sc Refined:} ``make the sky pink'' (0.263)\\
        \vspace{0.02in}}
    \end{subfigure}
    \hfill 
    \begin{subfigure}[t]{0.475\linewidth}
            \begin{small}
        \begin{tabbing}
            \hspace{1.6em}\= \hspace{6.3em} \= \kill
            \> Original \> Edited  \\
        \end{tabbing}    
        \end{small}
        \vspace{-0.3in}
        \includegraphics[width=\linewidth, trim=0 100px 0 0, clip]{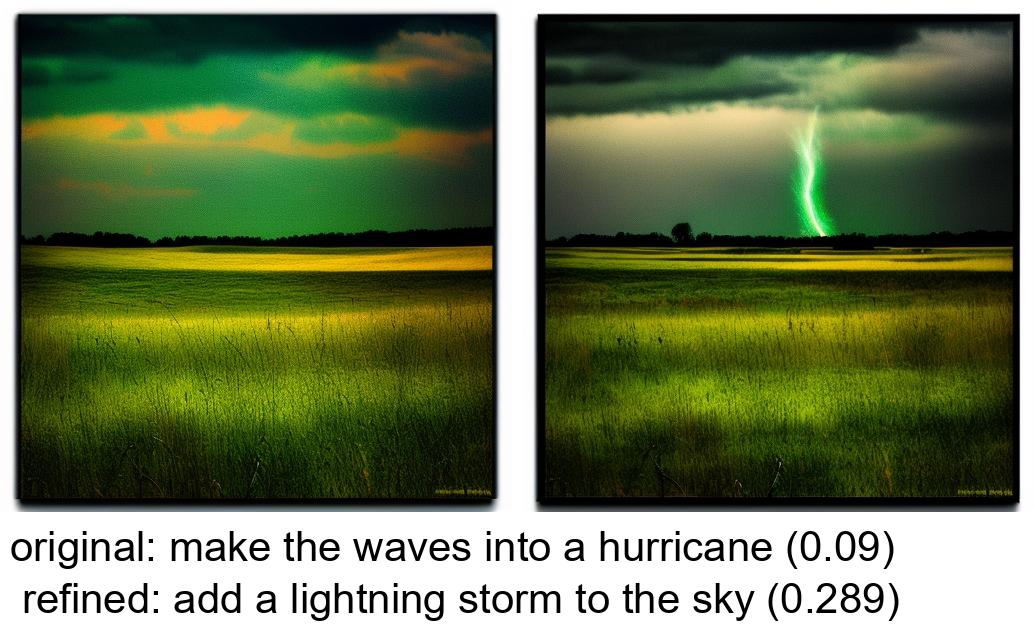}
        \parbox{\linewidth}{
        {\sc Original:} ``make the waves into a hurricane'' (0.090)\\
        {\sc Refined:} ``add a lightning storm to the sky'' (0.289)\\
        \vspace{0.02in}}
    \end{subfigure}

    \begin{subfigure}[t]{0.475\linewidth}
        \includegraphics[width=\linewidth, trim=0 100px 0 0, clip]{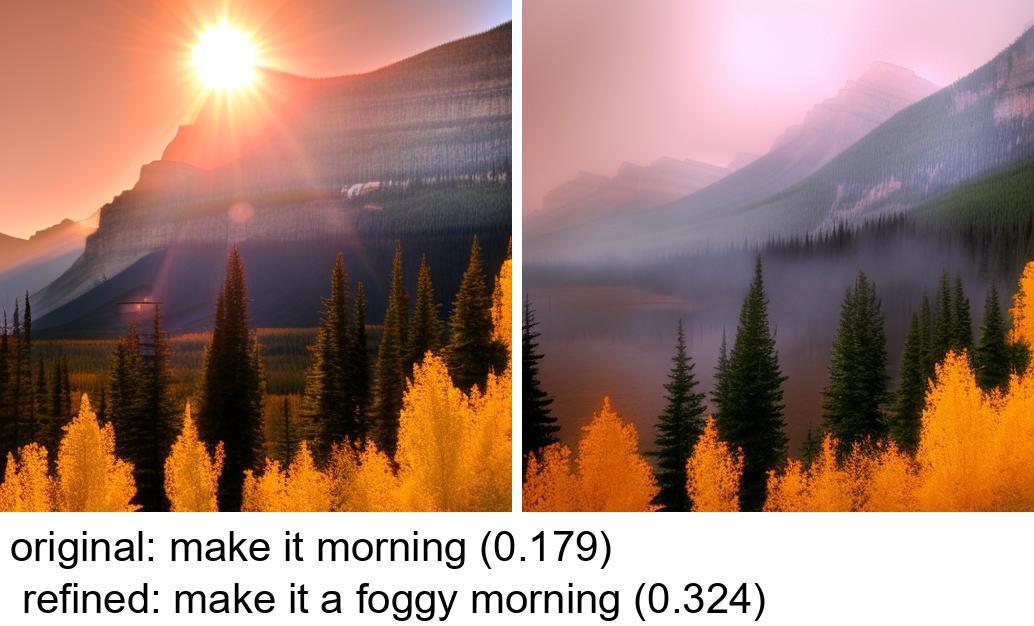}
        \parbox{\linewidth}{
        {\sc Original:} ``make it morning'' (0.179)\\
        {\sc Refined:} ``make it a foggy morning'' (0.324)\\
        \vspace{0.02in}}
    \end{subfigure}
    \hfill 
    \begin{subfigure}[t]{0.475\linewidth}
        \includegraphics[width=\linewidth, trim=0 100px 0 0, clip]{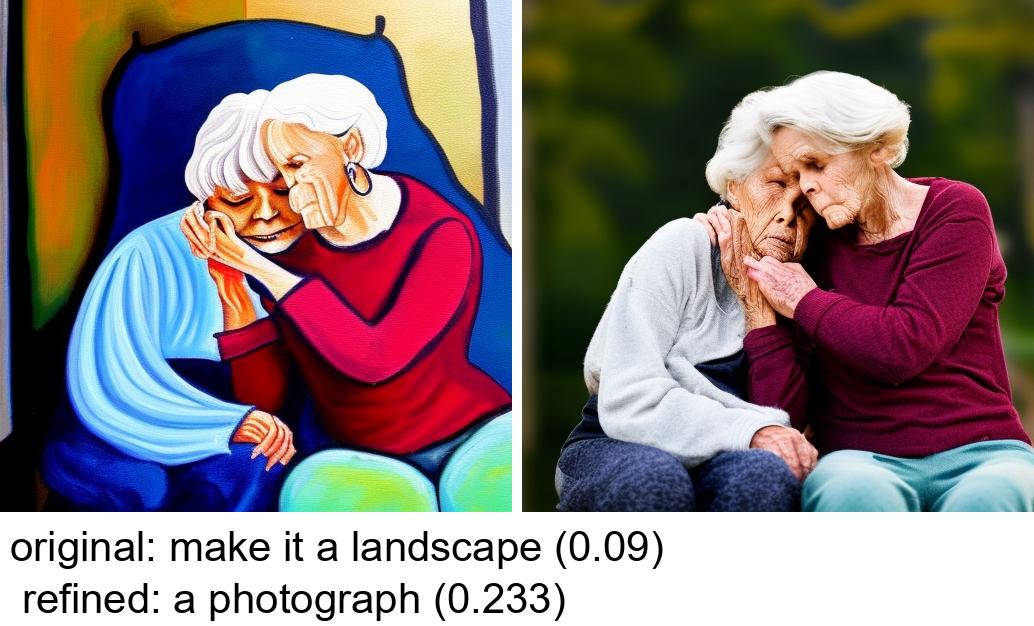}
        \parbox{\linewidth}{
        {\sc Original:} ``make it a landscape'' (0.090)\\
        {\sc Refined:} ``a photograph'' (0.233)\\
        \vspace{0.02in}}
    \end{subfigure}

    \begin{subfigure}[t]{0.475\linewidth}
        \includegraphics[width=\linewidth, trim=0 100px 0 0, clip]{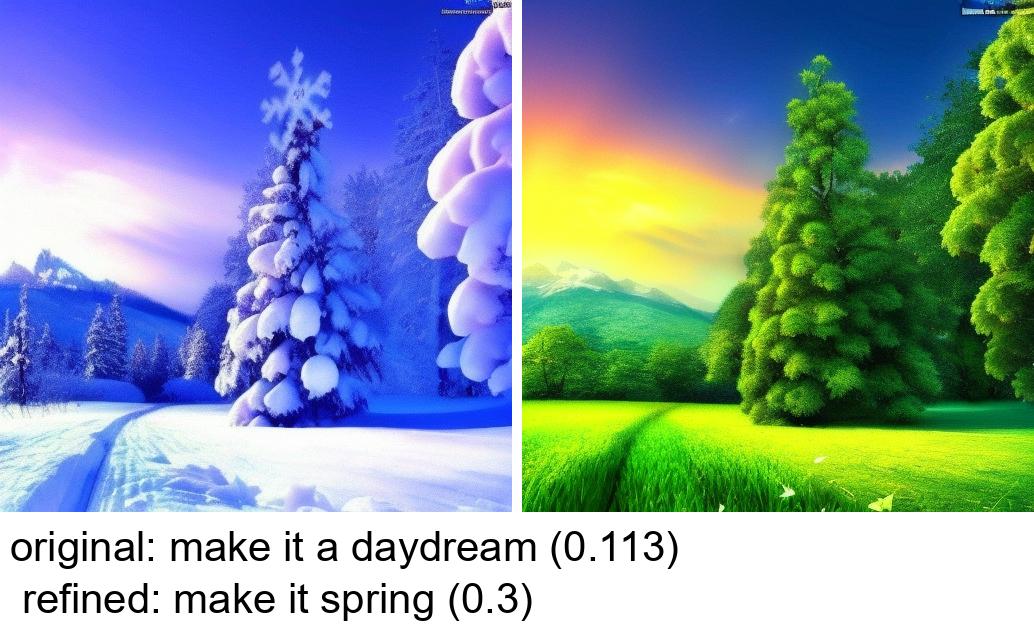}
        \parbox{\linewidth}{
        {\sc Original:} ``make it a daydream'' (0.113)\\
        {\sc Refined:} ``make it spring'' (0.300)\\
        \vspace{0.02in}}
    \end{subfigure}
    \hfill 
    \begin{subfigure}[t]{0.475\linewidth}
        \includegraphics[width=\linewidth, trim=0 100px 0 0, clip]{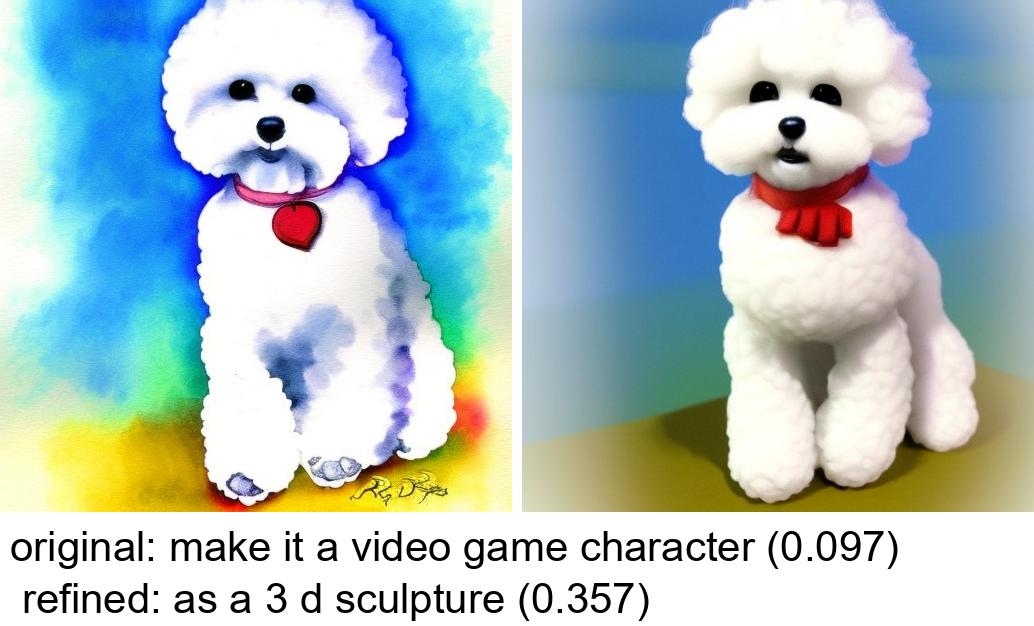}
        \parbox{\linewidth}{
        \mbox{{\sc Original:} ``make it a video game character'' (0.097)} \\ 
        {\sc Refined:} ``as a 3 d sculpture'' (0.357)\\
        \vspace{0.02in}}
    \end{subfigure}
    \end{tiny}
    \vspace{-0.1in}
    \caption{Problems with existing instruction-guided image-editing datasets~\cite{brooks2023instructpix2pix}. As shown, there are many examples where the dataset's original edit instruction does not match the actual changes in the images. Our \ourworkabbr{} approach refines edit instructions to match the visual change better and allows us to train a system that produces better outputs. The values in parentheses are the cosine similarity between the visual change from the original to the edited image and the edit instruction from \ourworkabbr{}. 
    }
    \label{fig:dataset_samples}

\end{figure}

Some have created these datasets by using T2I models to approximate the desired behavior of an instruction-guided image editing system. For example, several approaches~\cite{brooks2023instructpix2pix,zhao2024ultraedit,zhang2023hive} leverage Prompt-to-Prompt~\cite{hertz2022prompt} to generate a pair of images that look like one is an edited version of the other by using a pair of appropriately modified prompts. A separate, fine-tuned large-language model (LLM), such as GPT3~\cite{brown2020language}, is then used to generate plausible edit instructions from these prompts.

Given the generality of Prompt-to-Prompt and LLMs, such datasets can cover a wide range of instructions. However, the quality of image editing is often limited by the capabilities of the generative methods, resulting in problems when the changes between the original and the edited image are misaligned with the edit instruction (Fig.~\ref{fig:dataset_samples}). This affects the performance on models trained on these datasets, as we can see with the IP2P results in Fig.~\ref{fig:teaser}.

Although improving image-editing datasets would significantly enhance the performance of models trained on them, doing so is difficult. On one hand, manually creating edit instructions and corresponding outputs is extremely labor-intensive and impractical. On the other, it is non-trivial to construct a pipeline that can perform effective edits in order to create an instruction-guided, image-editing dataset. After all, if such a pipeline existed, it would be a successful instruction-guided, image-editing method itself.

Furthermore, creating a brand new dataset from scratch may not be necessary. Despite the inherent limitations of previous datasets, we observe that they still provide some ``signal'' for training image editing models. After all, as shown in Fig.~\ref{fig:teaser}, we see that models trained on them produce results that, while imperfect, show that the system is understanding {\em something} about the desired edit. Can we harness this signal to refine the dataset and address its issues?

Inspired by CLIP~\cite{radford2021learning}, which learns rich semantic alignments through contrastive language-image pre-training, we embed original-edited image pairs and their corresponding edit instructions into the same feature space in a similar manner with a neural  network, which we call  {\em \ourwork{}}. We then follow this with a separate text decoder to generate more precise instructions from the learned features~\cite{li2023decap}. 

\ourwork{} is equipped with a modified DINOv2~\cite{oquab2023dinov2} backbone capable of handling noisy latent images in Stable Diffusion (SD)~\cite{rombach2022high} so it can not only enhance training data quality in the pre-processing stage but also serve as an efficient learning objective during training. We use \ourwork{} to refine instructions in the IP2P dataset~\cite{brooks2023instructpix2pix}, resulting in 120K+ unique, enhanced samples (see Fig.~\ref{fig:dataset_samples}). The IP2P model is then fine-tuned on this dataset with an \ourwork{}-guided loss, enabling the trained instruction-guided editing model to produce results more faithfully aligned with the instructions (see Fig.~\ref{fig:qualitative}).

In summary, our contributions are:
\begin{itemize}
    \item A dual-purpose model that refines dataset instructions and enhances editing model training.
    \item A large-scale dataset with over 120K samples featuring more accurate and enriched instructions.
    \item An instruction-guided image editing method trained on above dataset.
\end{itemize}
\section{Related Work\label{sec:related_work}}

\subsection{Text-to-image diffusion-based image editing}

Diffusion models~\cite{song2020denoising,dhariwal2021diffusion,rombach2022high,podell2023sdxl,luo2023latent,esser2403scaling} have set a new standard for image generation quality, and form the basis of various text-to-image (T2I) editing applications. A range of methods~\cite{hertz2022prompt, mokady2023null, wu2023uncovering, wallace2023edict, parmar2023zero, tumanyan2023plug} aim to edit a given image to semantically align with a target prompt, using the input prompt as an anchor. However, traditional T2I models like Stable Diffusion (SD)~\cite{rombach2022high} often yield inconsistent results even with similar prompts, where both subject and context can change significantly. Prompt-to-Prompt~\cite{hertz2022prompt} addresses this by preserving attention maps for the shared components in the input and the target prompt to improve image visual alignment, but both images are synthesized so it cannot handle arbitrary input images.
Null-text inversion~\cite{mokady2023null} overcomes this by reconstructing an input image in SD based on the input prompt. They do this by optimizing the empty-text as the negative prompt, enabling controlled edits when combined with Prompt-to-Prompt.

Other methods similarly leverage diffusion inversion techniques paired with unique editing mechanisms: Diffusion Disentanglement~\cite{wu2023uncovering} aims to find the optimal blend of input and target prompt features to find the compromise between visual alignment with the input image and semantic alignment with the target prompt; EDICT~\cite{wallace2023edict} uses coupling inversion for improved reconstruction, which in turn enhance editing performance; pix2pix-zero~\cite{parmar2023zero} optimizes model attention maps for the target image to match the ones for the input image for visual consistency between the two; and Plug-and-Play~\cite{tumanyan2023plug} integrates self-attention maps from the input image to guide the edited image generation. While each method has its own advantages, a major shared shortcoming is they require prompts for both input and output images, which can be cumbersome for users.

\subsection{Instruction-guided image editing}

New approaches have recently emerged that leverage priors from Stable Diffusion and which offer a more user-friendly approach to image editing, requiring only a single edit instruction~\cite{brooks2023instructpix2pix,zhang2024magicbrush,zhang2023hive,yildirim2023inst,mirzaei2025watch,li2024zone}. For example, InstructPix2Pix (IP2P)~\cite{brooks2023instructpix2pix} replaces the prompt with an edit instruction that conditions the generation of the edited image, where the input image is concatenated as an extra condition as well. Among these instruction-guided image editing methods, HIVE~\cite{zhang2023hive} collects user feedback on outputs to train a reward model, which iteratively improves the model's output. Watch Your Step~\cite{mirzaei2025watch} and ZONE~\cite{li2024zone} both use a masking mechanism -- either by measuring the discrepancy between IP2P predictions with and without the instruction, or by employing a region intersection-over-union (IoU) scheme with a segmentation model. This helps them avoid editing instruction-irrelevant regions, resulting in better localized editing capabilities.

\subsection{Instruction-guided datasets}

Regardless what additional control mechanism is introduced to improve instruction-guided, image-editing results, the development of instruction-guided image editing models fundamentally relies on suitable training datasets. However, creating large-scale, high-quality datasets automatically presents a significant challenge, as such a data creation system would effectively constitute an editing method itself. Prior works have addressed this challenge by combining existing synthesis models to approximate the behavior of desired instruction-guided image editing methods, primarily following two distinct approaches. 

Several works~\cite{brooks2023instructpix2pix, zhao2024ultraedit, zhang2023hive} adopted the first approach, which leverages Prompt-to-Prompt~\cite{hertz2022prompt} to generate pairs of original/edited images using corresponding prompt pairs that describe the content of each image. These systems typically fine-tune a large language model (LLM) like GPT-3~\cite{brown2020language} to produce edit instructions from prompt pairs. While this approach generates diverse samples, the dataset quality is inherently constrained by the limitations of both Prompt-to-Prompt and the LLM, often resulting in misalignment between the edit instruction and the actual transformation from the original to the edited image (see Fig.~\ref{fig:dataset_samples}).

The second approach attempts to address these limitations by creating datasets through inpainting~\cite{yildirim2023inst,zhang2024magicbrush}. Here, edited ground truth images are generated by inpainting selected regions of original images. Although this technique improves the quality of localized image editing samples, it has two significant drawbacks: 1) it cannot generate samples with global modifications (such as style transfer), and 2) it often requires manual creation of inpainting regions. These limitations not only increase the cost of dataset creation but also substantially restrict the dataset size.

We now describe our approach to address the limitations in previous work.
 \section{\ourwork{}}

\begin{figure*} 
\centering 
\begin{subfigure}{0.45\linewidth} \includegraphics[width=\linewidth]{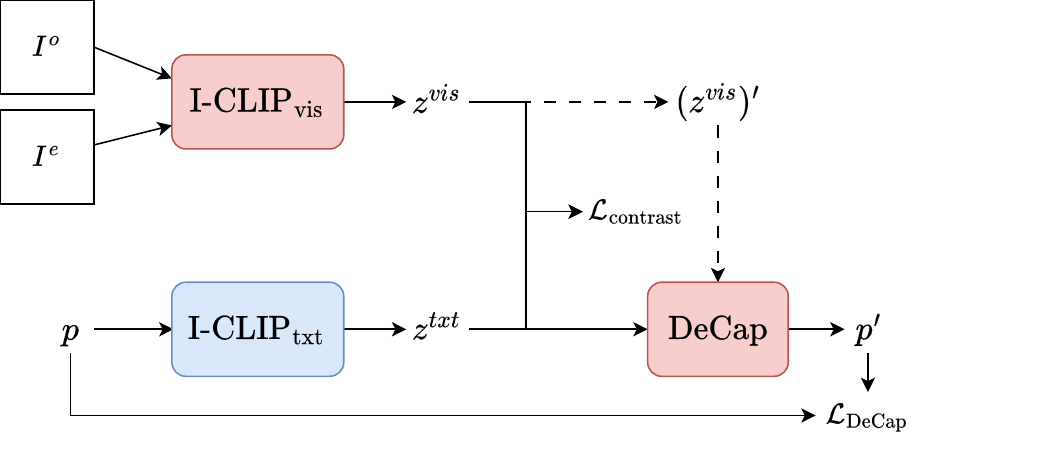}
\caption{Overall \ourwork{} system architecture}
\label{fig:overview} 
\end{subfigure} 
\hspace{20pt}
\begin{subfigure}{0.35\linewidth} 
\includegraphics[width=\linewidth]{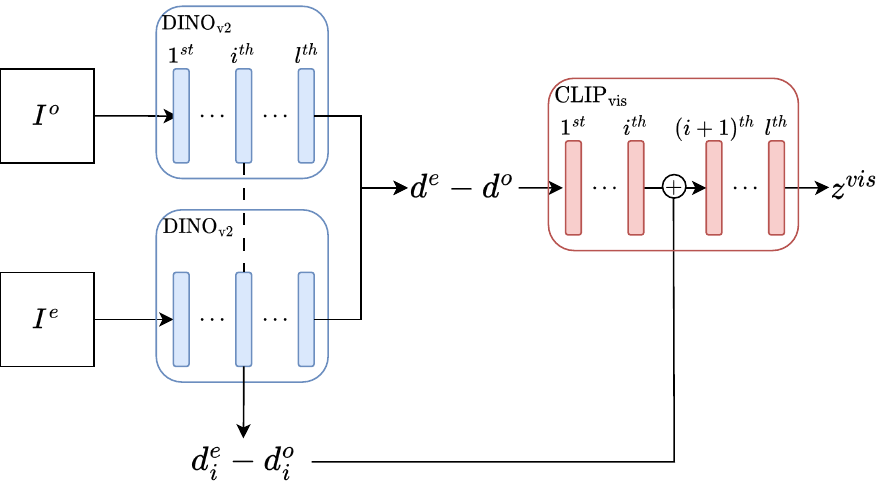}
\caption{\mbox{Architecture of visual-change encoder $\iclipvis(I^o, I^e)$}}
\label{fig:instructclip_indepth} 
\end{subfigure} 
\vspace{-0.05in}
\caption{\textbf{\ourwork{} architectures.}
{\bf (a)} Overview of \ourwork{} (\ourworkabbr{}), which embeds the {\em visual change} in the original/edited images $I^o$ and $I^e$ and the edit instruction $p$ into the same feature space through contrastive loss, $\mathcal{L}_\text{contrast}$ (Eq.~\ref{eq:loss_contrastive}). To obtain refined instruction $p$ from its \ourworkabbr{} embedding $z^\mytxt{}$, we adopt the same approach in DeCap~\cite{li2023decap} to decode $z^\mytxt{}$ back to $p$ using cross-entropy loss, $\mathcal{L}_\text{DeCap}$ (Eq.~\ref{eq:loss_decap}). At inference time, the text decoder takes the embedded visual change from the original to the edited image ($z^\myvis{}$) and decodes it to produce a new instruction. Due to the significant cosine similarity gap between $z^\myvis{}$ and $z^\mytxt{}$ even when they are well aligned, directly decoding $z^\myvis{}$ leads to suboptimal results. To achieve a representation of $z^\myvis{}$ closer to the text features that the instruction decoder learned during training, we compute $(z^\myvis{})'$ with Eq.~\ref{eq:decap_inference} and decode it to obtain the refined instruction $p'$, which is used to improve the dataset. {\bf (b)} The architecture of image encoder $\ourworkabbr{}_\myvis{}$ includes two shared-weighted \dinotext{}~\cite{oquab2023dinov2} modules in front of a standard $\text{CLIP}_\myvis{}$ encoder.
} 
\end{figure*}

Our goal is to improve previous instruction-guided image editing methods by enhancing the quality of available datasets~\cite{brooks2023instructpix2pix,zhang2023hive,hui2024hq,yang2024editworld}, which we propose to do by correcting their edit instructions. However, while instructions in these datasets sometimes do not reflect the actual changes between the original and edited images, we observe that the datasets overall still provide enough ``signal'' to train editing models. The main challenge is therefore how to properly leverage this ``signal'' to refine the given edit instructions and consequently improve the overall dataset quality.

To this end, we propose \ourwork{} (\ourworkabbr{}, for short), which embeds the {\em visual change} between the original image $I^o$ and the edited result $I^e$ into the same feature space as the corresponding edit instruction $p$ through contrastive learning with a neural network. Our approach is inspired by CLIP~\cite{radford2021learning}, which learns the semantic alignment between images and their captions. Likewise, \ourwork{} learns the relationship between the visual changes in the original-edit image pair and its edit instruction.

Just like in CLIP, our approach has an image encoder that in our case encodes the visual change between the input and edited images (denoted as $\iclipvis(I^o, I^e)$), and a text encoder that encodes the edit instruction ($\icliptxt(p)$). However, unlike the original CLIP image encoder, ours takes both the original and edited images as input so that it can encode their visual difference. As shown in Fig.~\ref{fig:overview}, we fine-tune the two encoders by computing and minimizing the contrastive loss between them (Eq.~\ref{eq:loss_contrastive}), similar to CLIP.

Furthermore, while the architecture of $\ourworkabbr{}_\mytxt{}$ is the same as CLIP, our $\ourworkabbr{}_\myvis{}$ adds two shared-weighted \dinotext{}~\cite{oquab2023dinov2} modules in front of $\text{CLIP}_\myvis{}$, as shown in Fig.~\ref{fig:instructclip_indepth}. These \dinotext{} units extract rich, robust visual features from the input images, allowing $\text{CLIP}_\myvis{}$ to focus on encoding the difference between the original and edited images. This is essential when we introduce Stable Diffusion's (SD) latent image encoding and diffusion into the network, as explained in Sec.~\ref{sec:latent_diffusion_dino}. 

To describe the architecture of $\ourworkabbr{}_\myvis{}$, we write the output of \dinotext{} for a given image $I$ as $d = \dinofunc(I)$, and the intermediate features from each of its $l$ layers as $\{d_i \mid i \in [1, l]\}$. So in $\ourworkabbr{}_\myvis{}$, we first we compute $d^o$ and $d^e$ by passing the input image $I^o$ and the edited image $I^e$ through \dinotext{}, respectively, then compute the \dinotext{} feature difference $d^e - d^o$ that we input to $\text{CLIP}_\myvis{}$. The intermediate feature differences $d_i^e - d_i^o$ are also added to input of the $i$th layer of $\text{CLIP}_\myvis{}$ before being processed by that layer. We find that this improves model performance in our experiments by providing additional information that may be lost from $d^e - d^o$ when being processed from the $(i+1)^{\text{th}}$ to $l^{\text{th}}$ layer, similar to skip connections~\cite{he2016deep}.

To train \ourwork{}, we first initialize $\text{CLIP}_\myvis{}$ and $\ourworkabbr{}_\mytxt{}$  with the respective blocks from the pre-trained CLIP model. Then, given a batch of $n$ data samples $\{(I_i^o, I_i^e, p_i) \mid i \in [1, n]\}$ from an existing instruction-guided image-editing dataset like InstructPix2Pix~\cite{brooks2023instructpix2pix}, we first compute the \ourwork{} features for both the visual difference and edit instruction: 
\begin{equation}
\begin{split}
    z_i^\myvis{} &= \iclipvis(I_i^o, I_i^e) \\
    z_i^\mytxt{} &= \icliptxt(p_i),
\end{split}
\end{equation}

\noindent and compute the contrastive loss $\mathcal{L}_\text{contrast}$ between them:
\begin{equation}
\small
\begin{aligned}
\mathcal{L}_\text{contrast} = 
-\frac{1}{n} \sum_{i=1}^{n} \Bigg( 
&\log \frac{\exp\left(\text{sim}(z_i^\myvis{}, z_i^\mytxt{}) / \tau\right)}
{\sum_{j=1}^{n} \exp\left(\text{sim}(z_i^\myvis{}, z_j^\mytxt{}) / \tau\right)} 
\\
+ &\log \frac{\exp\left(\text{sim}(z_i^\mytxt{}, z_i^\myvis{}) / \tau\right)}
{\sum_{j=1}^{n} \exp\left(\text{sim}(z_i^\mytxt{}, z_j^\myvis{}) / \tau\right)} 
\Bigg),
\end{aligned}
\label{eq:loss_contrastive}
\end{equation}

\noindent where scalar $\tau$ is a learnable ``temperature'' parameter that controls the sharpness of the similarity distribution and function $\text{sim}(\cdot, \cdot)$ measures normalized cosine similarity:
\begin{equation}
\text{sim}(z_1, z_2) = \frac{z_1 \cdot z_2}{\|z_1\| \|z_2\|}. 
\end{equation}
%
We first use this contrastive loss to fine-tune the $\ourworkabbr{}_\myvis{}$ and $\ourworkabbr{}_\mytxt{}$ modules in Fig.~\ref{fig:overview}. Once they are trained to embed the visual change and the instruction into the same feature space, we then use the approach of DeCap~\cite{li2023decap} to translate the features into text instructions as described next. 

\subsection{Predicting instructions by decoding features}
\label{sec:instruct_decoding}

Learning the alignment between original/edited image pairs and instructions does not automatically result in better instructions, since a decoder is required to translate \ourworkabbr{} latent-space features into actual text instructions. To do this, we leverage the approach of DeCap~\cite{li2023decap}, where we finetune pre-trained CLIP and GPT-2 decoding head~\cite{radford2019language} for our application to match the original edit instruction $p$ by either decoding $z^\myvis{} = \iclipvis(I^o, I^e)$ or $z^\mytxt{} = \icliptxt(p)$.

Since the original edit instructions $p$ in the datasets may not match the visible changes encoded in $z^\myvis{}$, we do not want to force the model to decode $z^\myvis{}$ directly to $p$ during training. Instead, we train it by decoding $z^\mytxt{}$ back to $p$, using the same approach as DeCap~\cite{li2023decap}. Formally, let $p = [c_1, c_2, \cdots, c_n]$ be an edit instruction's  token representation of length $n$, where each token $c_i$ is a one-hot vector. Similarly, the decoded/predicted instruction is $p' = [c_1', c_2', \cdots, c_n']$ (instructions can be represented by the same number of tokens through padding or truncation). This is basically a classification problem that requires identifying the correct ``element'' for each token, so the training objective of the text decoder is a cross-entropy loss:
\begin{equation}
    \mathcal{L}_\text{DeCap} = - \frac{1}{n} \sum_{i=1}^{n} \sum_{c \in C} y_i^p(c) \log y_i^{p'}(c),
\label{eq:loss_decap}
\end{equation} 
where $C$ is the set of all possible tokens, $y_i^p(c)$ is the ground-truth for each token $c$, and $y_i^{p'}(c)$ is the probability of token $c$ at position $i$ for $p'$ as predicted by the model. 

At inference time, our goal is to generate improved instructions that accurately capture the visual changes from the original image to the edited one by decoding $z^\myvis{}$. However, since the average cosine similarity between visual and text CLIP features is relatively low, even when well matched (around 0.2)~\cite{li2023decap}, a text decoder that has been trained on text features $z^\mytxt{}$ will struggle to handle $z^\myvis{}$ due to such differences, producing sub-optimal decoded results.

To address this, we would like the decoder to get as input a feature representation of $z^\myvis{}$ that is more similar to the text features the instruction decoder was trained on. A simple approach would be to compute the cosine similarity between $z^\myvis{}$ and each instruction's text feature in the dataset, selecting the feature with the highest similarity. However, this method would simply retrieve the most similar instruction, limiting both the diversity and accuracy of refined instructions. Instead, we follow the approach of DeCap~\cite{li2023decap}, which leverages information from all text features by weighting their influence by their cosine similarity to $z^\myvis$. Essentially, instructions with text features more similar to $z^\myvis$ should contribute more to the refined instruction.

Therefore, we calculate the probability of each instruction feature contributing to the refined representation using the softmax function over the cosine similarities between all text features and $z^\myvis$. Formally, for a dataset of size $n$, let $\{z_i^\mytxt = \icliptxt(p_i) \mid i \in [1, n]\}$ be the set of text features, each corresponding to an instruction $p_i$ in the dataset. The probability $w_i$ that instruction $p_i$'s text feature influences the refined instruction for $z^\myvis{}$ is defined as:
\begin{equation}
w_i = \frac{\exp{(\text{sim}(z_i^\mytxt, z^\myvis{}))}}{\sum_{j=1}^{n} \exp{(\text{sim}(z_j^\mytxt, z^\myvis{}))}}.
\end{equation}
We use these weights to project $z^\myvis{}$ to the text feature space:
\begin{equation}
(z^\myvis{})' = \sum_{i=1}^{n} (w_i \cdot z_i^\mytxt).
\label{eq:decap_inference}
\end{equation}
In the end, the refined instruction for $(I^o, I^e)$ is $p' = \text{DeCap}\left((z^\myvis{})'\right)$, where $\text{DeCap}(\cdot)$ denotes the instruction decoder, and so the refined dataset sample is $(I^o, I^e, p')$.

\subsection{Working in the latent-diffusion domain}
\label{sec:latent_diffusion_dino}

With \ourworkabbr{} and DeCap, we can now generate datasets with improved semantic alignment between original/edited image changes and their edit instructions (see results in Fig.~\ref{fig:dataset_samples}), thereby providing a better ``signal'' for training instruction-guided, image-editing models. So one might try to simply train models like InstructPix2Pix~\cite{brooks2023instructpix2pix} directly as-is on our improved dataset. While this does lead to some improvement (see Sec.~\ref{sec:ablation}), we find we get even more improvement if we reinforce the alignment between the visual change and the edit instruction {\em during} the training of the stable diffusion (SD) model itself. This means using \ourworkabbr{} as an integral part of the training objective beyond just dataset refinement.

\begin{figure}
    \centering
    \includegraphics[width=1.0\linewidth]{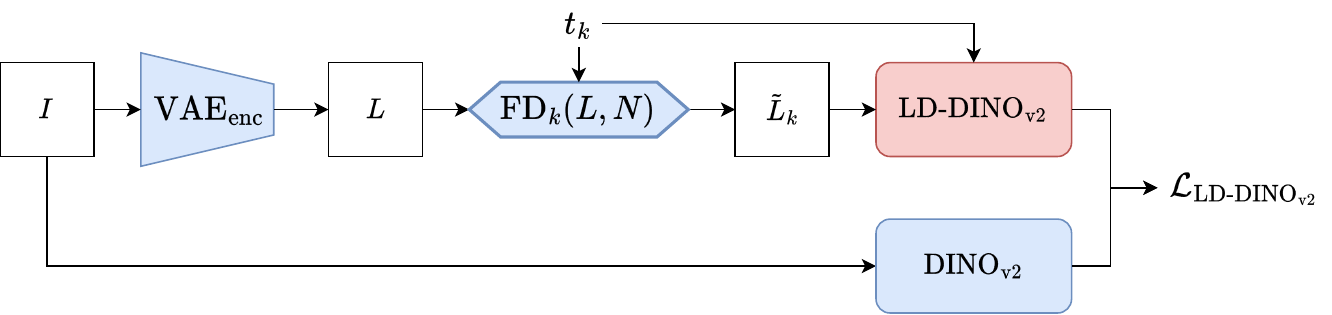}
    \vspace{-0.05in}
    \caption{\textbf{Training our \lddinotext{} model.} To use \ourworkabbr{} as part of the training objective for Stable Diffusion~\cite{rombach2022high}, it needs to handle noisy latent images. Therefore, we replace the original \dinotext{} backbone in \cref{fig:instructclip_indepth} with a latent-diffusion version of it we call \lddinotext{}, which takes both the noisy latent image $\tilde{L}_k$ from SD VAE encoding and forward-diffusion (FD) timestep $t_k$. We then train \lddinotext{} to ``ignore'' the noise and the latent-space compression and to extract the original \dinotext{} features using the training objective $\mathcal{L}_{\lddinofunc{}}$ (Eq.~\ref{eq:loss_lddinov2}).}
    \label{fig:lddino_training}
\end{figure}

To do this, \ourworkabbr{} must directly handle noisy, latent images in SD, defined as $\tilde{L}_k = \text{FD}(\text{VAE}_\text{enc}(I), N, t_k)$. Here, $\text{FD}(\cdot)$ is the $k^{\text{th}}$ step of the forward-diffusion process~\cite{song2020denoising} which takes in image $I$, embeds it in the latent domain with SD's variational autoencoder $\text{VAE}_\text{enc}(I)$, and then adds randomly sampled noise $N \sim \mathcal{N}(0, \mathbf{I})$ with respect to timestep $t_k$. Our \dinotext{} feature backbone in \ourworkabbr{}, which we call \lddinotext{} (see \cref{fig:lddino_training}), works with $\tilde{L}_k$ directly by also taking the corresponding timestep $t_k$ as input and is trained to ``ignore'' the noise and the latent-space compression to extract the original \dinotext{} features with the following training objective:
\begin{equation}
\small
\mathcal{L}_{\lddinofunc{}} = 1 - \text{sim}(d^{\text{LD}}\hspace{-0.04in}, d) + \frac{1}{l}\sum_{i=1}^{l}\left(1 - \text{sim}(d_i^{\text{LD}}\hspace{-0.04in}, d_i)\right),
\label{eq:loss_lddinov2}
\end{equation}
where $d^{\text{LD}}$ is the output of \lddinotext{}, $d$ is the original DINOv2 output, and the intermediate features from each of its $l$ layers are $\{d_i^{\text{LD}} \mid i \in [1, l]\}$. The visual feature of the original-edited image pair $(I^o, I^e)$ then becomes:
\begin{equation}
    z^\myvis{} = \iclipvis(\tilde{L}_{k_1}^o, t_{k_1}, \tilde{L}_{k_2}^e, t_{k_2}),
\end{equation}
\noindent where: 
\begin{equation}
\begin{split}
    \tilde{L}_{k_1}^o &= \text{FD}(\text{VAE}_\text{enc}(I^o), N_1, t_{k_1}) \\
    \tilde{L}_{k_2}^e &= \text{FD}(\text{VAE}_\text{enc}(I^e), N_2, t_{k_2}),
\end{split}
\end{equation}
are generated using randomly sampled Gaussian noise terms $N_1$ and $N_2$. 
Note that although for diffusion-based image editing training, no noise is added to the original image, we noisify the original image to train $\iclipvis$ to make it more robust.  For further details on the \lddinotext{} training process, please refer to the supplementary. Note that we set $k_1 = k_2 = 0$ during dataset refinement stage so \lddinotext{} will process the pristine latent images.

\subsection{Training our instruction-guided editing model}

At this point, we would like to leverage both the refined dataset and \ourworkabbr{} to enhance SD-based instruction-guided image editing methods, such as InstructPix2Pix (IP2P)~\cite{brooks2023instructpix2pix}.

Given data sample $(I^o, I^e, p)$, IP2P usually trains a denoising UNet to predict the noise added to the edited image:
\begin{equation}
    \tilde{N}_k = \text{UNet}_{\text{IP2P}}(L^o, \tilde{L}^e_k, t_k, p),
\end{equation}
which is conditioned on the original image in the latent domain ($L^o$) and the edit instruction $p$. The traditional training loss in this setup is the mean squared error (MSE) with respect to the actual noise $N$ added to the GT edited image:
\begin{equation}
\mathcal{L}_{\text{MSE}} = \left\| N - \tilde{N}_{k} \right\|_2^2,
\label{eq:mse_loss}
\end{equation}

Given refined sample $(I^o, I^e, p')$, we design a complementary loss function to incorporate our \ourwork{} guidance into the objective and force the visual change to match the refined edit instruction:
\begin{equation}
\begin{aligned}
\mathcal{L}_\ourworkabbr{} = 1 - \text{sim}\big(&\iclipvis(L^o, t_0, \tilde{L}^e_{k-1}, t_{k-1}), \\
&\icliptxt(p') \big),
\end{aligned}
\end{equation}
where the intermediate, denoised latent output using the predicted noise is calculated as:
\begin{equation}
\tilde{L}^e_{k-1} = \text{RD}_{k, k-1}(\tilde{L}^e_k, \tilde{N}_k).
\end{equation}
Here, $\text{RD}_{k, k-1}(\tilde{L}^e_k, \tilde{N}_k)$ is the reverse-diffusion process. Thus, our final training objective is:
\begin{equation}
\mathcal{L} = \mathcal{L}_\text{MSE} + \lambda\mathcal{L}_\ourworkabbr{},
\end{equation}
where $\lambda$ is set to $0.1$ to balance $\mathcal{L}_\ourworkabbr{}$ with the MSE loss.

\section{Experiments\label{sec:experiments}}

We implemented and trained \ourwork{} (\ourworkabbr{}) as described above and used it to refine the InstructPix2Pix (IP2P) dataset~\cite{brooks2023instructpix2pix} to get 120K+ refined instructions, which took around 10 hours on two A6000 GPUs. We then used this refined dataset to fine-tune the InstructPix2Pix model. Please refer to the supplementary for more details. We now describe the experiments to test our method.

\begin{figure*}
\centering
\resizebox{0.9\linewidth}{!}{ 
\begin{minipage}{\linewidth}

\begin{small}
    \begin{tabbing}
        \hspace{1.9em}\= \hspace{7.0em} \= \hspace{6.7em} \= \hspace{6.6em} \= \hspace{6.4em} \= \hspace{6.7em} \= \hspace{6.9em} \= \hspace{4.9em} \= \kill
        \> Original \> HIVE \> I-Inp \> WYS \> ZONE \> MagBr \> IP2P \> I-CLIP (Ours) \\
    \end{tabbing}
    \vspace{-0.27in}
    
    \begin{subfigure}{\linewidth}
        \begin{overpic}[width=\linewidth, trim=0 125px 0 0, clip]{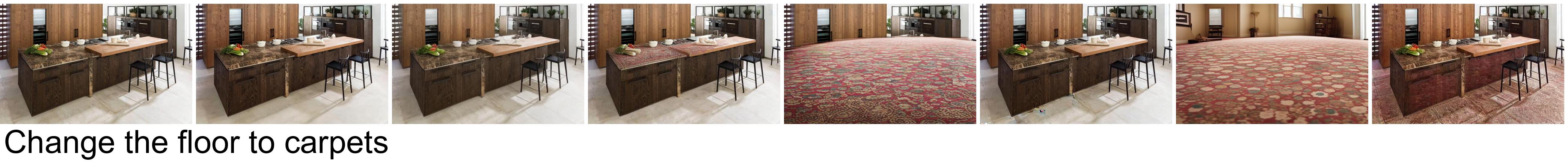}
            \put(12,6){
                \begin{tikzpicture}
                     \fill[black, opacity=0.5] (0,0) rectangle +(0.57,0.25); 
                    \node[anchor=west, text=white, inner sep=0pt] at (0,0.12) {\scriptsize 0.307};
                \end{tikzpicture}
            }

            \put(24.5,6){
                \begin{tikzpicture}
                     \fill[black, opacity=0.5] (0,0) rectangle +(0.57,0.25); 
                    \node[anchor=west, text=white, inner sep=0pt] at (0,0.12) {\scriptsize 0.303};
                \end{tikzpicture}
            }

            \put(37,6){
                \begin{tikzpicture}
                     \fill[black, opacity=0.5] (0,0) rectangle +(0.57,0.25); 
                    \node[anchor=west, text=white, inner sep=0pt] at (0,0.12) {\scriptsize \underline{0.325}};
                \end{tikzpicture}
            }

            \put(49.5,6){
                \begin{tikzpicture}
                     \fill[black, opacity=0.5] (0,0) rectangle +(0.57,0.25); 
                    \node[anchor=west, text=white, inner sep=0pt] at (0,0.12) {\scriptsize 0.225};
                \end{tikzpicture}
            }

            \put(62,6){
                \begin{tikzpicture}
                     \fill[black, opacity=0.5] (0,0) rectangle +(0.57,0.25); 
                    \node[anchor=west, text=white, inner sep=0pt] at (0,0.12) {\scriptsize 0.312};
                \end{tikzpicture}
            }

            \put(74.5,6){
                \begin{tikzpicture}
                     \fill[black, opacity=0.5] (0,0) rectangle +(0.57,0.25); 
                    \node[anchor=west, text=white, inner sep=0pt] at (0,0.12) {\scriptsize 0.224};
                \end{tikzpicture}
            }

            \put(87,6){
                \begin{tikzpicture}
                     \fill[black, opacity=0.5] (0,0) rectangle +(0.57,0.25); 
                    \node[anchor=west, text=white, inner sep=0pt] at (0,0.12) {\scriptsize 0.317};
                \end{tikzpicture}
            }
        \end{overpic}
        \parbox{\linewidth}{\vspace{-0.03in}
        ``Change the floor to carpets''
        \vspace{0.05in}}
    \end{subfigure}

    \begin{subfigure}{\linewidth}
        \begin{overpic}[width=\linewidth, trim=0 125px 0 0, clip]{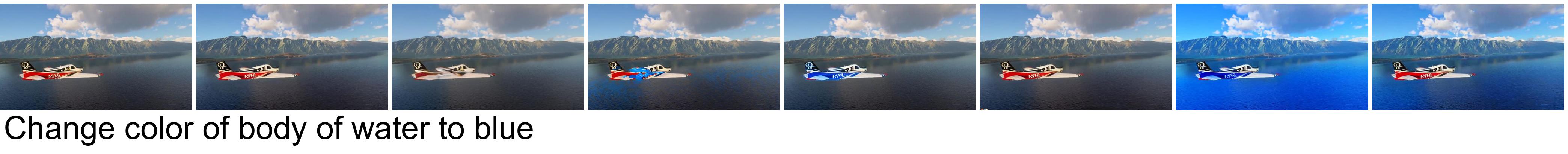}
            \put(12,5.1){
                \begin{tikzpicture}
                     \fill[black, opacity=0.4] (0,0) rectangle +(0.57,0.25); 
                    \node[anchor=west, text=white, inner sep=0pt] at (0,0.12) {\scriptsize 0.278};
                \end{tikzpicture}
            }

            \put(24.5,5.1){
                \begin{tikzpicture}
                     \fill[black, opacity=0.4] (0,0) rectangle +(0.57,0.25); 
                    \node[anchor=west, text=white, inner sep=0pt] at (0,0.12) {\scriptsize 0.276};
                \end{tikzpicture}
            }

            \put(37,5.1){
                \begin{tikzpicture}
                     \fill[black, opacity=0.4] (0,0) rectangle +(0.57,0.25); 
                    \node[anchor=west, text=white, inner sep=0pt] at (0,0.12) {\scriptsize 0.290};
                \end{tikzpicture}
            }

            \put(49.5,5.1){
                \begin{tikzpicture}
                     \fill[black, opacity=0.5] (0,0) rectangle +(0.57,0.25); 
                    \node[anchor=west, text=white, inner sep=0pt] at (0,0.12) {\scriptsize 0.299};
                \end{tikzpicture}
            }

            \put(62,5.1){
                \begin{tikzpicture}
                     \fill[black, opacity=0.4] (0,0) rectangle +(0.57,0.25); 
                    \node[anchor=west, text=white, inner sep=0pt] at (0,0.12) {\scriptsize 0.279};
                \end{tikzpicture}
            }

            \put(74.5,5.1){
                \begin{tikzpicture}
                     \fill[black, opacity=0.4] (0,0) rectangle +(0.57,0.25); 
                    \node[anchor=west, text=white, inner sep=0pt] at (0,0.12) {\scriptsize \underline{0.304}};
                \end{tikzpicture}
            }

            \put(87,5.1){
                \begin{tikzpicture}
                     \fill[black, opacity=0.4] (0,0) rectangle +(0.57,0.25); 
                    \node[anchor=west, text=white, inner sep=0pt] at (0,0.12) {\scriptsize 0.287};
                \end{tikzpicture}
            }

        \end{overpic}
        \parbox{\linewidth}{\vspace{-0.03in}
        ``Change color of body of water to blue''
        \vspace{0.05in}}
    \end{subfigure}

    \begin{subfigure}{\linewidth}
        \begin{overpic}[width=\linewidth, trim=0 125px 0 0, clip]{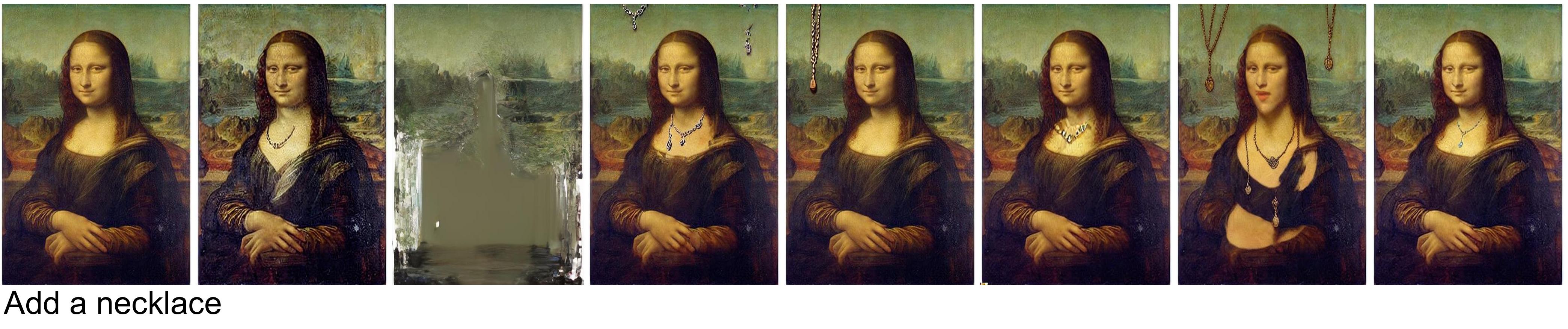}
            \put(12.9,16.4){\textcolor{white}{\scriptsize 0.362}}
            \put(25.4,16.4){\textcolor{white}{\scriptsize 0.241}}
            \put(37.9,16.4){\textcolor{white}{\scriptsize \underline{0.369}}}
            \put(50.4,16.4){\textcolor{white}{\scriptsize \underline{0.369}}}
            \put(62.9,16.4){\textcolor{white}{\scriptsize 0.362}}
            \put(75.4,16.4){\textcolor{white}{\scriptsize 0.365}}
            \put(87.9,16.4){\textcolor{white}{\scriptsize 0.360}}
        \end{overpic}
        \parbox{\linewidth}{\vspace{-0.03in}
        ``Add a necklace''
        \vspace{0.05in}}
    \end{subfigure}

    \begin{subfigure}{\linewidth}
        \begin{overpic}[width=\linewidth, trim=0 125px 0 0, clip]{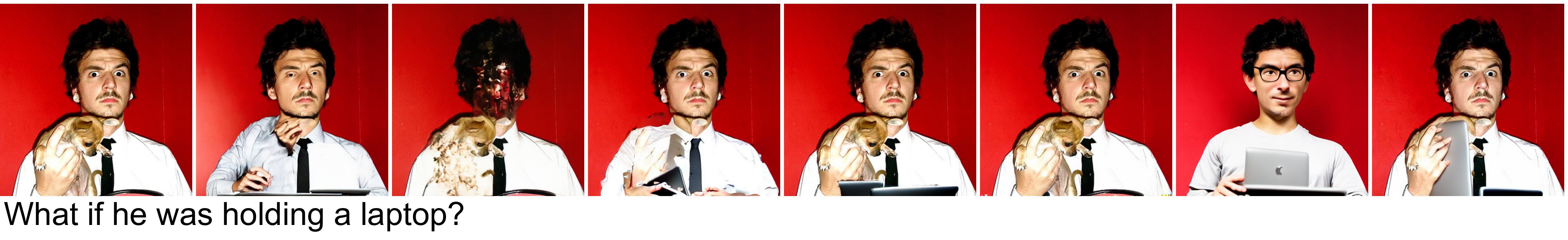}
            \put(12.7,10.8){\textcolor{white}{\scriptsize 0.238}}
            \put(25.2,10.8){\textcolor{white}{\scriptsize 0.196}}
            \put(37.7,10.8){\textcolor{white}{\scriptsize 0.216}}
            \put(50.2,10.8){\textcolor{white}{\scriptsize \underline{0.280}}}
            \put(62.7,10.8){\textcolor{white}{\scriptsize 0.267}}
            \put(75.2,10.8){\textcolor{white}{\scriptsize 0.243}}
            \put(87.7,10.8){\textcolor{white}{\scriptsize 0.240}}
        \end{overpic}
        \parbox{\linewidth}{\vspace{-0.03in}
        ``What if he was holding a laptop?''
        \vspace{0.05in}}
    \end{subfigure}

    \begin{subfigure}{\linewidth}
        \begin{overpic}[width=\linewidth, trim=0 125px 0 0, clip]{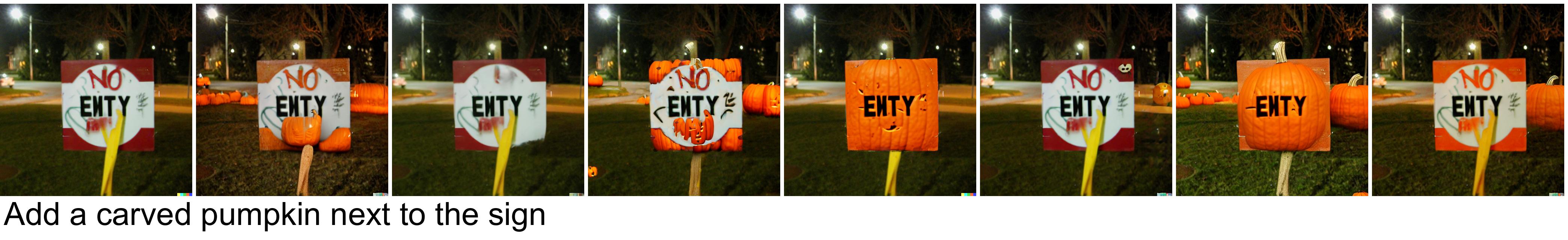}
            \put(12,10.6){
                \begin{tikzpicture}
                     \fill[black, opacity=0.5] (0,0) rectangle +(0.57,0.25); 
                    \node[anchor=west, text=white, inner sep=0pt] at (0,0.12) {\scriptsize 0.375};
                \end{tikzpicture}
            }

            \put(24.5,10.6){
                \begin{tikzpicture}
                     \fill[black, opacity=0.5] (0,0) rectangle +(0.57,0.25); 
                    \node[anchor=west, text=white, inner sep=0pt] at (0,0.12) {\scriptsize 0.284};
                \end{tikzpicture}
            }

            \put(37,10.6){
                \begin{tikzpicture}
                     \fill[black, opacity=0.5] (0,0) rectangle +(0.57,0.25); 
                    \node[anchor=west, text=white, inner sep=0pt] at (0,0.12) {\scriptsize 0.381};
                \end{tikzpicture}
            }

            \put(49.5,10.6){
                \begin{tikzpicture}
                     \fill[black, opacity=0.5] (0,0) rectangle +(0.57,0.25); 
                    \node[anchor=west, text=white, inner sep=0pt] at (0,0.12) {\scriptsize 0.375};
                \end{tikzpicture}
            }

            \put(62,10.6){
                \begin{tikzpicture}
                     \fill[black, opacity=0.5] (0,0) rectangle +(0.57,0.25); 
                    \node[anchor=west, text=white, inner sep=0pt] at (0,0.12) {\scriptsize \underline{0.395}};
                \end{tikzpicture}
            }

            \put(74.5,10.6){
                \begin{tikzpicture}
                     \fill[black, opacity=0.5] (0,0) rectangle +(0.57,0.25); 
                    \node[anchor=west, text=white, inner sep=0pt] at (0,0.12) {\scriptsize 0.351};
                \end{tikzpicture}
            }

            \put(87,10.6){
                \begin{tikzpicture}
                     \fill[black, opacity=0.5] (0,0) rectangle +(0.57,0.25); 
                    \node[anchor=west, text=white, inner sep=0pt] at (0,0.12) {\scriptsize 0.384};
                \end{tikzpicture}
            }
        \end{overpic}
        \parbox{\linewidth}{\vspace{-0.03in}
        ``Add a carved pumpkin next to the sign''
        \vspace{0.05in}}
    \end{subfigure}

    \begin{subfigure}{\linewidth}
        \begin{overpic}[width=\linewidth, trim=0 125px 0 0, clip]{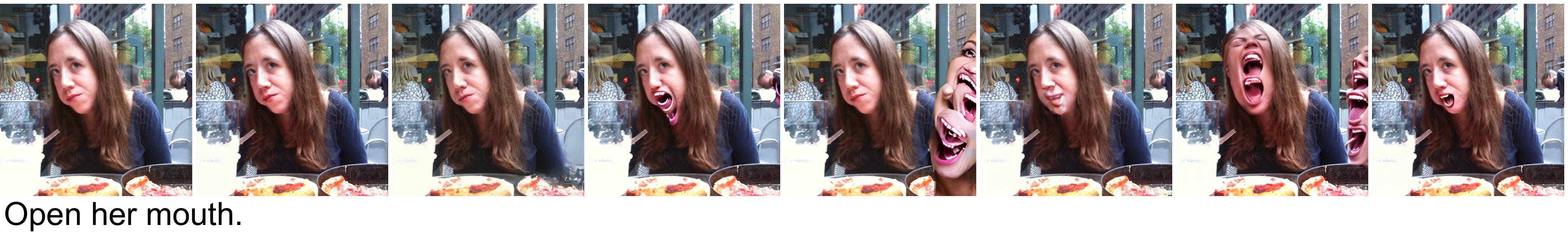}
            \put(12,10.6){
                \begin{tikzpicture}
                     \fill[black, opacity=0.3] (0,0) rectangle +(0.57,0.25); 
                    \node[anchor=west, text=white, inner sep=0pt] at (0,0.12) {\scriptsize 0.311};
                \end{tikzpicture}
            }

            \put(24.5,10.6){
                \begin{tikzpicture}
                     \fill[black, opacity=0.3] (0,0) rectangle +(0.57,0.25); 
                    \node[anchor=west, text=white, inner sep=0pt] at (0,0.12) {\scriptsize 0.311};
                \end{tikzpicture}
            }

            \put(37,10.6){
                \begin{tikzpicture}
                     \fill[black, opacity=0.3] (0,0) rectangle +(0.57,0.25); 
                    \node[anchor=west, text=white, inner sep=0pt] at (0,0.12) {\scriptsize \underline{0.335}};
                \end{tikzpicture}
            }

            \put(49.5,10.6){
                \begin{tikzpicture}
                     \fill[black, opacity=0.3] (0,0) rectangle +(0.57,0.25); 
                    \node[anchor=west, text=white, inner sep=0pt] at (0,0.12) {\scriptsize 0.287};
                \end{tikzpicture}
            }

            \put(62,10.6){
                \begin{tikzpicture}
                     \fill[black, opacity=0.3] (0,0) rectangle +(0.57,0.25); 
                    \node[anchor=west, text=white, inner sep=0pt] at (0,0.12) {\scriptsize 0.324};
                \end{tikzpicture}
            }

            \put(74.5,10.6){
                \begin{tikzpicture}
                     \fill[black, opacity=0.3] (0,0) rectangle +(0.57,0.25); 
                    \node[anchor=west, text=white, inner sep=0pt] at (0,0.12) {\scriptsize 0.295};
                \end{tikzpicture}
            }

            \put(87,10.6){
                \begin{tikzpicture}
                     \fill[black, opacity=0.3] (0,0) rectangle +(0.57,0.25); 
                    \node[anchor=west, text=white, inner sep=0pt] at (0,0.12) {\scriptsize 0.332};
                \end{tikzpicture}
            }
        \end{overpic}
        \parbox{\linewidth}{\vspace{-0.03in}
        ``Open her mouth.''
        \vspace{0.05in}}
    \end{subfigure}

    \begin{subfigure}{\linewidth}
        \begin{overpic}[width=\linewidth, trim=0 125px 0 0, clip]{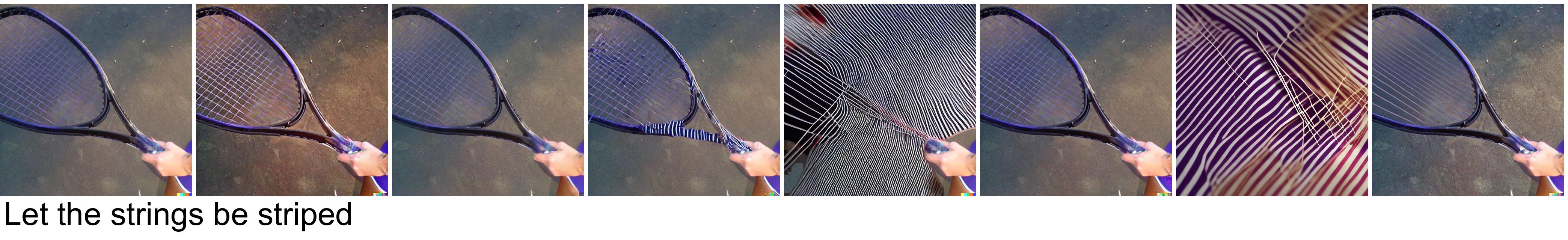}
            \put(12,10.6){
                \begin{tikzpicture}
                     \fill[black, opacity=0.5] (0,0) rectangle +(0.57,0.25); 
                    \node[anchor=west, text=white, inner sep=0pt] at (0,0.12) {\scriptsize 0.318};
                \end{tikzpicture}
            }

            \put(24.5,10.6){
                \begin{tikzpicture}
                     \fill[black, opacity=0.5] (0,0) rectangle +(0.57,0.25); 
                    \node[anchor=west, text=white, inner sep=0pt] at (0,0.12) {\scriptsize 0.315};
                \end{tikzpicture}
            }

            \put(37,10.6){
                \begin{tikzpicture}
                     \fill[black, opacity=0.5] (0,0) rectangle +(0.57,0.25); 
                    \node[anchor=west, text=white, inner sep=0pt] at (0,0.12) {\scriptsize \underline{0.342}};
                \end{tikzpicture}
            }

            \put(49.5,10.6){
                \begin{tikzpicture}
                     \fill[black, opacity=0.5] (0,0) rectangle +(0.57,0.25); 
                    \node[anchor=west, text=white, inner sep=0pt] at (0,0.12) {\scriptsize 0.273};
                \end{tikzpicture}
            }

            \put(62,10.6){
                \begin{tikzpicture}
                     \fill[black, opacity=0.5] (0,0) rectangle +(0.57,0.25); 
                    \node[anchor=west, text=white, inner sep=0pt] at (0,0.12) {\scriptsize 0.306};
                \end{tikzpicture}
            }

            \put(74.5,10.6){
                \begin{tikzpicture}
                     \fill[black, opacity=0.5] (0,0) rectangle +(0.57,0.25); 
                    \node[anchor=west, text=white, inner sep=0pt] at (0,0.12) {\scriptsize 0.217};
                \end{tikzpicture}
            }

            \put(87,10.6){
                \begin{tikzpicture}
                     \fill[black, opacity=0.5] (0,0) rectangle +(0.57,0.25); 
                    \node[anchor=west, text=white, inner sep=0pt] at (0,0.12) {\scriptsize 0.305};
                \end{tikzpicture}
            }
        \end{overpic}
        \parbox{\linewidth}{\vspace{-0.03in}
        ``Let the strings be striped''
        \vspace{0.05in}}
    \end{subfigure}

    \begin{subfigure}{\linewidth}
        \begin{overpic}[width=\linewidth, trim=0 125px 0 0, clip]{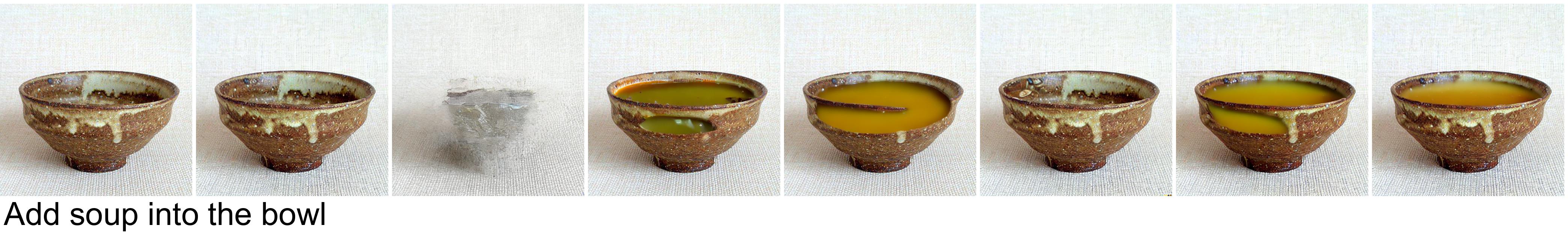}
            \put(12.7,10.8){\textcolor{black}{\scriptsize 0.314}}
            \put(25.2,10.8){\textcolor{black}{\scriptsize 0.232}}
            \put(37.7,10.8){\textcolor{black}{\scriptsize 0.327}}
            \put(50.2,10.8){\textcolor{black}{\scriptsize 0.333}}
            \put(62.7,10.8){\textcolor{black}{\scriptsize 0.313}}
            \put(75.2,10.8){\textcolor{black}{\scriptsize \underline{0.337}}}
            \put(87.7,10.8){\textcolor{black}{\scriptsize 0.333}}
        \end{overpic}
        \parbox{\linewidth}{\vspace{-0.03in}
        ``Add soup into the bowl''
        \vspace{0.05in}}
    \end{subfigure}

    \begin{subfigure}{\linewidth}
        \begin{overpic}[width=\linewidth, trim=0 125px 0 0, clip]{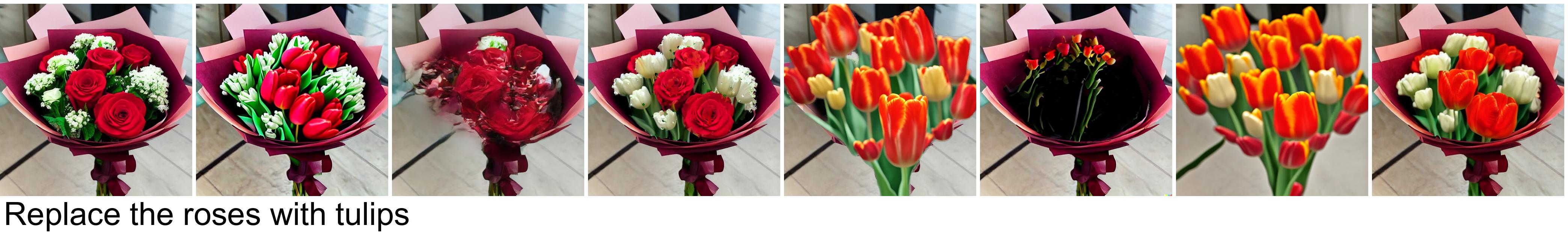}
            \put(12,10.6){
                \begin{tikzpicture}
                     \fill[black, opacity=0.5] (0,0) rectangle +(0.57,0.25); 
                    \node[anchor=west, text=white, inner sep=0pt] at (0,0.12) {\scriptsize 0.314};
                \end{tikzpicture}
            }

            \put(24.5,10.6){
                \begin{tikzpicture}
                     \fill[black, opacity=0.5] (0,0) rectangle +(0.57,0.25); 
                    \node[anchor=west, text=white, inner sep=0pt] at (0,0.12) {\scriptsize 0.256};
                \end{tikzpicture}
            }

            \put(37,10.6){
                \begin{tikzpicture}
                     \fill[black, opacity=0.5] (0,0) rectangle +(0.57,0.25); 
                    \node[anchor=west, text=white, inner sep=0pt] at (0,0.12) {\scriptsize 0.296};
                \end{tikzpicture}
            }

            \put(49.5,10.6){
                \begin{tikzpicture}
                     \fill[black, opacity=0.5] (0,0) rectangle +(0.57,0.25); 
                    \node[anchor=west, text=white, inner sep=0pt] at (0,0.12) {\scriptsize 0.311};
                \end{tikzpicture}
            }

            \put(62,10.6){
                \begin{tikzpicture}
                     \fill[black, opacity=0.5] (0,0) rectangle +(0.57,0.25); 
                    \node[anchor=west, text=white, inner sep=0pt] at (0,0.12) {\scriptsize 0.294};
                \end{tikzpicture}
            }

            \put(74.5,10.6){
                \begin{tikzpicture}
                     \fill[black, opacity=0.5] (0,0) rectangle +(0.57,0.25); 
                    \node[anchor=west, text=white, inner sep=0pt] at (0,0.12) {\scriptsize 0.308};
                \end{tikzpicture}
            }

            \put(87,10.6){
                \begin{tikzpicture}
                     \fill[black, opacity=0.5] (0,0) rectangle +(0.57,0.25); 
                    \node[anchor=west, text=white, inner sep=0pt] at (0,0.12) {\scriptsize \underline{0.315}};
                \end{tikzpicture}
            }
        \end{overpic}
        \parbox{\linewidth}{\vspace{-0.03in}
        ``Replace the roses with tulips''
        \vspace{0.05in}}
    \end{subfigure}
\end{small}
\end{minipage}
} 
\vspace{-0.2in}
\caption{Comparison with state-of-the-art approaches for instruction-guided  image editing, including HIVE~\cite{zhang2023hive}, Inst-Inpaint (I-Inp)~\cite{yildirim2023inst}, Watch Your Steps (WYS)~\cite{mirzaei2025watch}, ZONE~\cite{li2024zone}, MagicBrush (MagBr)~\cite{zhang2024magicbrush}, InstructPix2Pix (IP2P)~\cite{brooks2023instructpix2pix} showcasing the strength of our approach. CLIP-T value of each output is shown at its top-left corner, with the best value per row underlined. Note that the image with the best CLIP-T score is not necessarily the visually best result, underscoring the deficiencies of conventional metrics (including CLIP-I and DINO-I shown in the supplemental) for measuring the quality of image edits.}
\label{fig:qualitative}
\end{figure*}
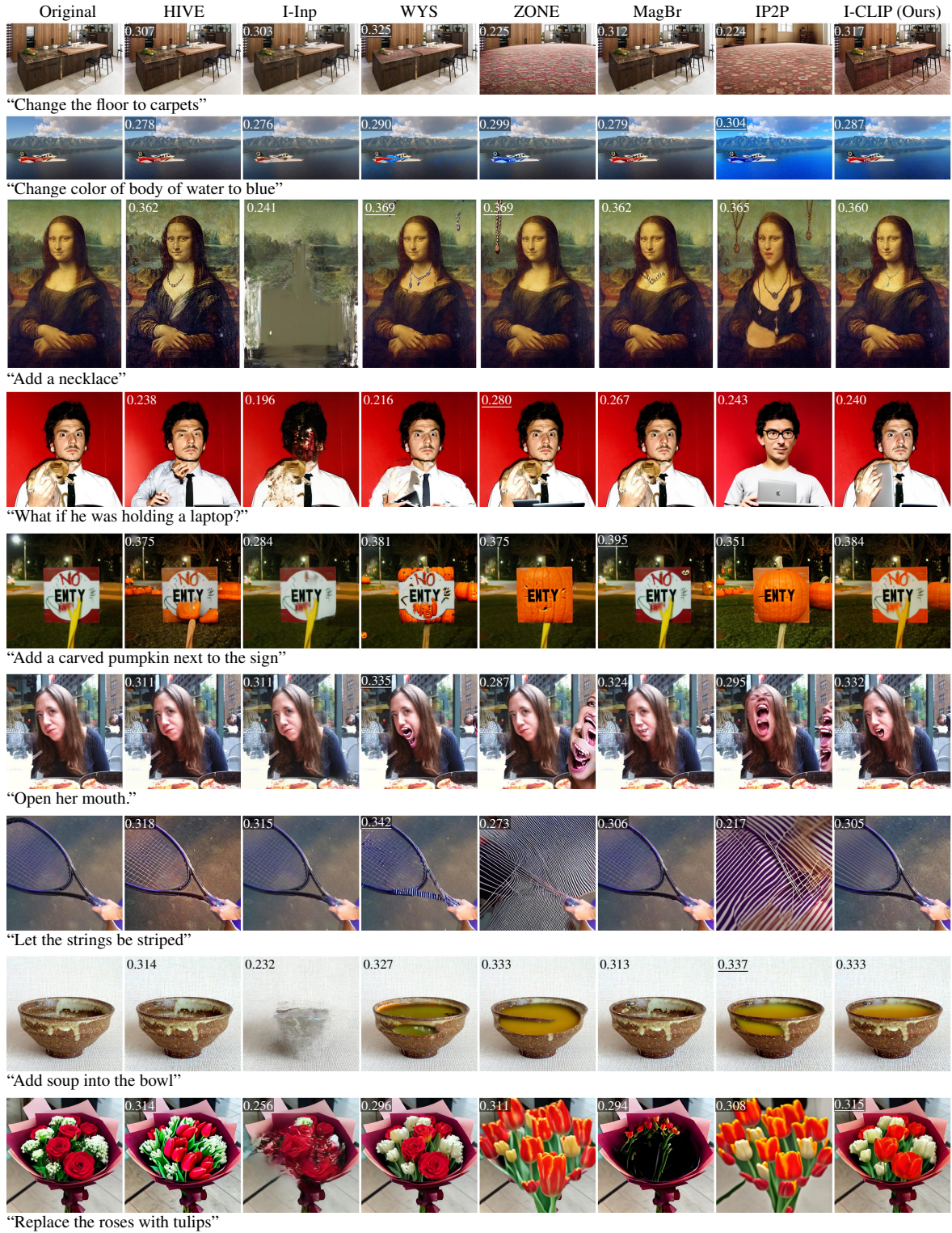

\subsection{Baselines and benchmarks}


We compare our editing model with several baselines: Inst-Inpaint (I-Inp)~\cite{yildirim2023inst}, MagicBrush (MagBr), HIVE~\cite{zhang2023hive}, InstructPix2Pix (IP2P)~\cite{brooks2023instructpix2pix}, Watch Your Steps (WYS)~\cite{mirzaei2025watch}, and ZONE~\cite{li2024zone} (see Sec.~\ref{sec:related_work}). We evaluate methods on two instruction-guided image-editing benchmarks: MagicBrush (MagBr)~\cite{zhang2024magicbrush} and ZONE~\cite{li2024zone}. The former contains {\em multi-turn} edits where multiple instructions are used to edit one image iteratively, as opposed to {\em single-turn} edit where the image is edited once. We compare results quantitatively with the following metrics used in prior work~\cite{brooks2023instructpix2pix,zhang2024magicbrush,mirzaei2025watch,li2024zone}:
\begin{itemize}
    \item $\text{CLIP-T} = \text{sim}(\text{CLIP}_\text{vis}((I^e)'), \text{CLIP}_\text{txt}(p^e))$
    \item $\text{CLIP-I} = \text{sim}(\text{CLIP}_\text{vis}((I^e)'), \text{CLIP}_\text{vis}(I^e))$
    \item $\text{DINO-I} = \text{sim}(\text{DINO}_\text{v2}((I^e)'), \text{DINO}_\text{v2}(I^e))$
\end{itemize}

\noindent Here, $(I^e)'$ denotes the method output corresponding to a benchmark sample $(I^o, I^e, p^o, p^e)$, where $I^o$ is the original image, $I^e$ is the ground-truth edited image (available for MagBr, not for ZONE), $p^o$ is the caption for the original image, and $p^e$ is the intended caption for the edited image. 
The CLIP-T score assesses semantic alignment between the result and its intended caption in the benchmark while the CLIP and DINO similarity scores (CLIP-I and DINO-I, respectively) evaluate the visual alignment between the result and the ground-truth, if available. However, we note that as shown in Fig.~\ref{fig:qualitative} and in the supplemental material, these metrics do not always match the visual quality of the image edits, but are presented here for completeness.

\subsection{Instruction refinement results}

We present samples from our refined dataset in~Fig.~\ref{fig:dataset_samples} (see supplementary for more samples). As we can see, \ourwork{} is able to correct wrong instructions such as ``make the waves into a hurricane'' to ``add a lightning storm to the sky'' ($1^{st}$ row right). By correcting these samples, we reduce the noise due to instructions not reflecting the actual edits, which in turn helps improve the performance of the model trained on this dataset as we will see next.

\subsection{Image-editing results}
\label{sec:editing_results}

We found a diverse set of samples that showcase the strength of our method and compared it to baselines in \cref{fig:teaser,fig:qualitative} (see supplemental for more). Our model not only addresses many issues present in IP2P -- such as unintended changes to regions irrelevant to the instructions -- but also generates results aligned more accurately with the edit instructions. This is particularly important for multi-turn editing to avoid results diverse further from the desired outcome across edit turns as shown in the supplementary. Notably, this is achieved {\em without} manually created datasets or masking mechanisms to explicitly locate the editing region, as used in other approaches. 

Furthermore, we present the CLIP-T value of each output in Fig.~\ref{fig:qualitative}, with the best value per sample underlined. As shown, these values -- as well as the CLIP-I and DINO-I values provided in the supplementary material -- are not indicative of visual quality, but are included for completeness. Table~\ref{tab:quantitative} provides a quantitative comparison of our model in both single-turn and multi-turn edit scenarios.

\begin{table}
\centering
\setlength{\tabcolsep}{2pt}
\resizebox{\columnwidth}{!}{  
\begin{tabular}{l|ccc|ccc|c}
    \multirow{2}{*}{} & \multicolumn{3}{c|}{MagBr data (single-turn)} & \multicolumn{3}{c|}{MagBr data (multi-turn)} & ZONE data \\
    \cline{2-8}
    & CLIP-T $\uparrow$ & CLIP-I $\uparrow$ & DINO-I $\uparrow$ & CLIP-T $\uparrow$ & CLIP-I $\uparrow$ & DINO-I $\uparrow$ & CLIP-T $\uparrow$ \\
    \midrule
    HIVE              & 0.303 & 0.892 & 0.746 & 0.299 & 0.852 & 0.659 & 0.297 \\
    I-Inp             & 0.285 & 0.887 & 0.729 & 0.277 & 0.859 & 0.654 & 0.267 \\
    MagBr                & 0.307 & 0.929 & 0.836 & 0.303 & 0.896 & 0.759 & 0.292 \\
    WYS               & 0.313 & 0.924 & 0.815 & 0.313 & 0.887 & 0.727 & 0.301 \\
    ZONE              & 0.301 & 0.929 & 0.824 & 0.307 & 0.896 & 0.750 & 0.296 \\
    \hdashline
    IP2P              & 0.300 & 0.854 & 0.645 & 0.298 & 0.824 & 0.573 & 0.296 \\
    Ours              & 0.305 & 0.911 & 0.803 & 0.301 & 0.871 & 0.721 & 0.297 \\
    & (\textcolor{darkgreen}{+1.67\%}) & (\textcolor{darkgreen}{+6.67\%}) & (\textcolor{darkgreen}{+24.5\%}) & (\textcolor{darkgreen}{+1.01\%}) & (\textcolor{darkgreen}{+5.70\%}) & (\textcolor{darkgreen}{+25.83\%}) & (\textcolor{darkgreen}{+0.34\%}) \\
\end{tabular}
}
\vspace{-0.1in}
\caption{Quantitative comparison with baselines. An up arrow ($\uparrow$) indicates higher values are better. The percentage improvements over baselines are in parentheses. }
\label{tab:quantitative}
\end{table}

\begin{table}[]
\centering
\setlength{\tabcolsep}{2pt}
\resizebox{\columnwidth}{!}{ 
\begin{tabular}{|l|c|c||l|c|c||l|c|c|}
\hline
vs. MagBr & \# & \% & vs. IP2P & \# & \% & IP2P vs.\ MagBr & \# & \% \\
\hline
ours win & 566 & 54.16\% & ours win & 310 & 29.67\% & IP2P win & 485 & 46.41\% \\
tie & 168 & 16.08\% & tie & 611 & 58.47\% & tie & 148 & 14.16\% \\
MagBr win & 311 & 29.76\% & IP2P win & 124 & 11.87\% & MagBr win & 412 & 39.43\% \\
\hline
\end{tabular}
}
\vspace{-0.1in}
\caption{Method pairwise comparison responses from user study.}
\label{tab:win_lose}
\end{table}

\subsection{User study}

We also conducted a user study with 104 participants (60 male, 43 female, 1 non-binary) to compare our method with MagicBrush (MagBr)~\cite{zhang2024magicbrush} and InstructPix2Pix (IP2P)~\cite{brooks2023instructpix2pix} on the InstructBrush benchmark~\cite{zhao2024instructbrush}, ensuring fairness since no method was trained on similar data. Each person evaluated 33 randomly sampled output pairs based on the input image and edit instruction, and select which was better or if they were similar. We balanced the study across all pairwise comparisons, with 1,045 responses per pair. As shown in Table~\ref{tab:win_lose}, participants overall found our method better than IP2P (17.8\% more wins) and MagBr (24.4\%).

\subsection{Ablation study\label{sec:ablation}}

\begin{table}
\centering
\small
\setlength{\tabcolsep}{2pt}
\begin{tabular}{l|ccc|c}
\multirow{2}{*}{} & \multicolumn{3}{c|}{MagBr data} & ZONE data \\
\cline{2-5}
                  & CLIP-T $\uparrow$ & CLIP-I $\uparrow$ & DINO-I $\uparrow$ & CLIP-T $\uparrow$ \\
\midrule
IP2P              & 0.300 & 0.854 & 0.645 & 0.296 \\
Ours (w/o data)   & 0.299 & 0.862 & 0.671 & 0.295 \\
Ours (w/o loss)   & 0.303 & 0.903 & 0.782 & 0.298 \\
Ours              & 0.305 & 0.911 & 0.803 & 0.297 \\
\end{tabular}
\vspace{-0.1in}
\caption{Ablation study. An up arrow ($\uparrow$) means higher values are better. We trained two variants for our model: one trained on the original IP2P dataset without refined instructions (`Ours (w/o data)'') and the other trained on our refined dataset without the \ourwork{} guidance loss (``Ours (w/o loss)''). We can see both the refined data and the loss help improve results.}
\vspace{-0.1in}
\label{tab:ablation}
\end{table}

\begin{figure}
    \centering
    \begin{scriptsize}
        \begin{tabbing}
            \hspace{1.8em}\= \hspace{6.9em} \= \hspace{4.7em} \= \hspace{6.6em} \= \hspace{8.5em} \= \kill
            \> Original \> IP2P \> Ours (w/o data) \> Ours (w/o loss) \> Ours \\
    \end{tabbing}
    \end{scriptsize}
    \vspace{-0.27in}

    \begin{footnotesize}
    \begin{subfigure}{\linewidth}
        \includegraphics[width=\linewidth, trim=0 125px 0 0, clip]{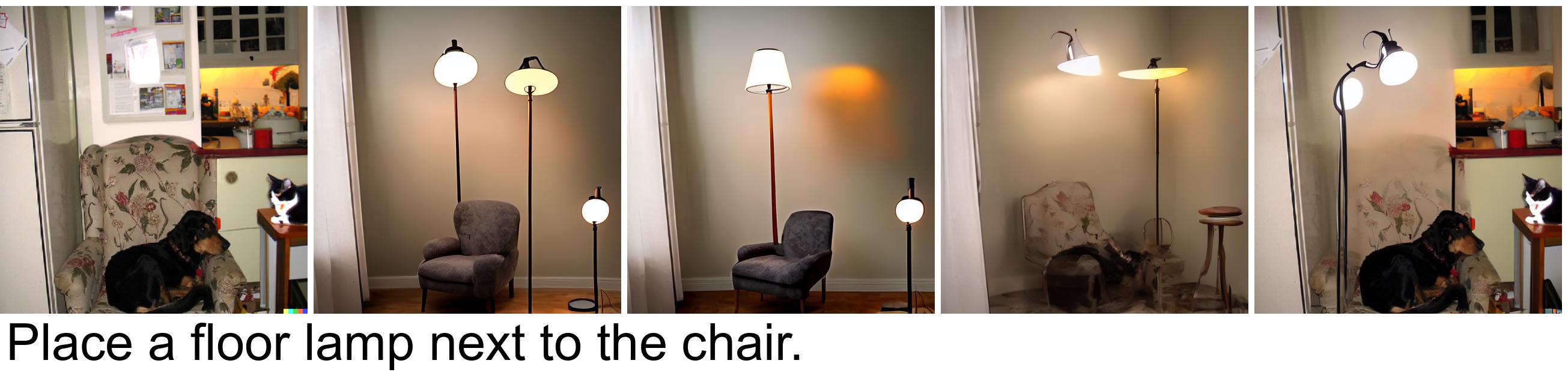}
        \parbox{\linewidth}{
        ``Place a floor lamp next to the chair.''
        \vspace{0.02in}}
    \end{subfigure}

    \begin{subfigure}{\linewidth}
        \includegraphics[width=\linewidth, trim=0 125px 0 0, clip]{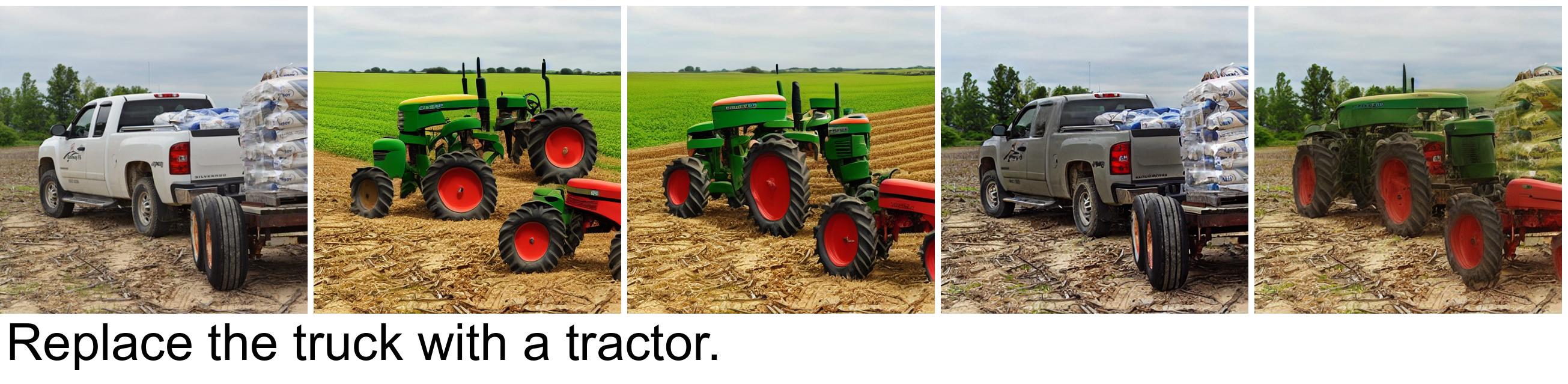}
        \parbox{\linewidth}{
        ``Replace the truck with a tractor''
        \vspace{0.02in}}
    \end{subfigure}
    \end{footnotesize}
    \vspace{-0.25in}

    \caption{Effect of refined instructions and \ourwork{} guidance loss. Compared to variants of our model trained without refined instructions (``Ours (w/o data)'') or without the guidance loss (``Ours (w/o loss)''), our model produce superior results.}
    \label{fig:ablation}
\end{figure}

To assess the impact of refined instructions and our \ourwork{} guidance loss, we trained two model variants: one on the original IP2P dataset with the \ourwork{} guidance loss (``Ours (w/o data)'') and another on our refined dataset without the guidance loss (``Ours (w/o loss)''). We compare these with IP2P and our full method in Table~\ref{tab:ablation}, and visual examples are provided in Fig.~\ref{fig:ablation}. Results show that refined data significantly improve visual alignment with ground truth, evidenced by higher CLIP and DINO scores, while the guidance loss further enhances this alignment. 

\subsection{Limitations}

\begin{figure}
\centering
\resizebox{\linewidth}{!}{ 
\begin{minipage}{\linewidth}

\begin{small}
    \begin{tabbing}
        \hspace{1em}\= \hspace{2.7em} \= \hspace{2.4em} \= \hspace{3.4em} \= \hspace{3.8em}\= \hspace{2.7em} \= \hspace{2.5em} \= \hspace{3.5em} \= \kill
        \> Ori \> IP2P \> MagBr \> Ours \> Ori \> IP2P \> MagBr \> Ours \\
    \end{tabbing}
    \vspace{-0.32in}

    \begin{subfigure}{0.49\linewidth}
        \includegraphics[width=\linewidth, trim=0 37px 10px 32px, clip]{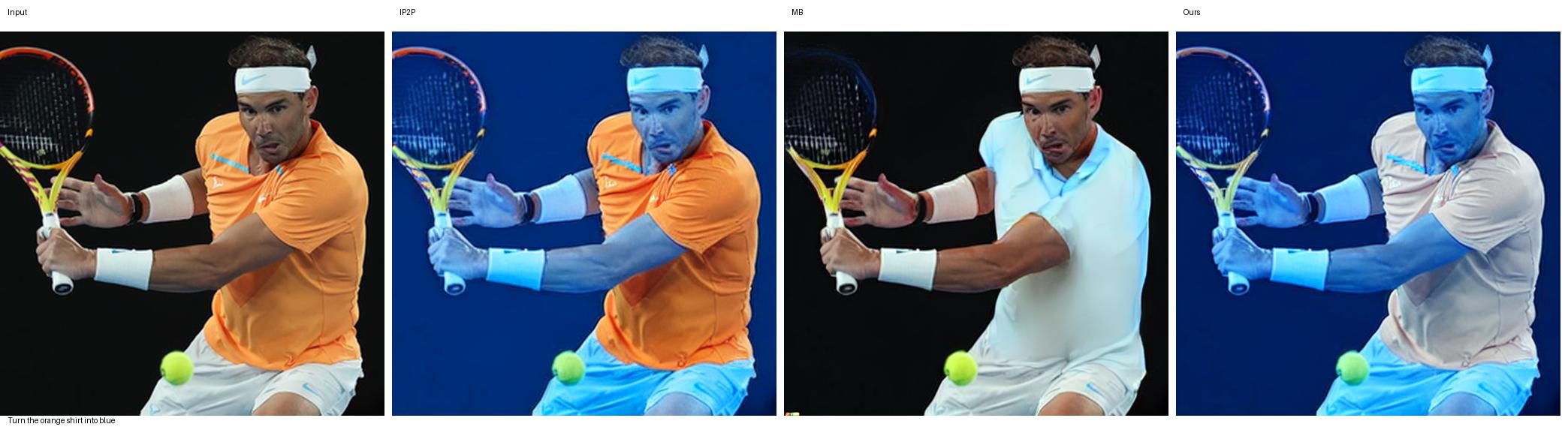}
        \parbox{\linewidth}{\vspace{-0.03in}
        ``Turn orange shirt into blue''
        \vspace{0.02in}}
    \end{subfigure}
    \begin{subfigure}{0.49\linewidth}
        \includegraphics[width=\linewidth, trim=0 37px 10px 32px, clip]{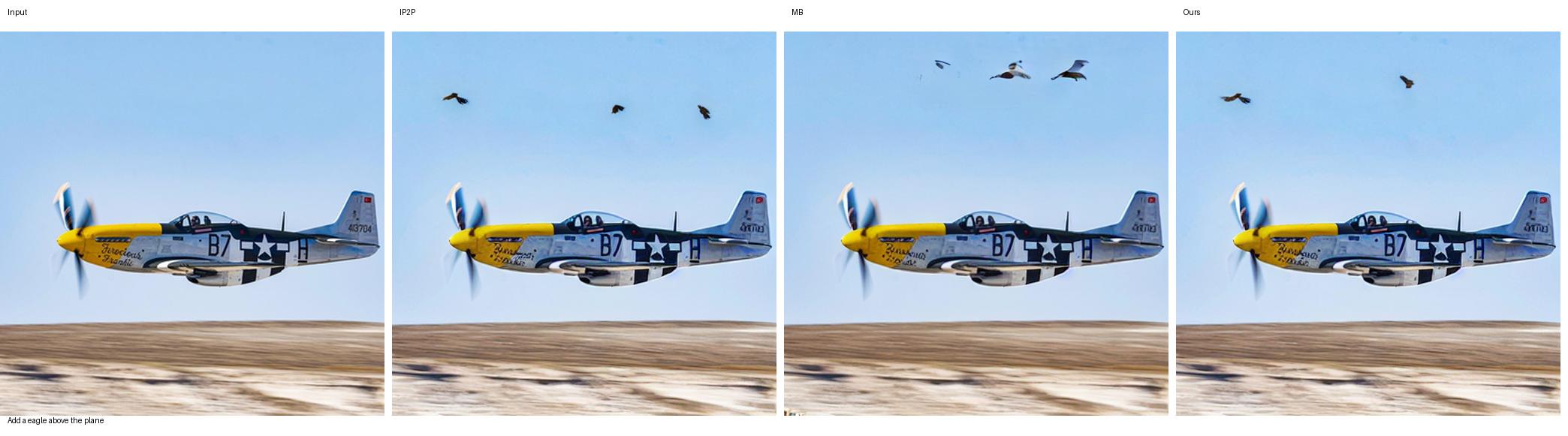}
        \parbox{\linewidth}{\vspace{-0.03in}
        ``Add an eagle above the plane''
        \vspace{0.02in}}
    \end{subfigure}

    \begin{subfigure}{0.49\linewidth}
        \includegraphics[width=\linewidth, trim=0 37px 10px 32px, clip]{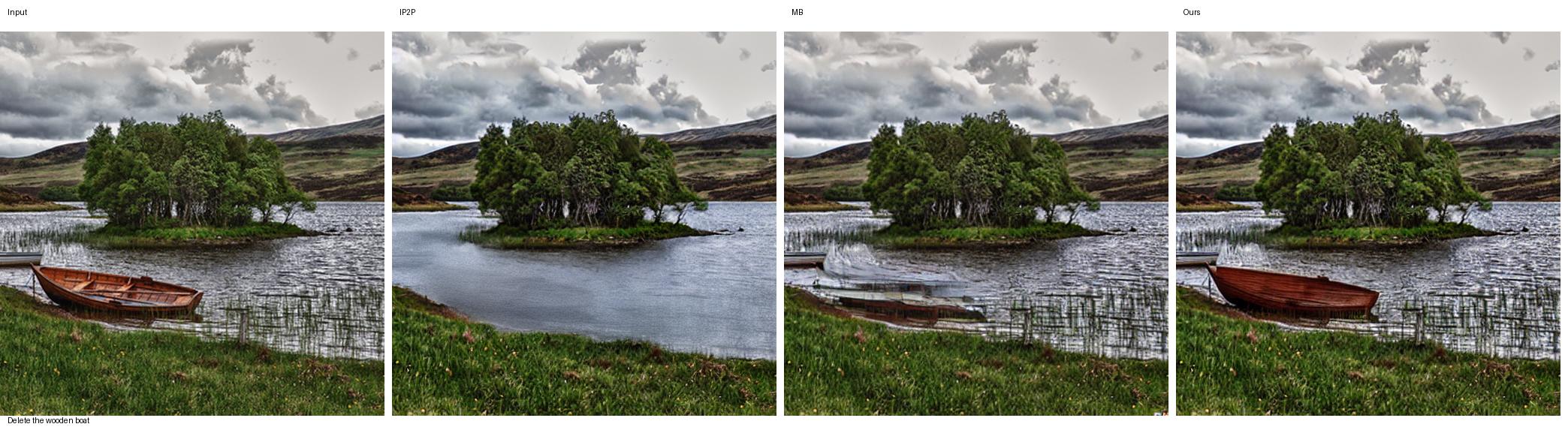}
        \parbox{\linewidth}{\vspace{-0.03in}
        ``Delete the wooden boat''
        \vspace{0.02in}}
    \end{subfigure}
    \begin{subfigure}{0.49\linewidth}
        \includegraphics[width=\linewidth, trim=0 37px 10px 32px, clip]{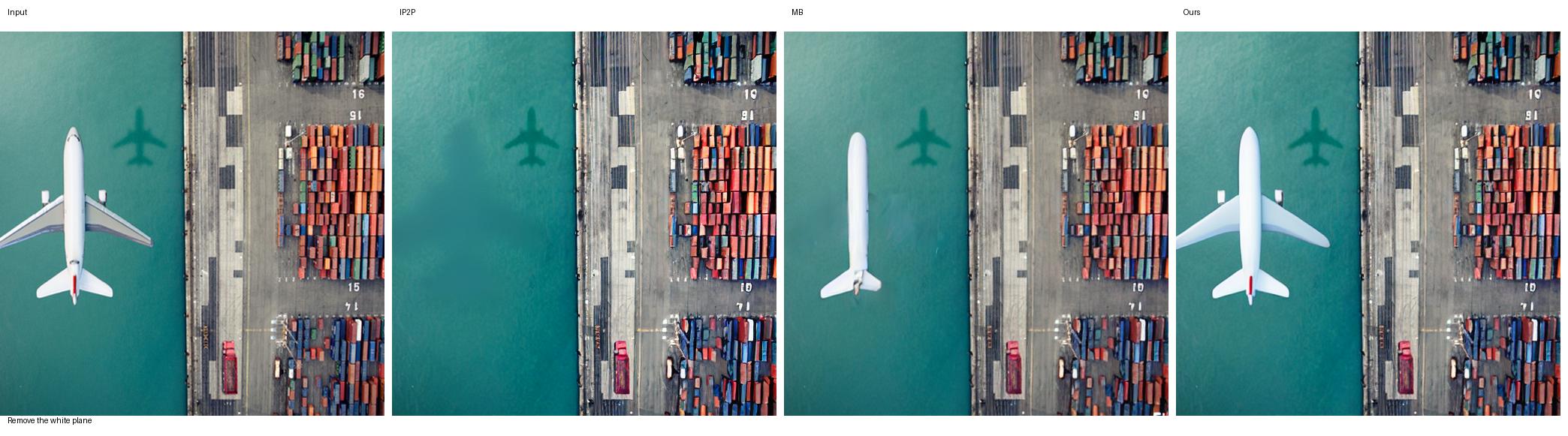}
        \parbox{\linewidth}{\vspace{-0.03in}
        ``Remove the white plane''
        \vspace{0.02in}}
    \end{subfigure}

\end{small}
\end{minipage}
} 
\vspace{-0.1in}
\caption{Examples of failure cases compared with InstructPix2Pix (IP2P)~\cite{brooks2023instructpix2pix} and MagicBrush (MagBr)~\cite{zhang2024magicbrush} performed on the original (Ori) images.}
\label{fig:failure_cases}
\end{figure}

Despite being able to correct a lot of inaccurate edit instructions in the training set, our method is still heavily influenced by the limitations of the generative method~\cite{hertz2022prompt} these training images come from, which include color bleeding outside the intended edit region (Fig.~\ref{fig:failure_cases}, top left), and incomplete object addition (Fig.~\ref{fig:failure_cases}, top right). While our model respects the original images better, it sometimes struggles to remove objects in the original images (Fig.~\ref{fig:failure_cases}, bottom).

\section{Conclusion}

We have presented \ourwork{}, a self-supervised method for instruction-guided image editing that learns the semantic changes between original and edited images to refine edit instructions in datasets. Applying \ourworkabbr{} to the InstructPix2Pix dataset yields over 120K refined samples, which we use to fine-tune its model with our \ourworkabbr{}-guided loss function and generate better edit results.


{
    \small
    \bibliographystyle{ieeenat_fullname}
    \bibliography{main}
}

\clearpage
\noindent\textbf{\Large Appendix}
\vspace{0.1in}
\appendix
\setcounter{table}{3}
\setcounter{figure}{7}

In this supplementary material, we first discuss the differences between our approach and CLIP directional similarity (Sec.~\ref{sec:clip_directional_supplementary}). Next, we provide additional implementation details in Sec.~\ref{sec:implementation_supplementary}. We then compare the performance of edit instruction refinement between our approach and vision-language models in Sec.~\ref{sec:comparison_vlm_supplementary}. Following that, we highlight the limitations of CLIP/DINO metrics in Sec.~\ref{sec:limitation_quantitative_supplementary}. Finally, we present additional editing results, refined editing instructions, and further failure cases in Sec.~\ref{sec:user_study_additional_results_supplementary}.

\section{CLIP Directional Similarity Comparison\label{sec:clip_directional_supplementary}}

One key difference between our \ourworkabbr{} and the CLIP directional similarity~\cite{gal2022stylegan} used in InstructPix2Pix~\cite{brooks2023instructpix2pix} is that \ourworkabbr{} leverages edit instructions rather than individual image prompts, which require two prompts per image pair. This makes \ourworkabbr{} readily applicable across instruction-guided image-editing datasets, even when image prompts are unavailable. Furthermore, individual prompts can often be lengthy and verbose, while edit instructions are typically more concise, reducing irrelevant information in the corresponding text embeddings.

For example, consider a prompt pair from the IP2P dataset: ``Infinity walk by Marcelo Archila - Black \& White Landscapes (contrast, monochrome, hdr, black and white, fine art, long exposure)'' and ``Infinity walk by Marcelo Archila - Black \& White Landscapes (contrast, monochrome, hdr, black and white, commercial).'' At first glance, it may be challenging to infer the edit instruction, which in this case is simply: ``make it commercial.''

\section{Implementation and Dataset Refinement\label{sec:implementation_supplementary}}

\begin{table}[b]
\centering
\includegraphics[width=\linewidth]{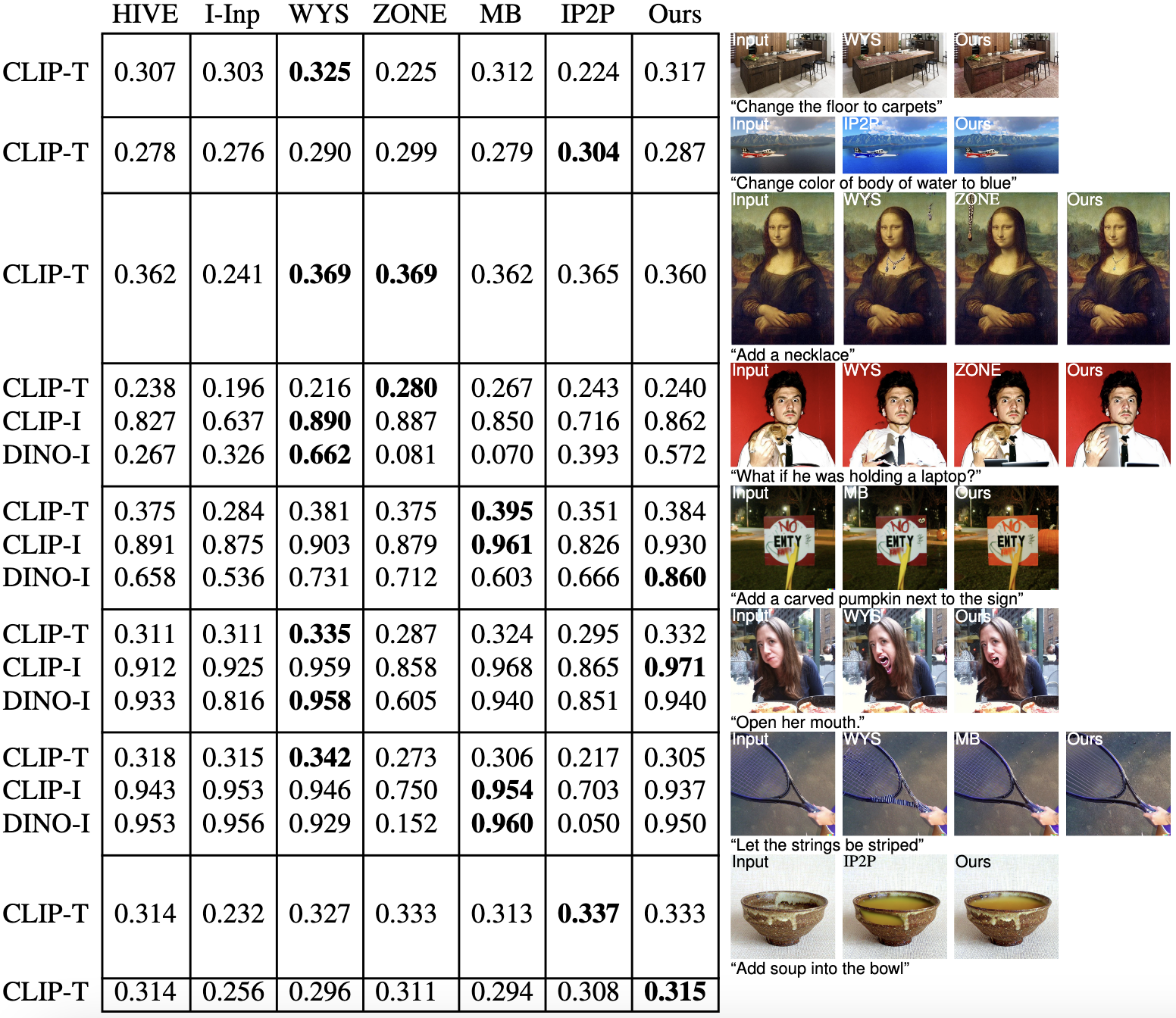}
\caption{Metrics (CLIP-I/DINO-I if GT exists) for Fig.\ 5 outputs, with best bolded and shown.}
\label{tab:sample_metrics}
\end{table}

\begin{figure*}[]
\centering
\includegraphics[width=0.85\linewidth]{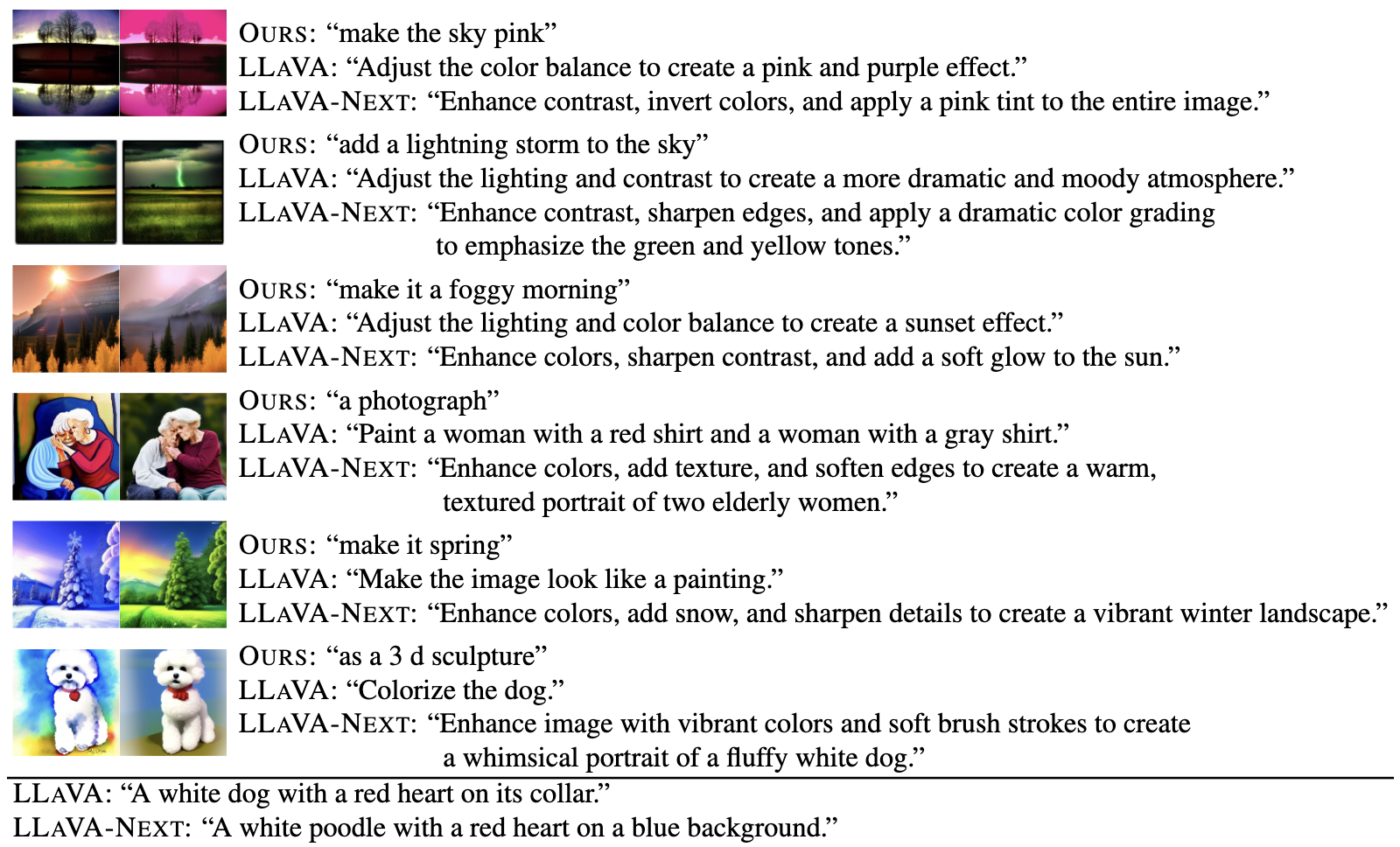}
\caption{\textbf{Top:} VLM outputs with respect to prompt ``Provide the edit instruction that can transform the source image to the target image in one phrase:'' and image pairs in Fig.~2. \textbf{Bottom:} VLM outputs with respect to prompt ``Describe the image in one phrase:'' and the input (leftmost) image in the last row.}
\label{fig:vlm}
\end{figure*}

\begin{figure}
    \centering
    \includegraphics[width=0.8\linewidth]{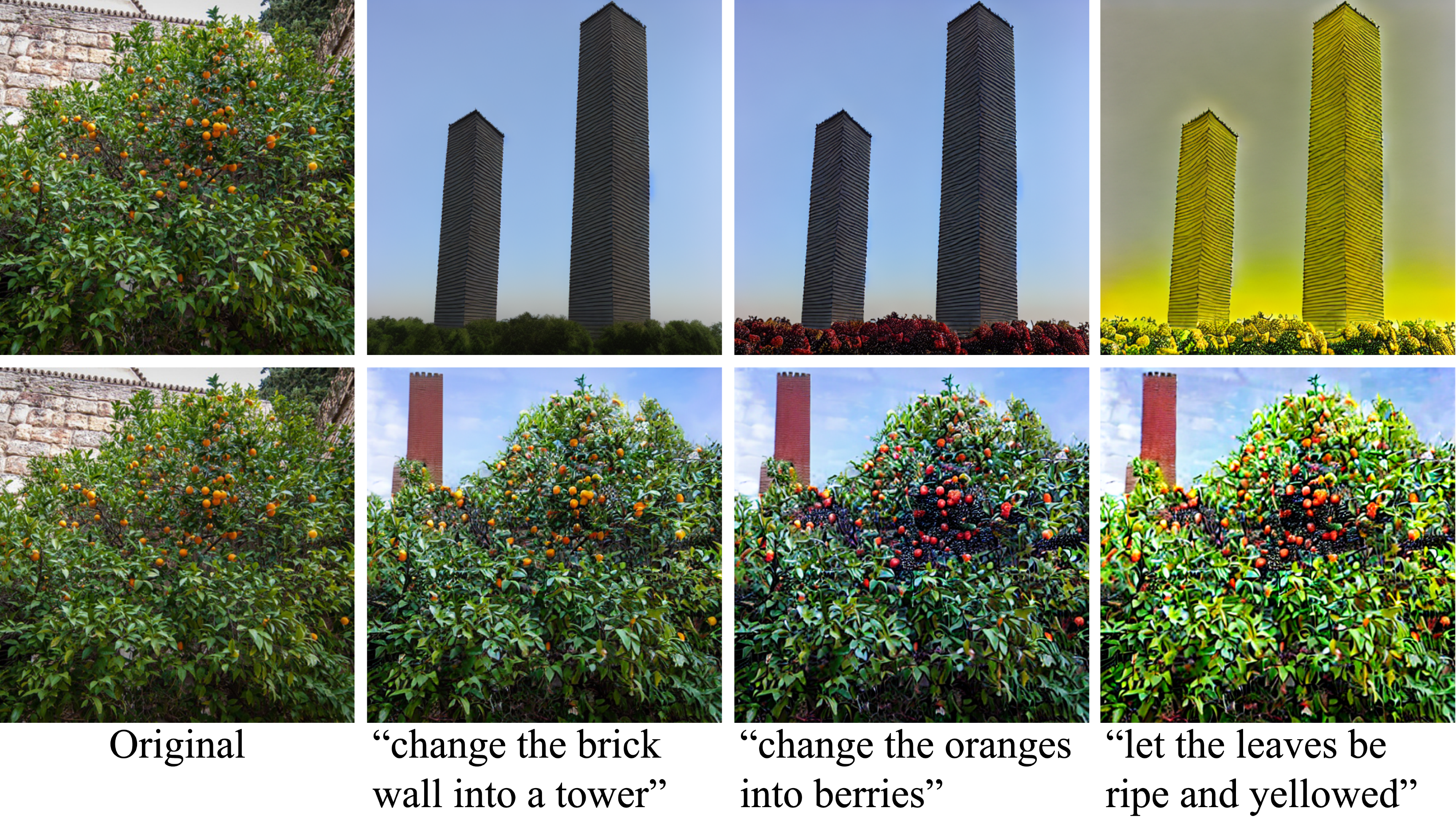}
    \vspace{-0.1in}
    \caption{Multi-turn edit comparison with InstructPix2Pix (IP2P)~\cite{brooks2023instructpix2pix}. Note how IP2P (top row) gradually diverges from the desired result more and more, unlike our approach (bottom row) which produces results more consistent with the original.}
    \label{fig:multiturn}
\end{figure}

LD-DINOv2 is initialized from a ViT-L/14 DINOv2 model~\cite{oquab2023dinov2}. To accommodate Stable Diffusion (SD) VAE~\cite{rombach2022high} encoded images, the patch embedding projection layer is replaced. Additionally, the timestep embedding projection module is initialized to handle timestep inputs following the SD timestep encoding implementation.

The model is trained on the InstructPix2Pix (IP2P)~\cite{brooks2023instructpix2pix} dataset, with all images resized to $256 \times 256$. Training is conducted with a learning rate of $10^{-5}$, a batch size of 32, and a total of 100K training steps.

During the first 10K steps, the timesteps are fixed to 0. This ensures that latent image inputs are not noisified, which allows the patch embedding projection layer to learn to encode latent images effectively. Then, the upper bound of the timestep is linearly increased in proportion to the training step number, reaching the maximum value of 1000 at the $90K^{\text{th}}$ step. During this period, the timestep value is uniformly sampled between 0 and the current upper bound for each training step. This gradual increase in the timestep value sampling range helps preserve the knowledge learned by the patch embedding projection layer while simultaneously adapting the timestep embedding projection module.

For the last 10K steps, timesteps are randomly sampled across the entire range. This strategy ensures that the model learns to handle the full distribution of timestep values.

\ourwork{} (\ourworkabbr{} in short) is initialized from a ViT-L/14 CLIP model~\cite{radford2021learning}. We freeze the text decoder, the aforementioned LD-DINOv2, and finetune the CLIP image encoder. The instruction decoder follows the architecture of DeCap~\cite{li2023decap} with a pre-trained GPT-2 backbone~\cite{radford2019language} and is trained along with the image encoder on the IP2P dataset with a learning rate of $10^{-5}$, a batch size of 32, and a total of 100K training steps. 

The advantage of training LD-DINOv2 ahead of time is that we can sample timesteps randomly within its maximum possible range without repeating the above training procedure, as LD-DINOv2 has already learned to ``ignore'' the noise added to the latent image.

After training, we refine the IP2P dataset. For each data sample $(I^o, I_e, p)$ and its corresponding refined instruction $p'$, we update the sample if the \ourworkabbr{} cosine similarity between the visual changes in the original/edited image and the refined instruction differs significantly from that with the original instruction:
\begin{equation}
\small
\begin{aligned}
&\text{sim}\big(\text{\ourwork{}}_\text{vis}(L^o, 0, L^e, 0), \text{\ourwork{}}_\text{text}(p^R)\big) \\
&> \text{sim}\big(\text{\ourwork{}}_\text{vis}(L^o, 0, L^e, 0), \text{\ourwork{}}_\text{text}(p)\big) + \phi,
\end{aligned}
\end{equation}
where
\begin{equation}
\begin{split}
L^o &= \text{VAE}_\text{enc}(I^o), \\
L^e &= \text{VAE}_\text{enc}(I^e),
\end{split}
\end{equation}
and $\phi = 0.1$ is the margin. This results in over 120K new instructions out of 313,010 samples in the IP2P dataset. We retain the original instructions for the remaining samples.

The image editing model is initialized from the IP2P model~\cite{brooks2023instructpix2pix}, where the UNet~\cite{ronneberger2015u} is fine-tuned using Low-Rank Adaptation (LoRA)~\cite{hu2021lora} with parameters $r = \alpha = 32$ on the newly generated samples. The training is performed with a learning rate of $10^{-4}$, a batch size of 64, and a total of 10K training steps. The rest of the training configuration follows the original IP2P work.

\section{Limitation of CLIP/DINO Metrics\label{sec:limitation_quantitative_supplementary}}

There are several reasons for the gap between our qualitative and quantitative results, which we include for completeness. First, while these metrics are widely used, they have well-documented limitations and do not align with human judgment, as highlighted by VIEScore~\cite{ku2023viescore}. Specifically, Yuksekgonul et al.~\cite{yuksekgonul2022and} show that CLIP’s Bag-of-Words behavior is insensitive to word order, leading to weak correlations with human evaluations.

This issue is also evident in Table~\ref{tab:sample_metrics}; despite the superior qualitative performance of our results in Fig.~5, our metrics are usually lower than the baselines. Additionally, the results presented in Table 1 (MagBr data) are computed on the MagBr test set, which has a distribution similar to their training set, giving MagBr an inherent advantage.

\section{Comparison with Vision Language Models\label{sec:comparison_vlm_supplementary}}

To compare our method with vision-language models (VLMs) in terms of edit instruction refinement, we evaluate LLaVA~\cite{liu2023llava} and LLaVA-Next~\cite{liu2024llavanext}, two widely used open-source VLMs, as shown in Fig.~\ref{fig:vlm} (top). Both VLMs fail to generate effective edit instructions compared to our method.

In Fig.~\ref{fig:vlm} (bottom), we use these VLMs to generate a caption for the input image (last row), which serves as the input prompt for editing methods that require separate prompts for the input and target images. While the caption accurately describes the image, it fails to capture its watercolor style—a crucial detail needed for the intended style editing in this sample pair. Consequently, users still need to manually refine the input prompt and compose the target image prompt, which is significantly more cumbersome than using a single edit instruction.

\section{Additional Results\label{sec:user_study_additional_results_supplementary}} 

\begin{figure*}
    \centering
    \begin{tabbing}
        \hspace{4.4em}\= \hspace{7.8em} \= \hspace{5.6em} \= \hspace{8.4em}\= \hspace{8.1em} \= \hspace{5.6em} \= \kill
        \> Original \> IP2P \> \ourworkabbr{} (Ours) \> Original \> IP2P \> \ourworkabbr{} (Ours)  \\
    \end{tabbing}    
    \vspace{-0.3in}

    \begin{subfigure}{0.45\linewidth}
        \includegraphics[width=\linewidth, trim=0 80px 0 0, clip]{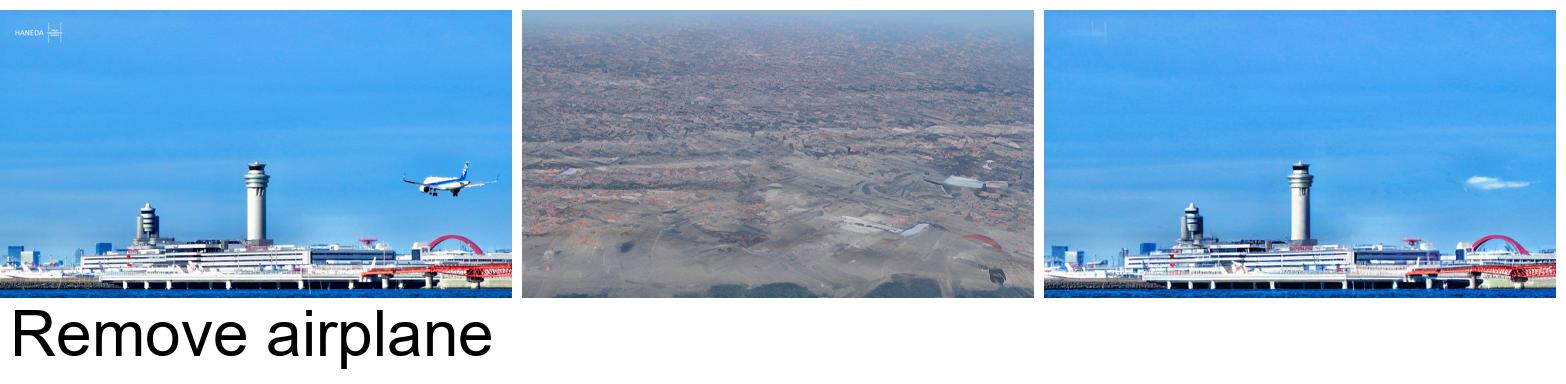}
        \parbox{\linewidth}{\vspace{-0.02in}
        ``Remove airplane''
        \vspace{0.05in}}
    \end{subfigure}
    \begin{subfigure}{0.45\linewidth}
        \includegraphics[width=\linewidth, trim=0 76px 0 0, clip]{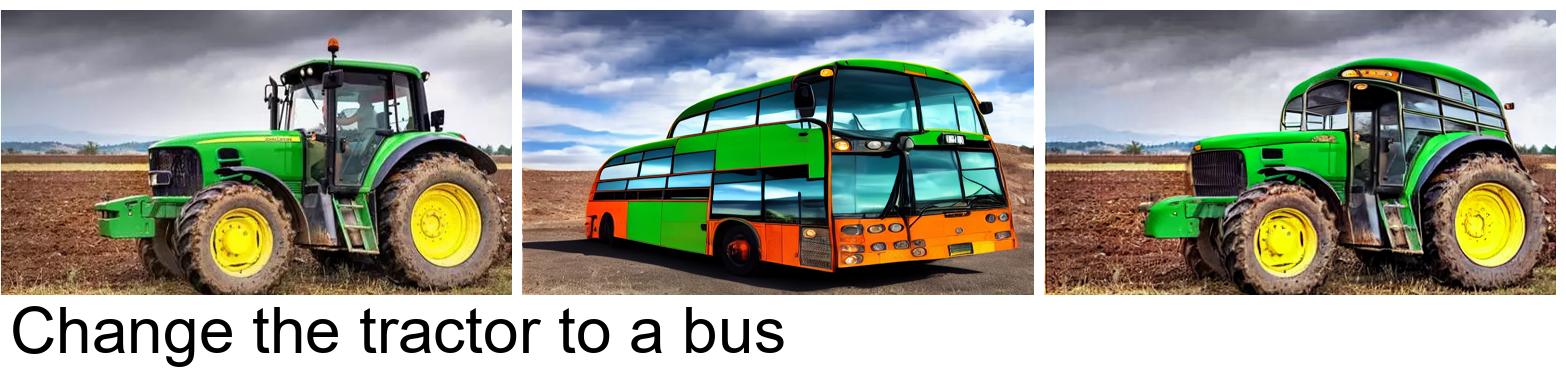}
        \parbox{\linewidth}{\vspace{-0.02in}
        ``Change the tractor to a bus''
        \vspace{0.05in}}
    \end{subfigure}

    \begin{subfigure}{0.45\linewidth}
        \includegraphics[width=\linewidth, trim=0 135px 0 0, clip]{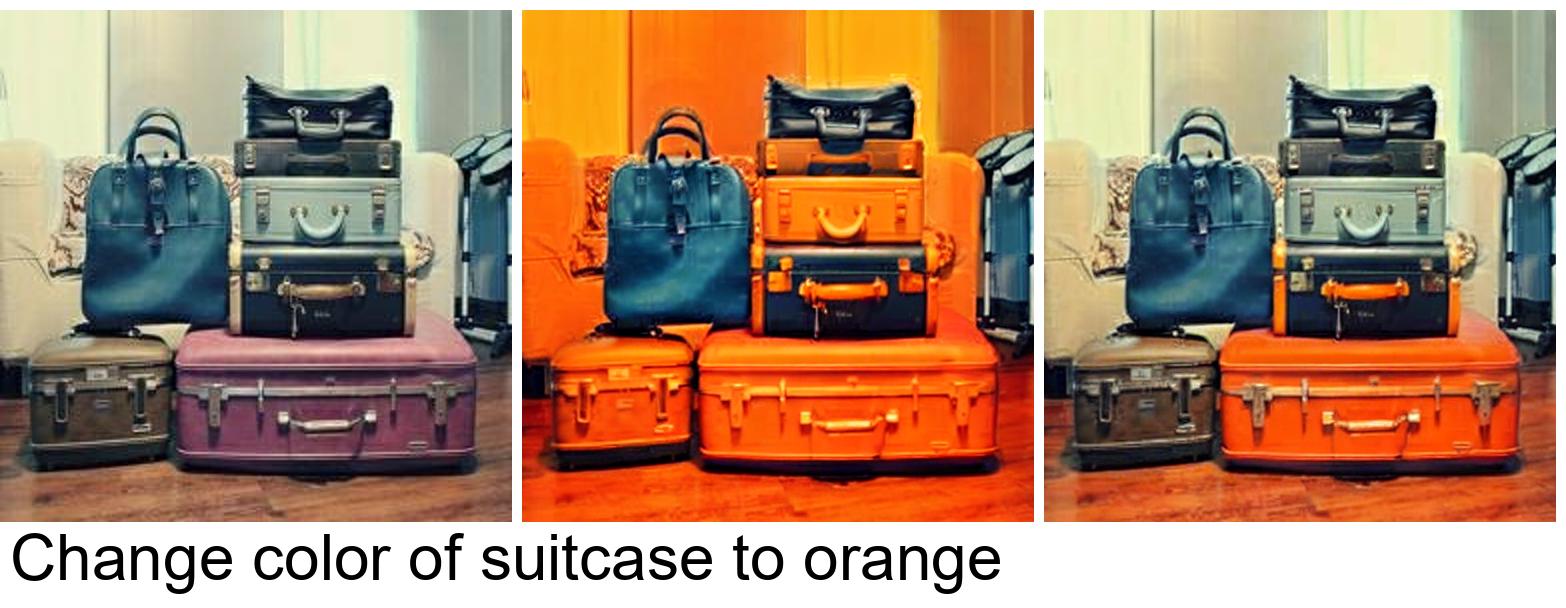}
        \parbox{\linewidth}{\vspace{-0.02in}
        ``Change color of suitcase to orange''
        \vspace{0.05in}}
    \end{subfigure}
    \begin{subfigure}{0.45\linewidth}
        \includegraphics[width=\linewidth, trim=0 135px 0 0, clip]{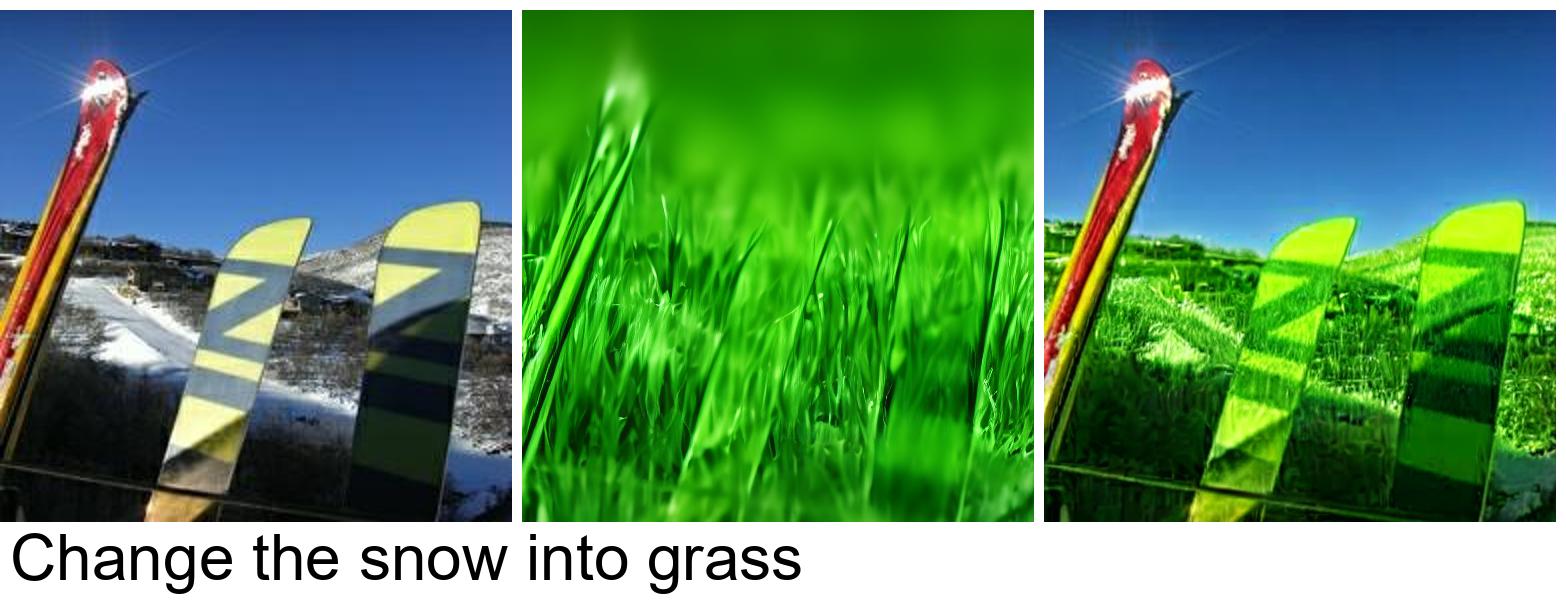}
        \parbox{\linewidth}{\vspace{-0.02in}
        ``Change the snow into grasses''
        \vspace{0.05in}}
    \end{subfigure}

    \begin{subfigure}{0.45\linewidth}
        \includegraphics[width=\linewidth, trim=0 135px 0 0, clip]{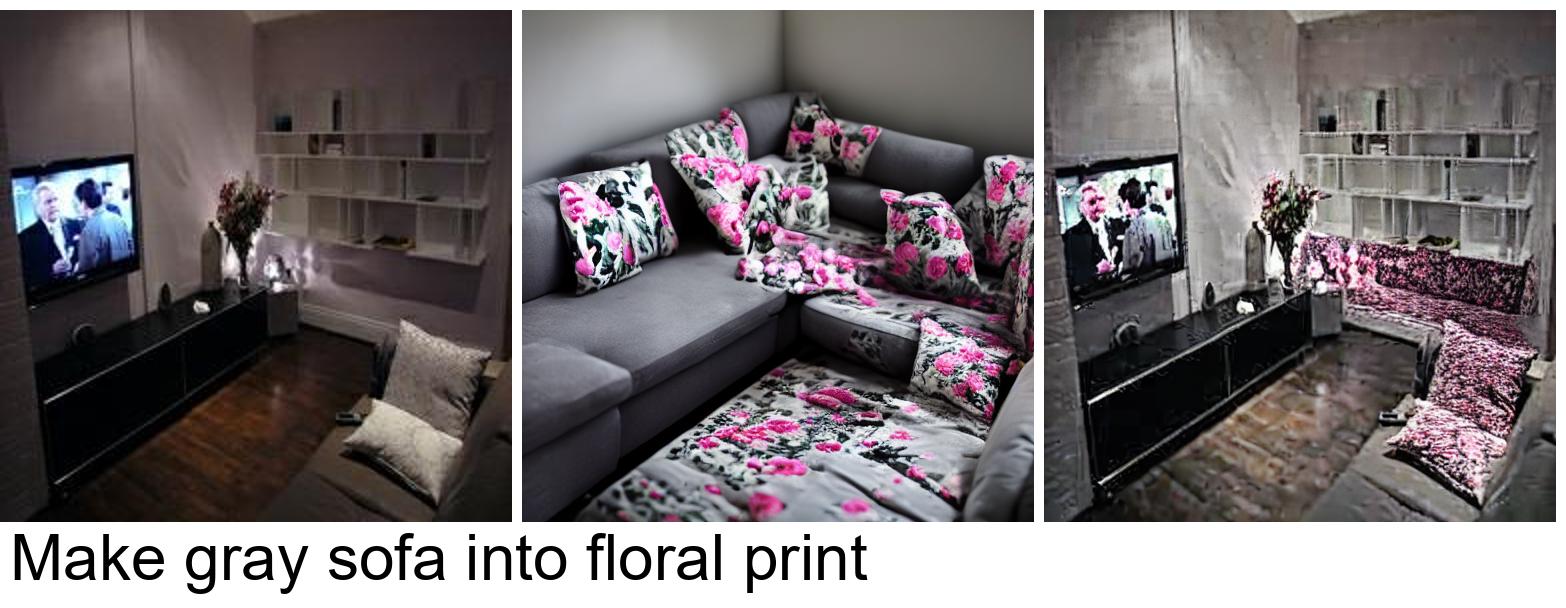}
        \parbox{\linewidth}{\vspace{-0.02in}
        ``Make gray sofa into floral print''
        \vspace{0.05in}}
    \end{subfigure}
    \begin{subfigure}{0.45\linewidth}
        \includegraphics[width=\linewidth, trim=0 135px 0 0, clip]{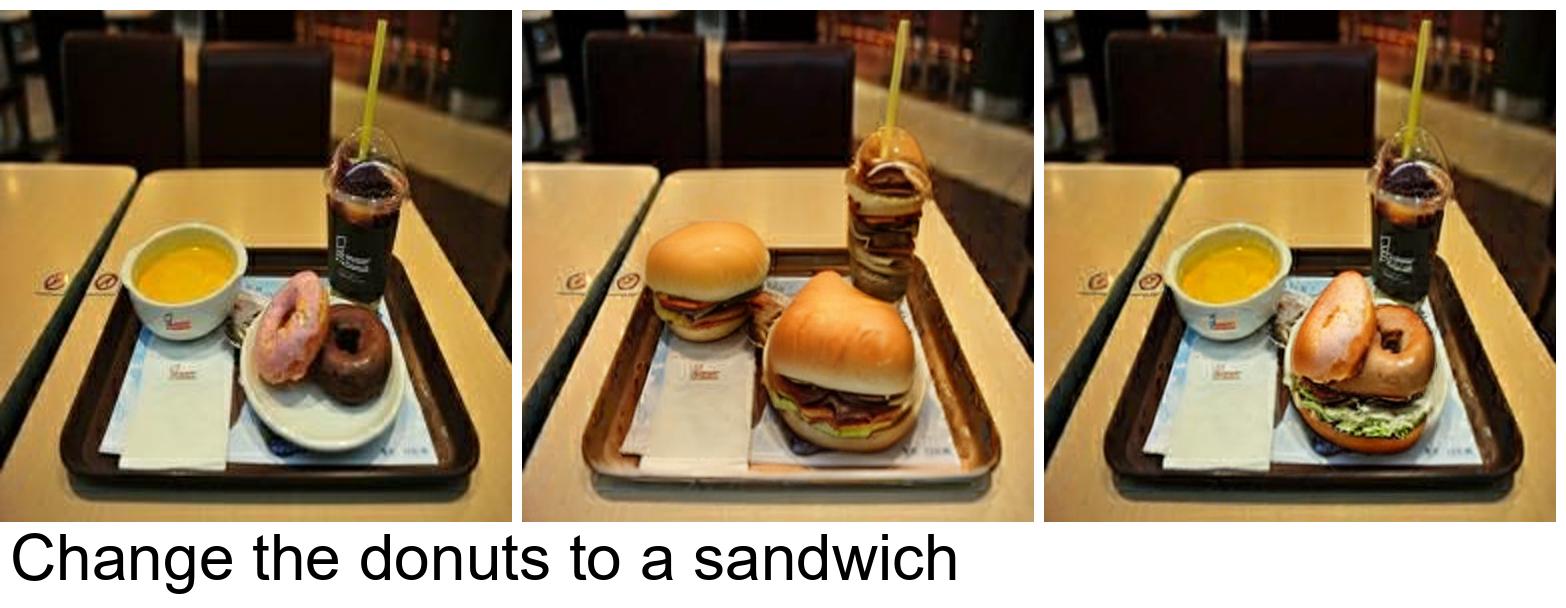}
        \parbox{\linewidth}{\vspace{-0.02in}
        ``Change the donuts to a sandwich''
        \vspace{0.05in}}
    \end{subfigure}

    \begin{subfigure}{0.45\linewidth}
        \includegraphics[width=\linewidth, trim=0 135px 0 0, clip]{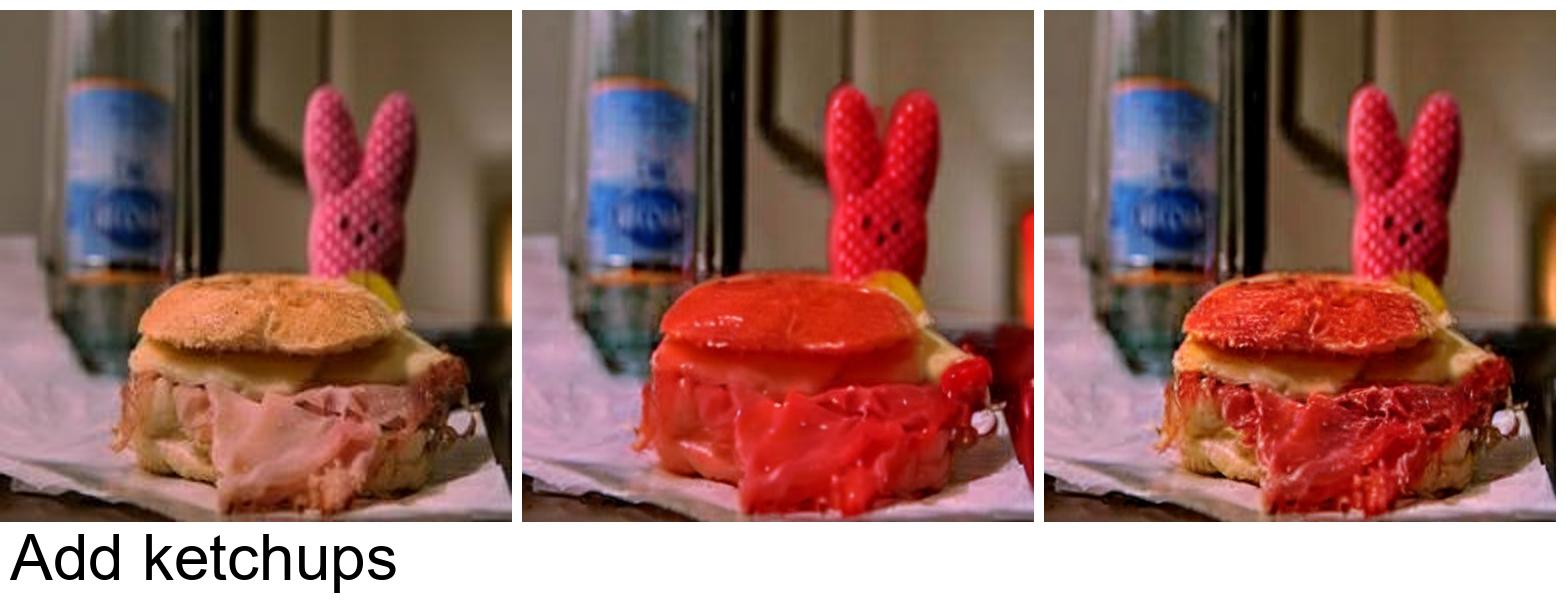}
        \parbox{\linewidth}{\vspace{-0.02in}
        ``Add ketchups''
        \vspace{0.05in}}
    \end{subfigure}
    \begin{subfigure}{0.45\linewidth}
        \includegraphics[width=\linewidth, trim=0 135px 0 0, clip]{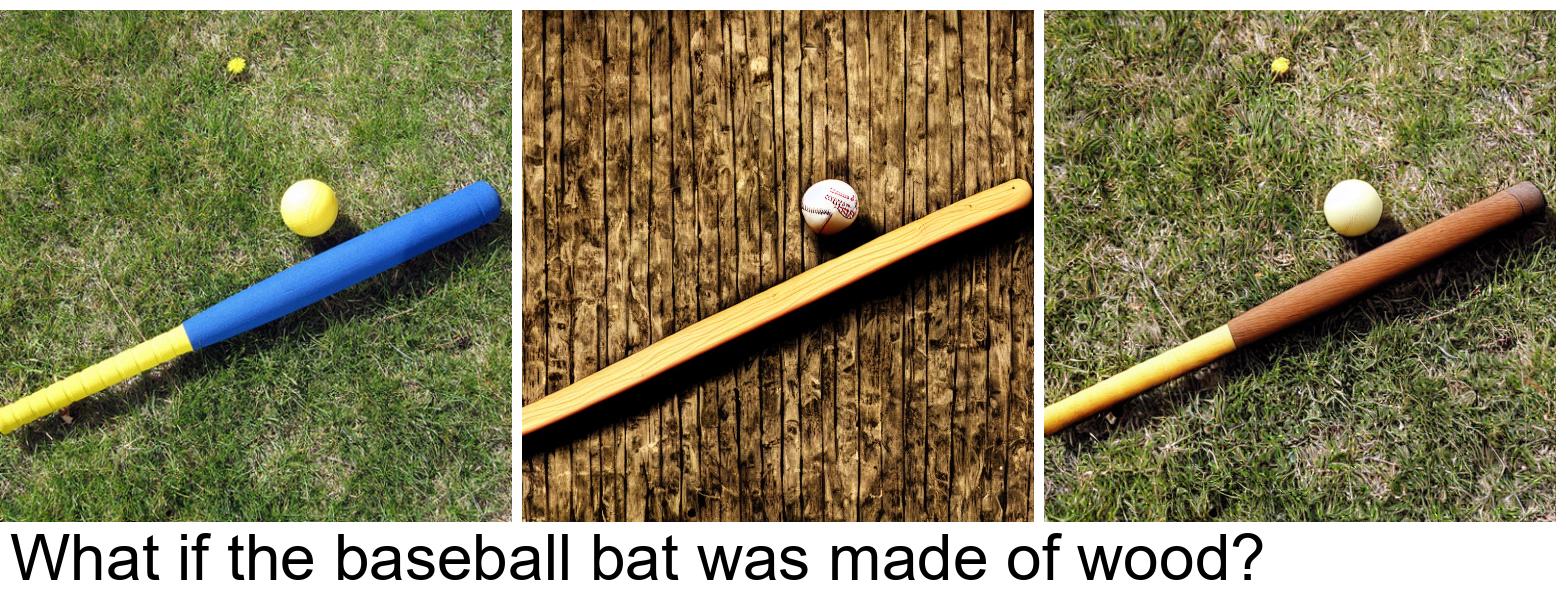}
        \parbox{\linewidth}{\vspace{-0.02in}
        ``What if the baseball bat was made of wood?''
        \vspace{0.05in}}
    \end{subfigure}

    \begin{subfigure}{0.45\linewidth}
        \includegraphics[width=\linewidth, trim=0 75px 0 0, clip]{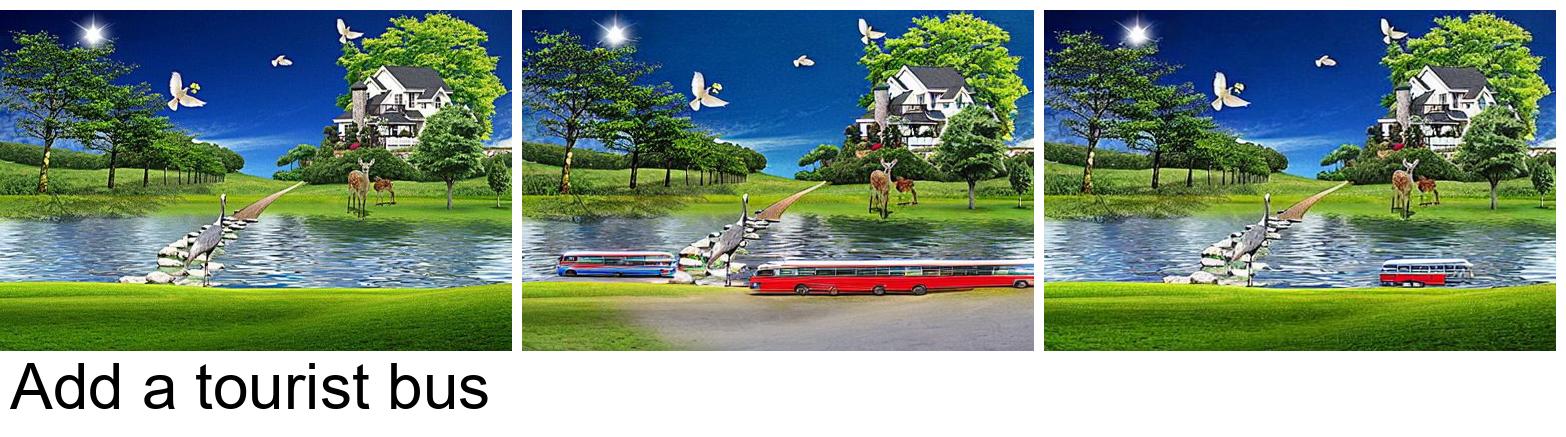}
        \parbox{\linewidth}{\vspace{-0.02in}
        ``Add a tourist bus''
        \vspace{0.05in}}
    \end{subfigure}
    \begin{subfigure}{0.45\linewidth}
        \includegraphics[width=\linewidth, trim=0 75px 0 0, clip]{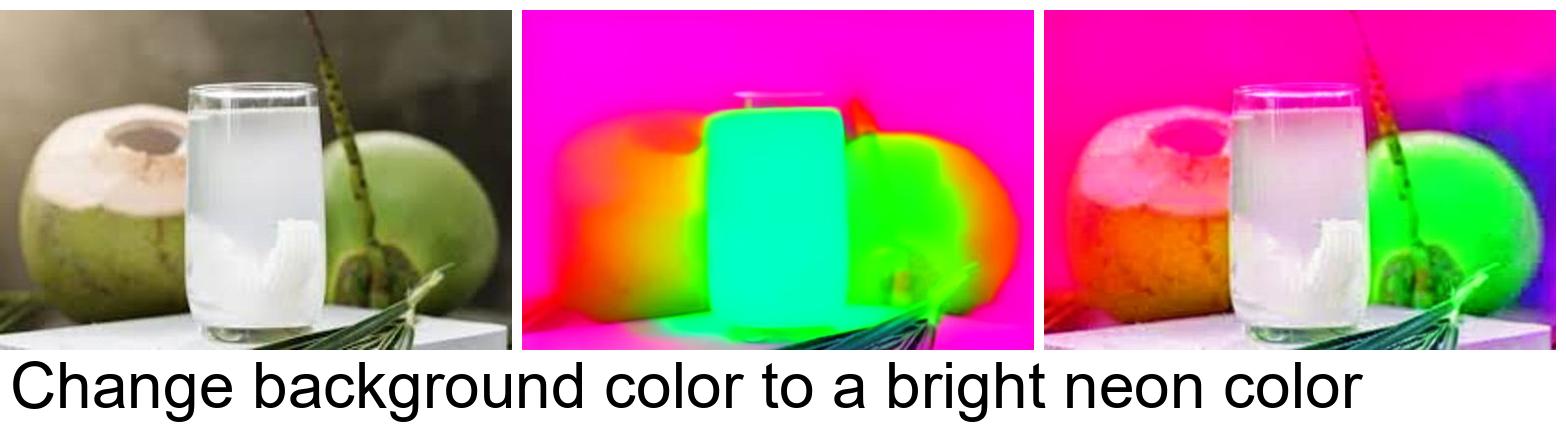}
        \parbox{\linewidth}{\vspace{-0.02in}
        ``Change background color to a bright neon color''
        \vspace{0.05in}}
    \end{subfigure}

    \begin{subfigure}{0.45\linewidth}
        \includegraphics[width=\linewidth, trim=0 75px 0 0, clip]{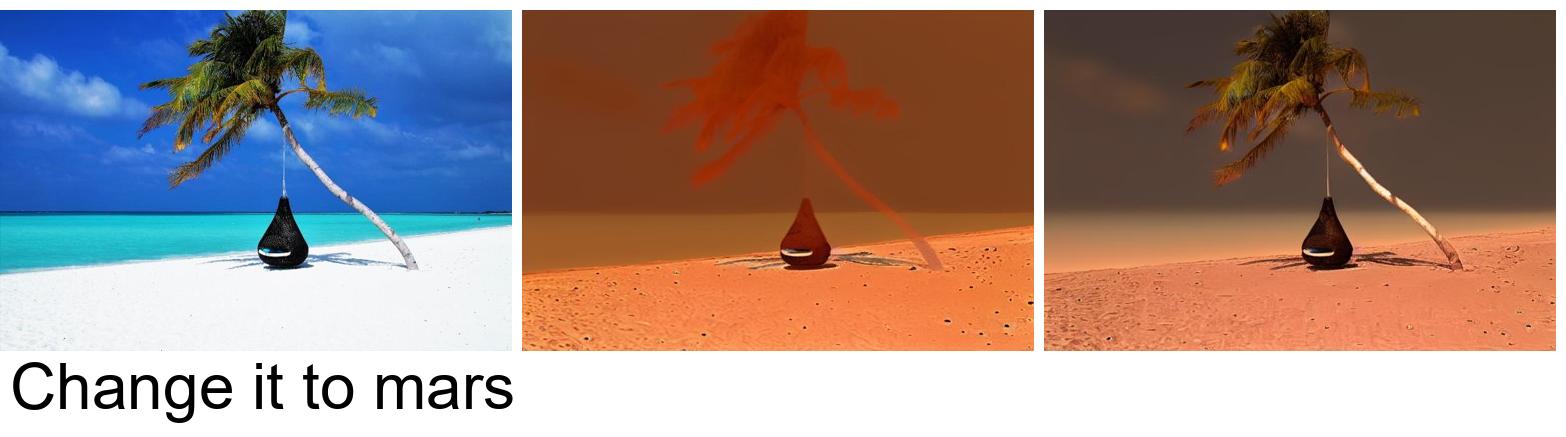}
        \parbox{\linewidth}{\vspace{-0.02in}
        ``Change it to mars''
        \vspace{0.05in}}
    \end{subfigure}
    \begin{subfigure}{0.45\linewidth}
        \includegraphics[width=\linewidth, trim=0 75px 0 0, clip]{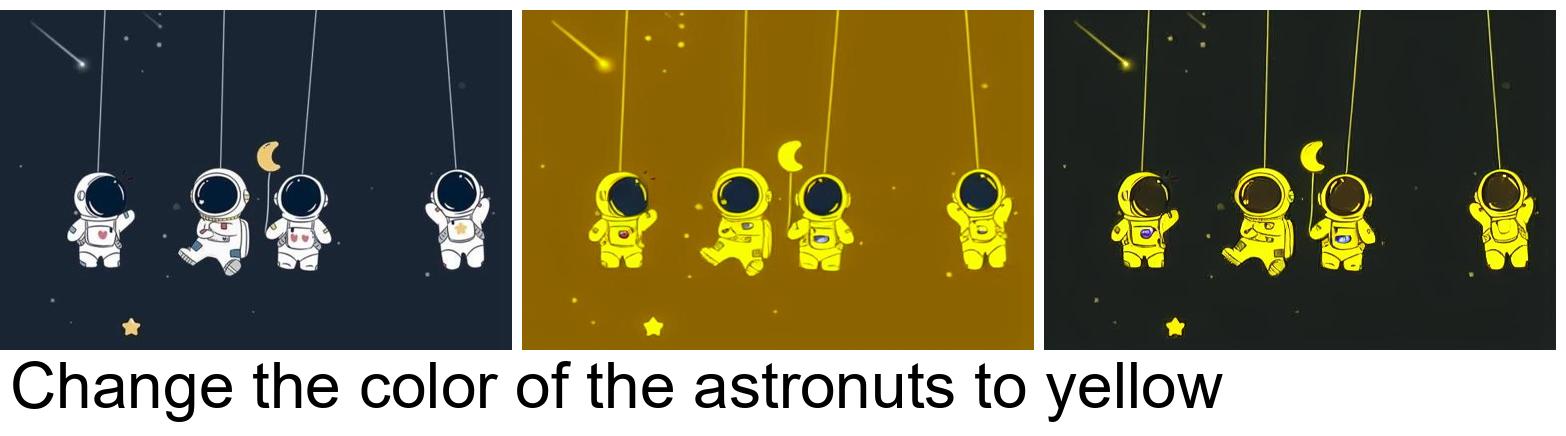}
        \parbox{\linewidth}{\vspace{-0.02in}
        ``Change the color of the astronuts to yellow''
        \vspace{0.05in}}
    \end{subfigure}

    \begin{subfigure}{0.45\linewidth}
        \includegraphics[width=\linewidth, trim=0 100px 0 0, clip]{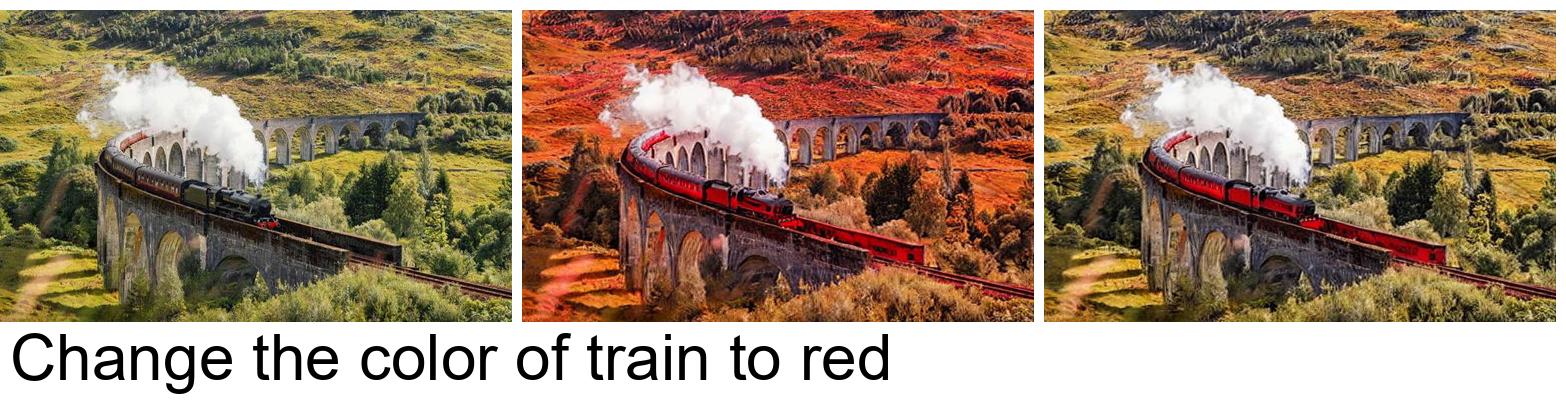}
        \parbox{\linewidth}{\vspace{-0.02in}
        ``Change the color of train to red''
        \vspace{0.05in}}
    \end{subfigure}
    \begin{subfigure}{0.45\linewidth}
        \includegraphics[width=\linewidth, trim=0 75px 0 0, clip]{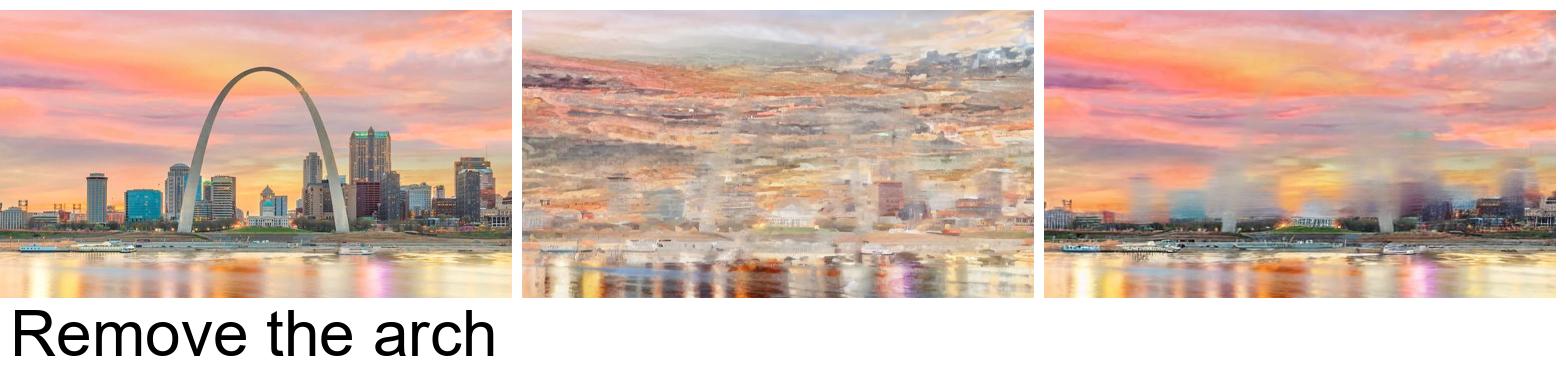}
        \parbox{\linewidth}{\vspace{-0.02in}
        ``Remove the arch''
        \vspace{0.05in}}
    \end{subfigure}

    \begin{subfigure}{0.45\linewidth}
        \includegraphics[width=\linewidth, trim=0 75px 0 0, clip]{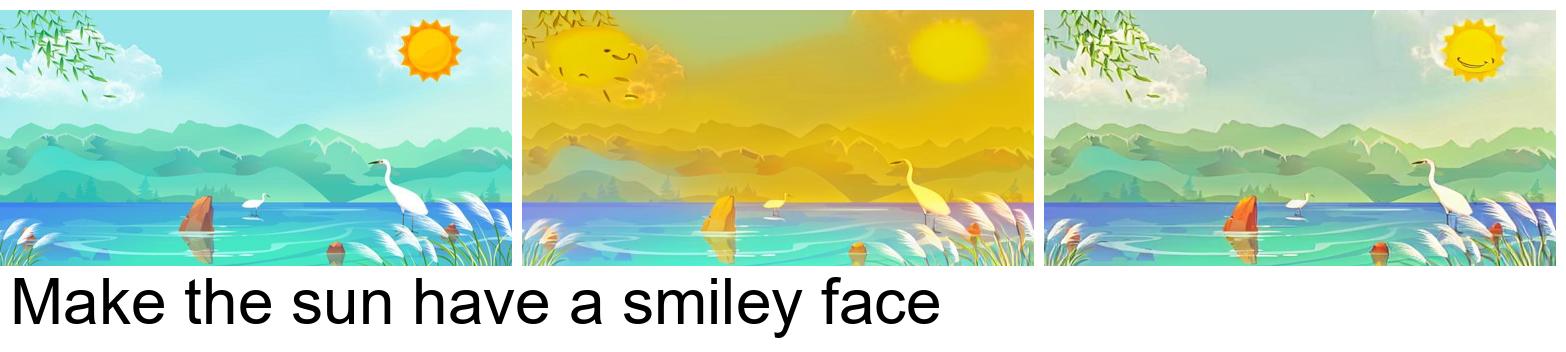}
        \parbox{\linewidth}{\vspace{-0.02in}
        ``Make the sun have a smiley face''
        \vspace{0.05in}}
    \end{subfigure}
    \begin{subfigure}{0.45\linewidth}
        \includegraphics[width=\linewidth, trim=0 230px 0 0, clip]{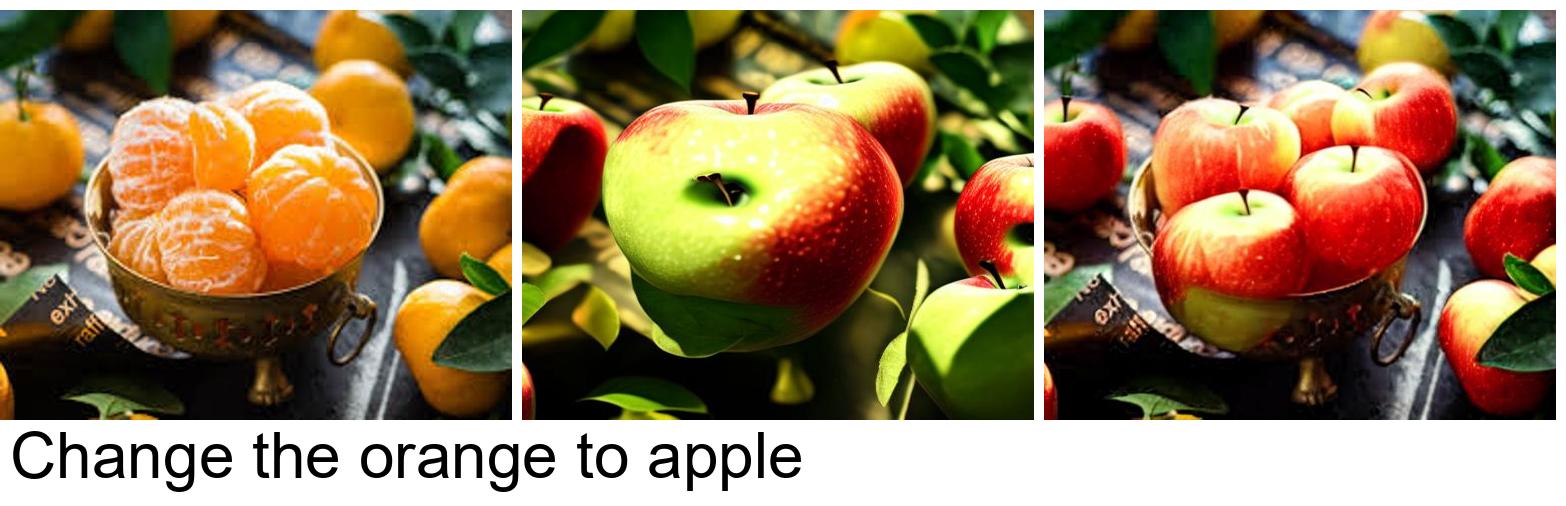}
        \parbox{\linewidth}{\vspace{-0.02in}
        ``Change the orange to apple''
        \vspace{0.05in}}
    \end{subfigure}
    
    \caption{Additional results from our \ourwork{} image editing method on benchmarks~\cite{zhang2024magicbrush,zhang2023hive,zhao2024instructbrush,li2024zone} (Part 1/2)}
    \label{fig:additional_editing_results_1}
    \vspace{0.2in}
\end{figure*}

\begin{figure*}
    \centering
    \begin{tabbing}
        \hspace{4.4em}\= \hspace{7.8em} \= \hspace{5.6em} \= \hspace{8.4em}\= \hspace{8.1em} \= \hspace{5.6em} \= \kill
        \> Original \> IP2P \> \ourworkabbr{} (Ours) \> Original \> IP2P \> \ourworkabbr{} (Ours)  \\
    \end{tabbing}    
    \vspace{-0.3in}

    \begin{subfigure}{0.45\linewidth}
        \includegraphics[width=\linewidth, trim=0 75px 0 0, clip]{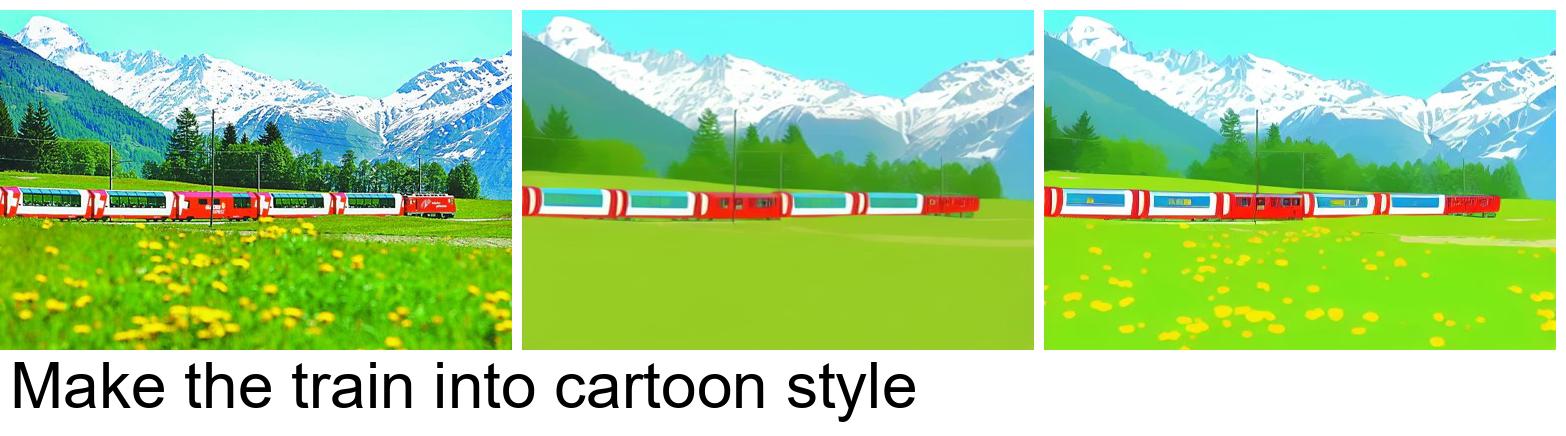}
        \parbox{\linewidth}{\vspace{-0.02in}
        ``Make the train into cartoon style''
        \vspace{0.05in}}
    \end{subfigure}
    \begin{subfigure}{0.45\linewidth}
        \includegraphics[width=\linewidth, trim=0 75px 0 0, clip]{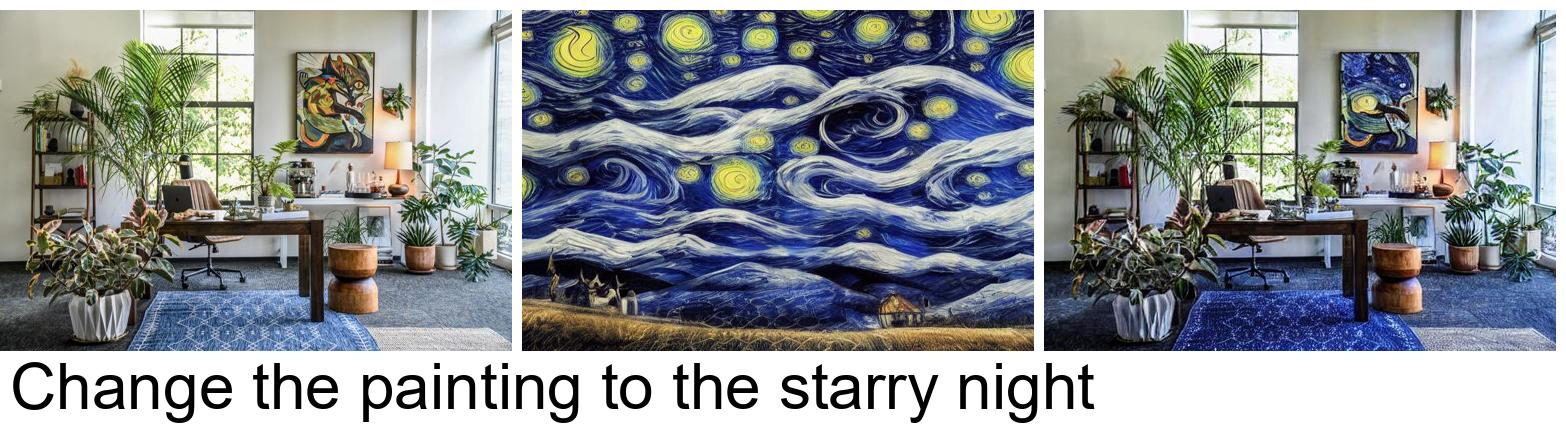}
        \parbox{\linewidth}{\vspace{-0.02in}
        ``Change the painting into the starry night''
        \vspace{0.05in}}
    \end{subfigure}

    \begin{subfigure}{0.45\linewidth}
        \includegraphics[width=\linewidth, trim=0 77px 0 0, clip]{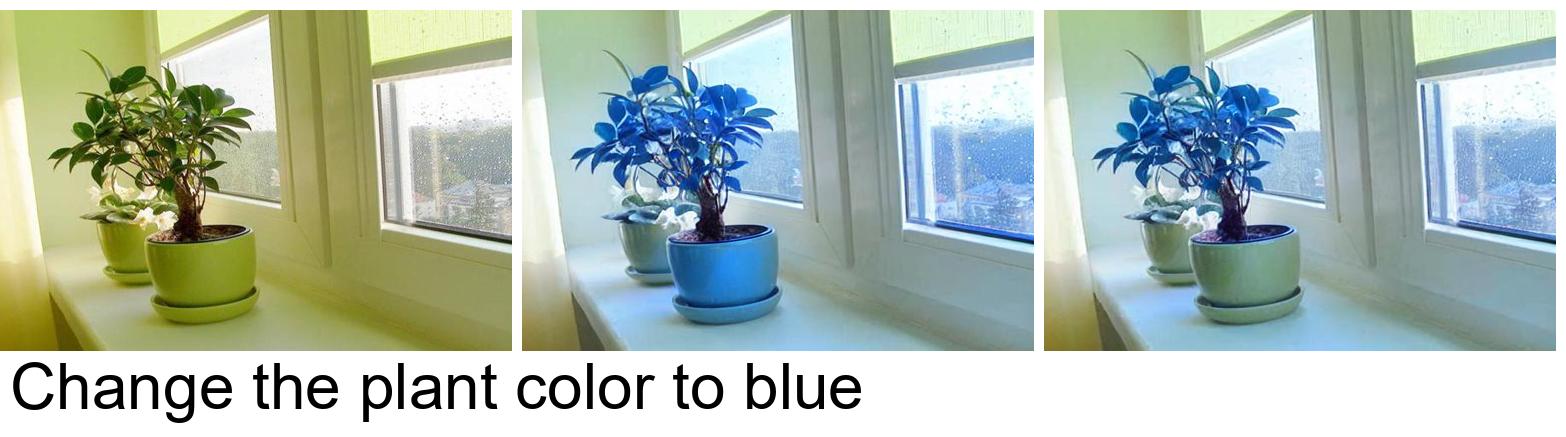}
        \parbox{\linewidth}{\vspace{-0.02in}
        ``Change the plant color to blue''
        \vspace{0.05in}}
    \end{subfigure}
    \begin{subfigure}{0.45\linewidth}
        \includegraphics[width=\linewidth, trim=0 77px 0 0, clip]{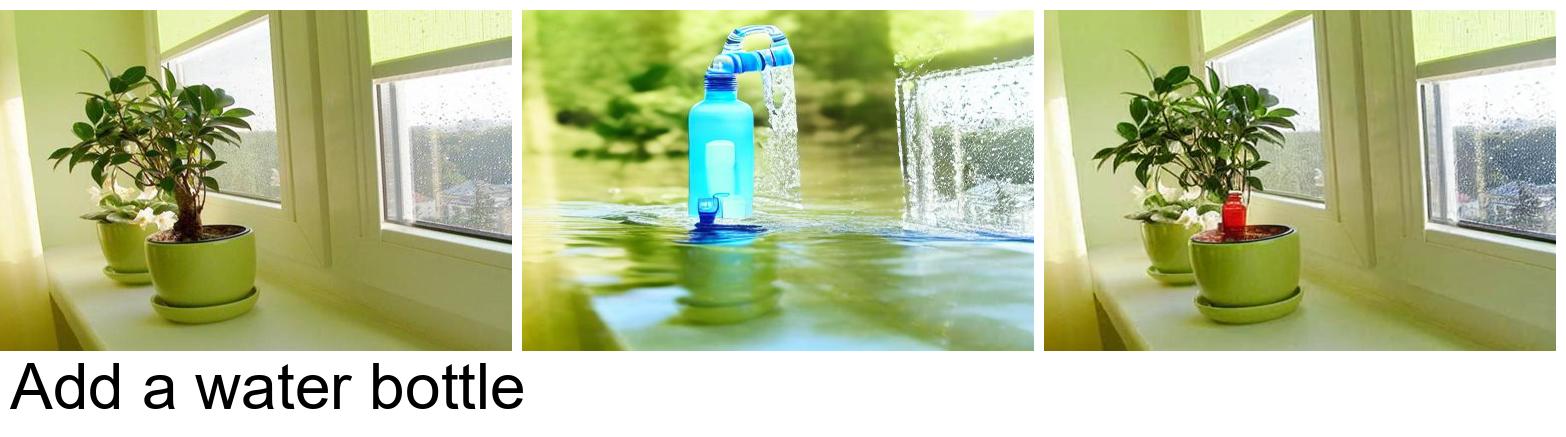}
        \parbox{\linewidth}{\vspace{-0.02in}
        ``Add a water bottle''
        \vspace{0.05in}}
    \end{subfigure}

    \begin{subfigure}{0.45\linewidth}
        \includegraphics[width=\linewidth, trim=0 75px 0 0, clip]{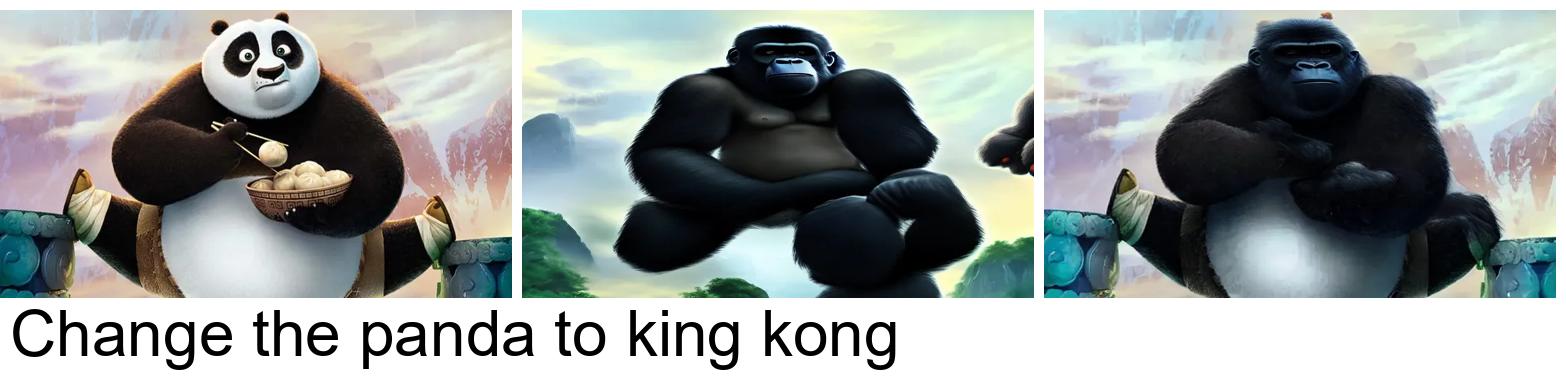}
        \parbox{\linewidth}{\vspace{-0.02in}
        ``Change the panda to king kong''
        \vspace{0.05in}}
    \end{subfigure}
    \begin{subfigure}{0.45\linewidth}
        \includegraphics[width=\linewidth, trim=0 75px 0 0, clip]{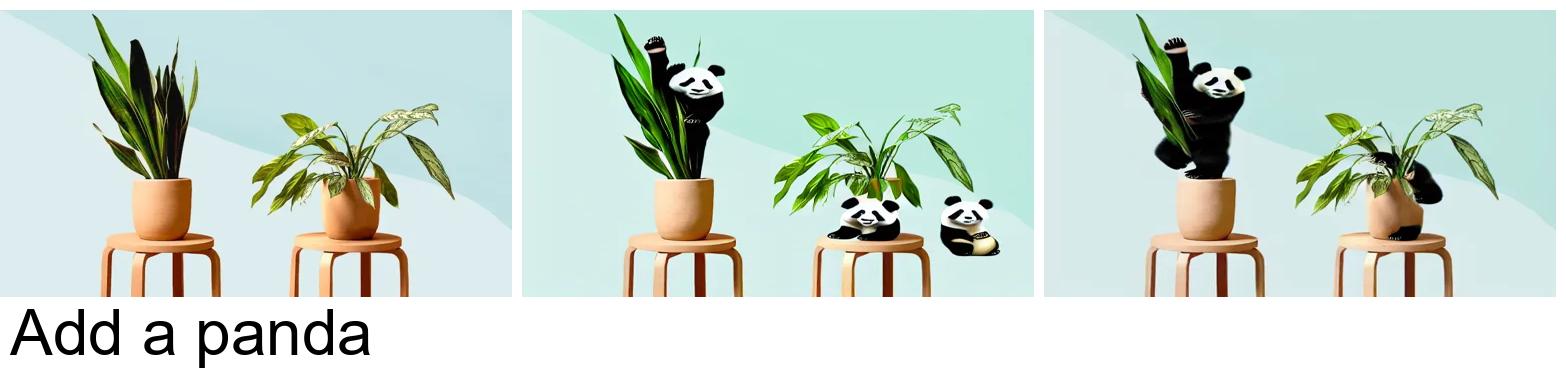}
        \parbox{\linewidth}{\vspace{-0.02in}
        ``Add a panda''
        \vspace{0.05in}}
    \end{subfigure}

    \begin{subfigure}{0.45\linewidth}
        \includegraphics[width=\linewidth, trim=0 135px 0 0, clip]{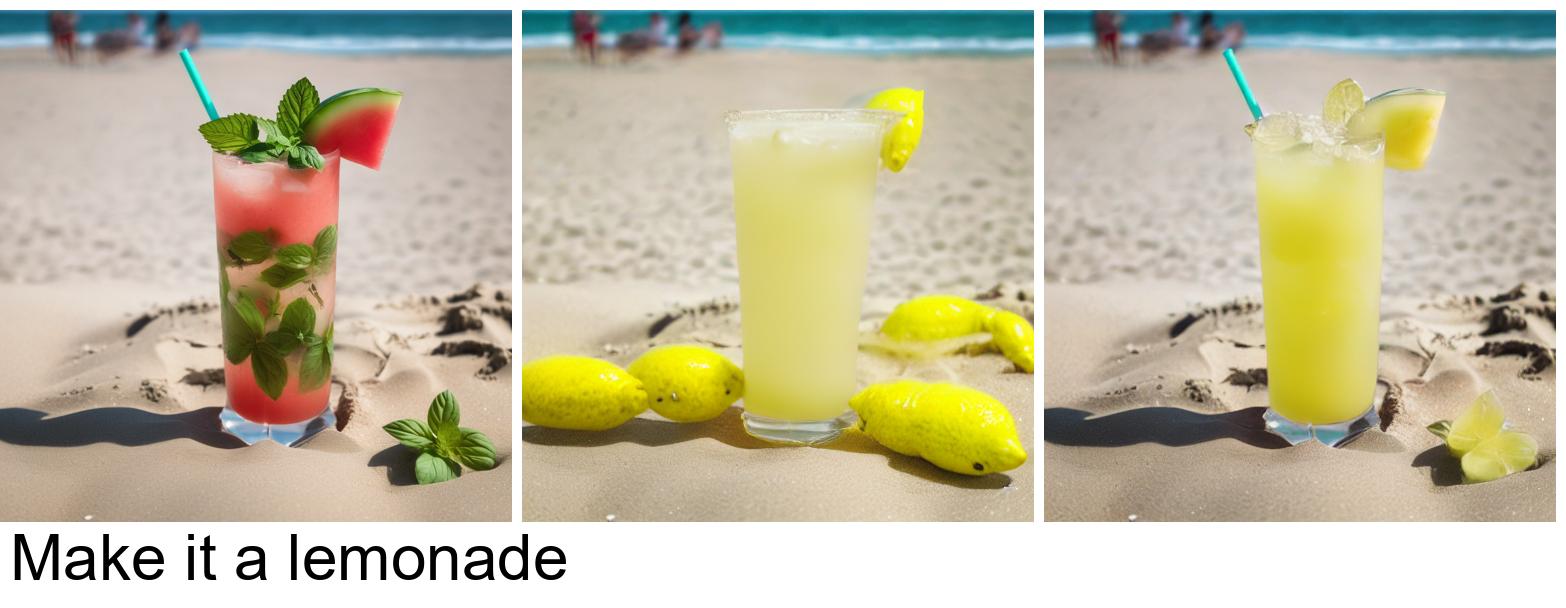}
        \parbox{\linewidth}{\vspace{-0.02in}
        ``Make it a lemonade''
        \vspace{0.05in}}
    \end{subfigure}
    \begin{subfigure}{0.45\linewidth}
        \includegraphics[width=\linewidth, trim=0 135px 0 0, clip]{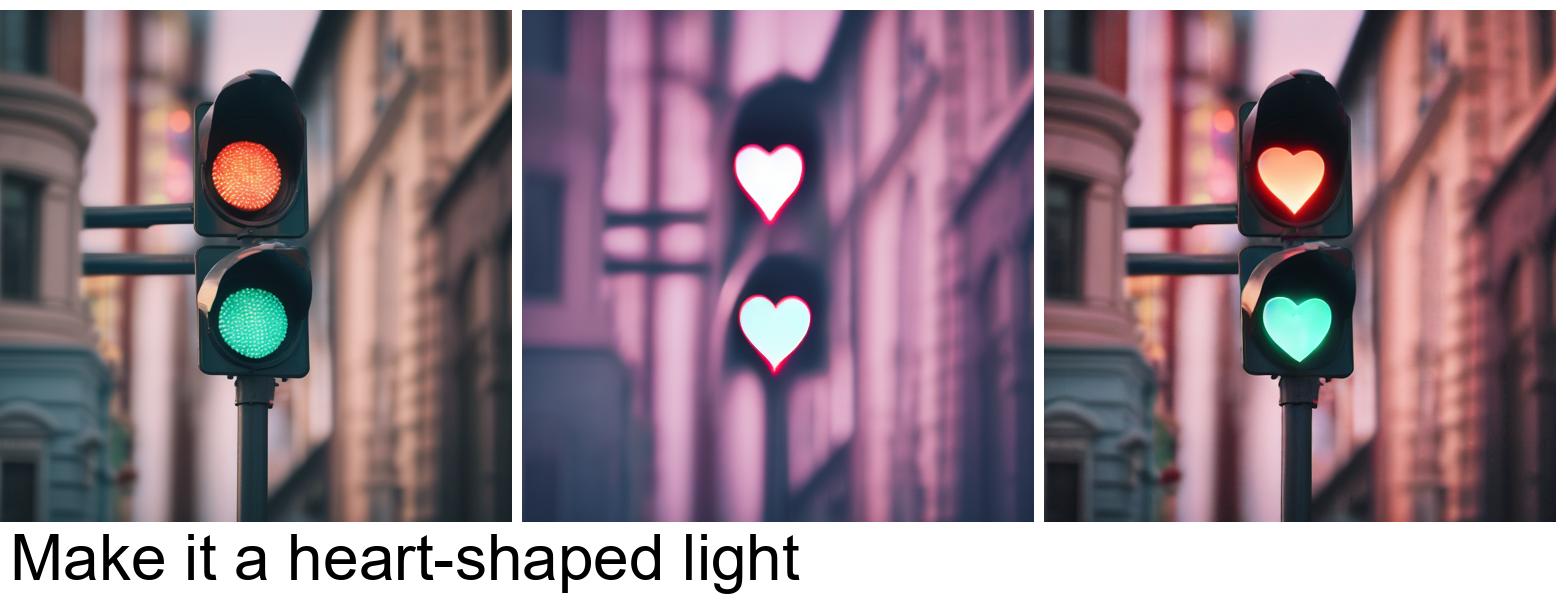}
        \parbox{\linewidth}{\vspace{-0.02in}
        ``Make it a heart-shaped light''
        \vspace{0.05in}}
    \end{subfigure}

    \begin{subfigure}{0.45\linewidth}
        \includegraphics[width=\linewidth, trim=0 135px 0 0, clip]{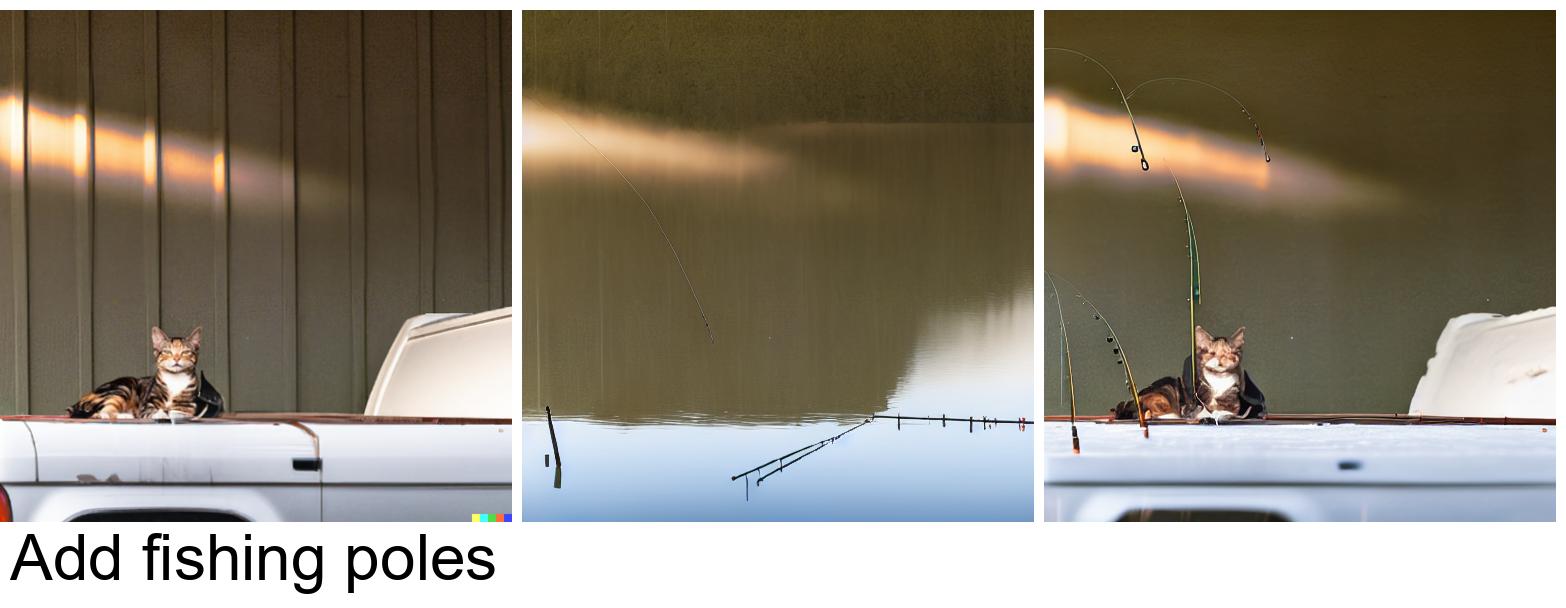}
        \parbox{\linewidth}{\vspace{-0.02in}
        ``Add fishing poles''
        \vspace{0.05in}}
    \end{subfigure}
    \begin{subfigure}{0.45\linewidth}
        \includegraphics[width=\linewidth, trim=0 135px 0 0, clip]{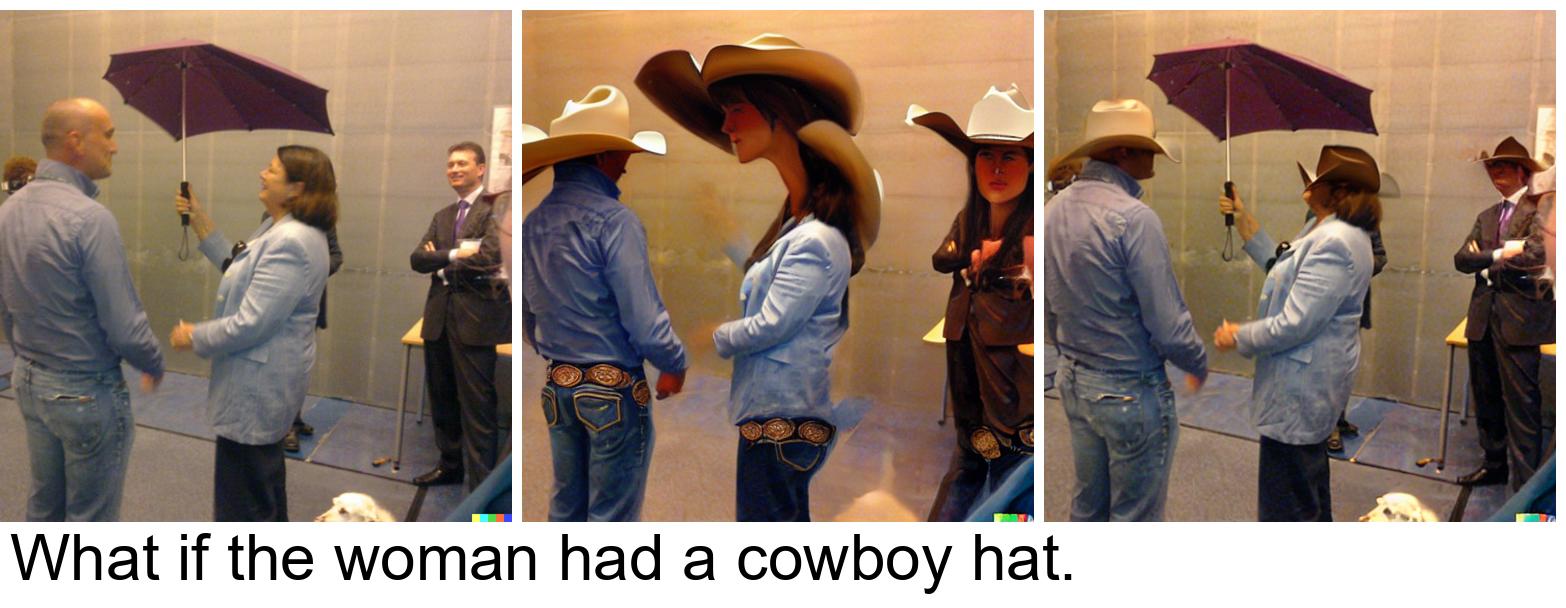}
        \parbox{\linewidth}{\vspace{-0.02in}
        ``What if the woman had a cowboy hat''
        \vspace{0.05in}}
    \end{subfigure}

    \begin{subfigure}{0.45\linewidth}
        \includegraphics[width=\linewidth, trim=0 135px 0 0, clip]{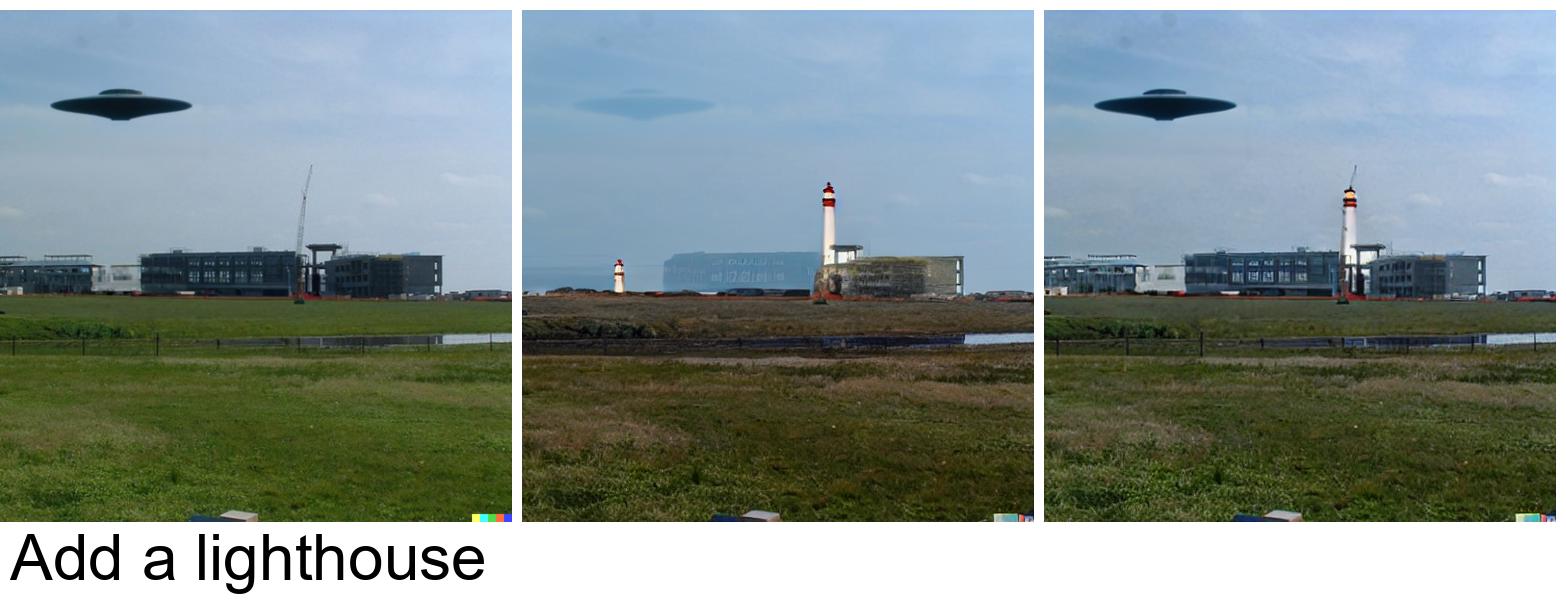}
        \parbox{\linewidth}{\vspace{-0.02in}
        ``Add a lighthouse''
        \vspace{0.05in}}
    \end{subfigure}
    \begin{subfigure}{0.45\linewidth}
        \includegraphics[width=\linewidth, trim=0 135px 0 0, clip]{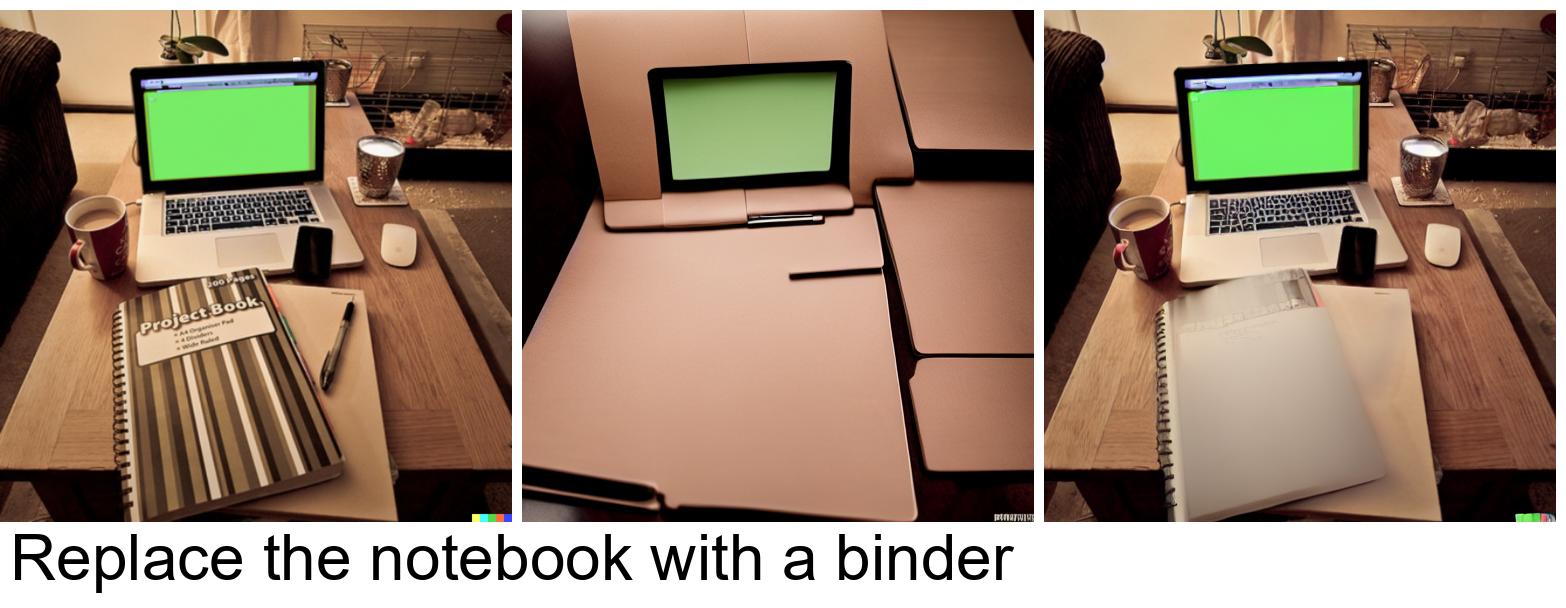}
        \parbox{\linewidth}{\vspace{-0.02in}
        ``Replace the notebook with a binder''
        \vspace{0.05in}}
    \end{subfigure}

    \begin{subfigure}{0.45\linewidth}
        \includegraphics[width=\linewidth, trim=0 135px 0 0, clip]{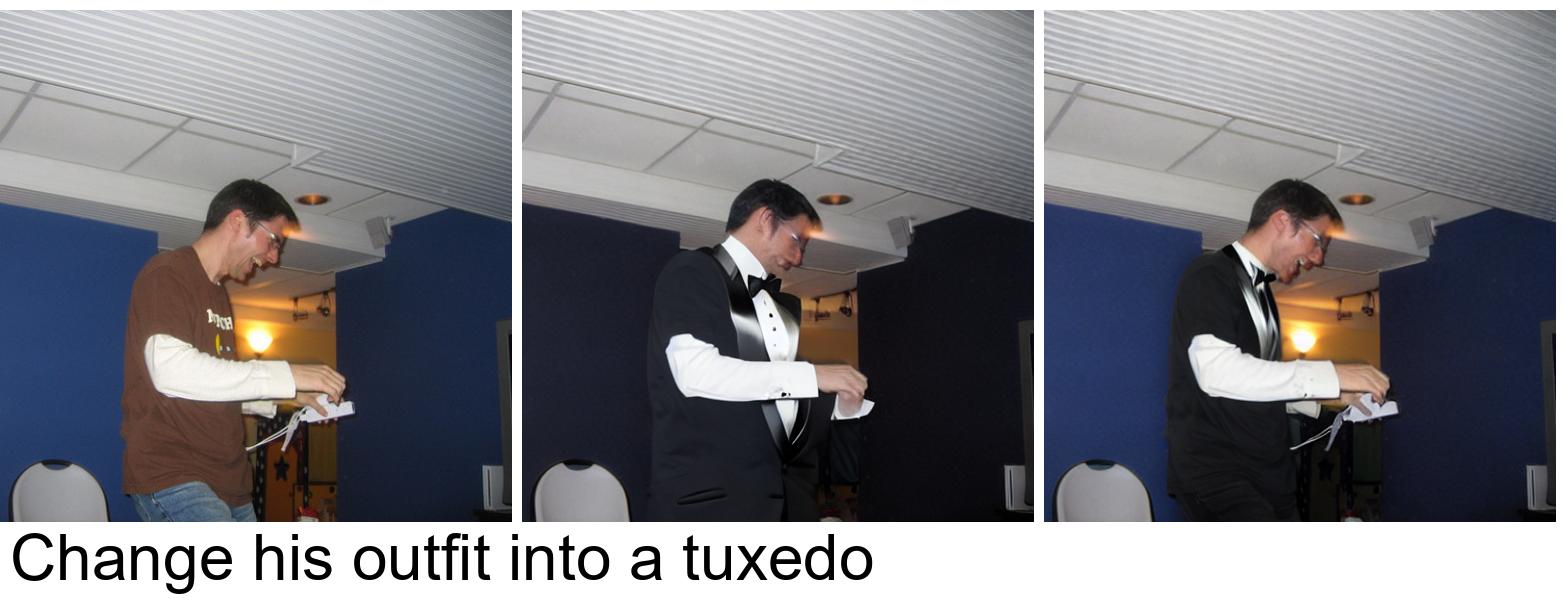}
        \parbox{\linewidth}{\vspace{-0.02in}
        ``Change his outfit into a tuxedo''
        \vspace{0.05in}}
    \end{subfigure}
    \begin{subfigure}{0.45\linewidth}
        \includegraphics[width=\linewidth, trim=0 135px 0 0, clip]{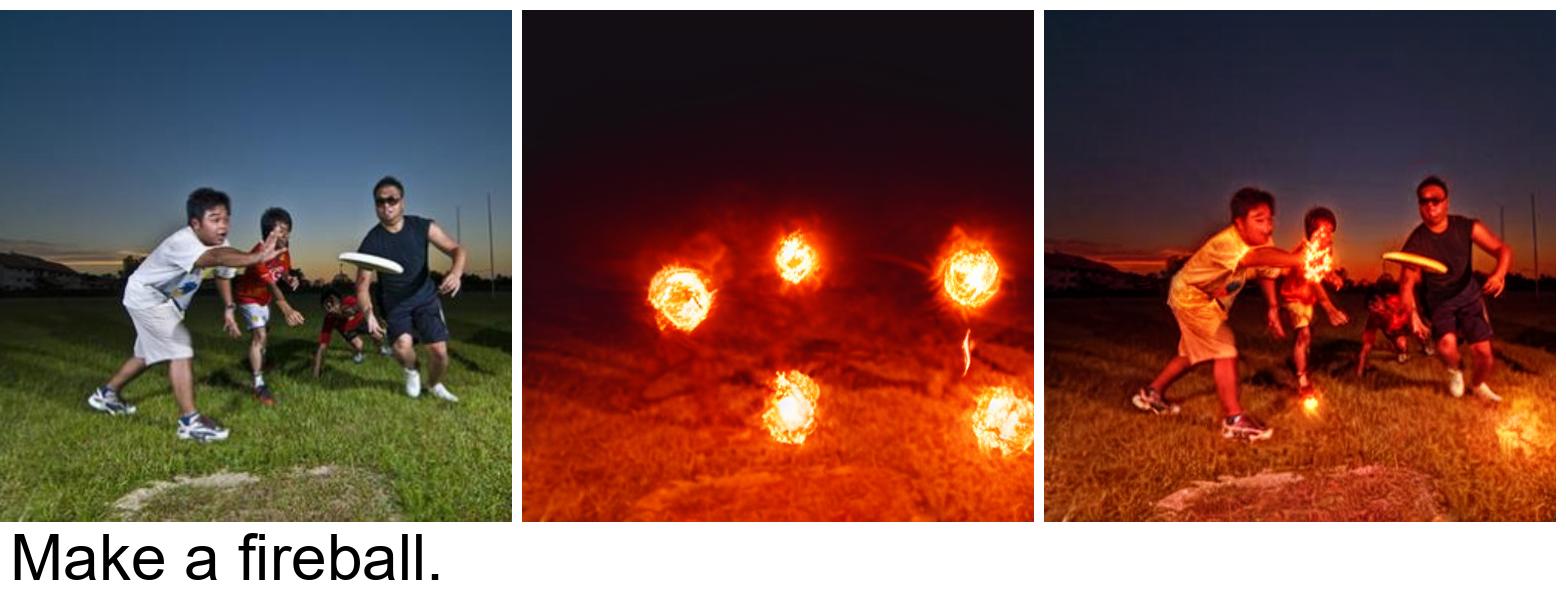}
        \parbox{\linewidth}{\vspace{-0.02in}
        ``Make a fireball''
        \vspace{0.05in}}
    \end{subfigure}

    \begin{subfigure}{0.45\linewidth}
        \includegraphics[width=\linewidth, trim=0 135px 0 0, clip]{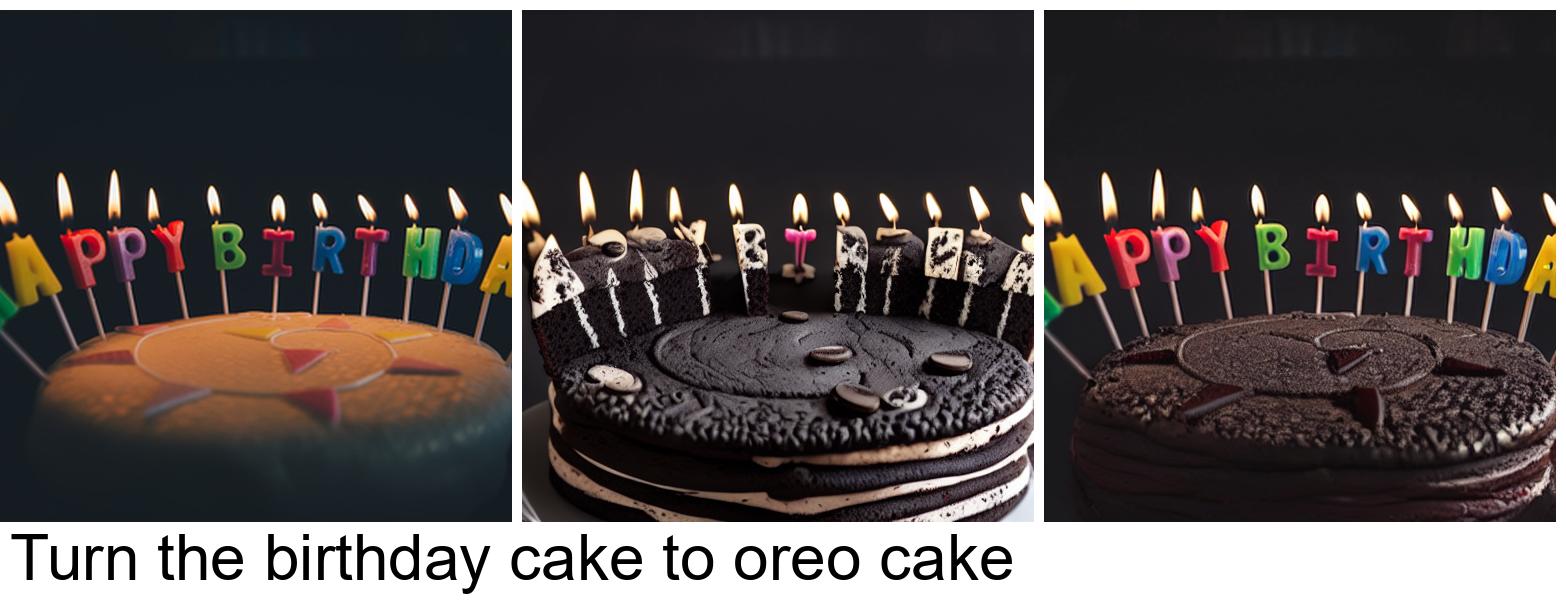}
        \parbox{\linewidth}{\vspace{-0.02in}
        ``Turn the birthday cake to oreo cake''
        \vspace{0.05in}}
    \end{subfigure}
    \begin{subfigure}{0.45\linewidth}
        \includegraphics[width=\linewidth, trim=0 135px 0 0, clip]{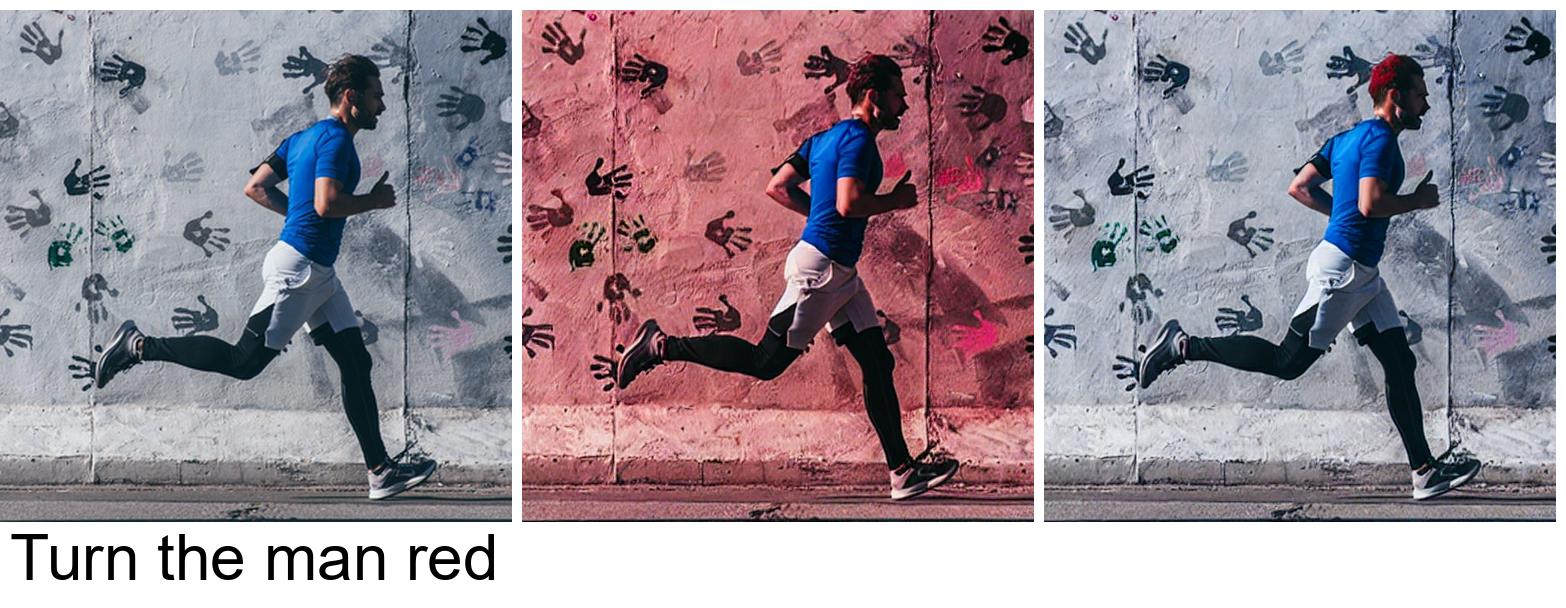}
        \parbox{\linewidth}{\vspace{-0.02in}
        ``Turn the man red''
        \vspace{0.05in}}
    \end{subfigure}

    \caption{Additional results from our \ourwork{} image editing method on benchmarks~\cite{zhang2024magicbrush,zhang2023hive,zhao2024instructbrush,li2024zone} (Part 2/2)}
    \label{fig:additional_editing_results_2}
\end{figure*}

\begin{figure*}
    \centering
    \begin{tabbing}
        \hspace{2.7em}\= \hspace{5.9em} \= \hspace{5.5em}\= \hspace{5.9em} \= \hspace{5.7em}\= \hspace{5.7em} \= \hspace{5.7em}\= \hspace{5.7em} \= \kill
        \> Original \> Edited \> Original \> Edited  \> Original \> Edited  \> Original \> Edited \\
    \end{tabbing} 
    \vspace{-0.3in}

    \tiny
    \begin{subfigure}{0.23\linewidth}
        \includegraphics[width=\linewidth, trim=0 100px 0 0, clip]{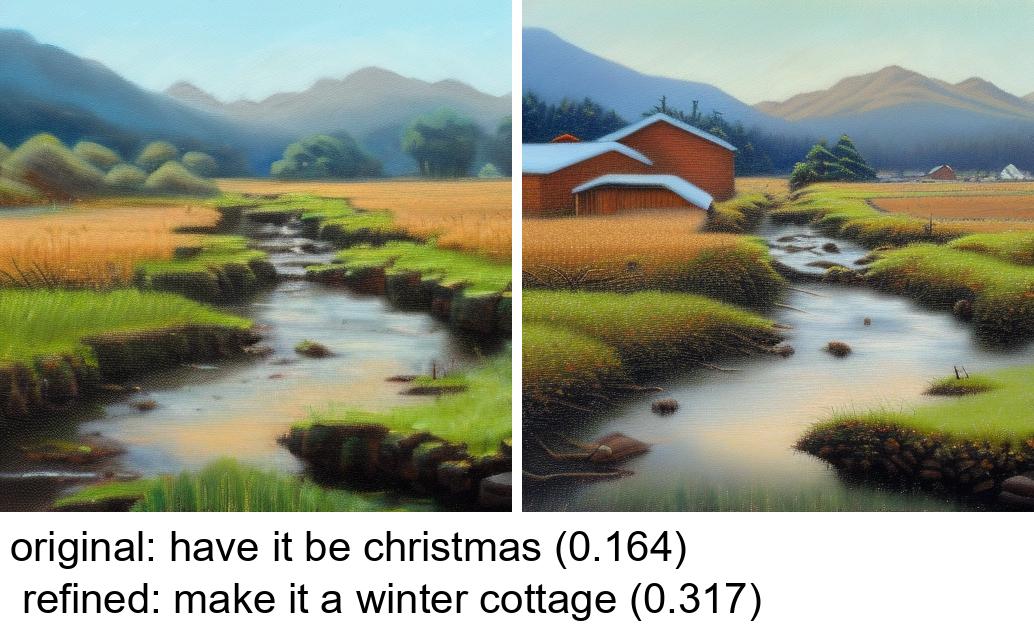}
        \parbox{\linewidth}{
        {\sc Original:} ``have it be christmas'' (0.164)\\
        {\sc Refined:} ``make it a winter cottage'' (0.317)\\
        }
    \end{subfigure}
    \begin{subfigure}{0.23\linewidth}
        \includegraphics[width=\linewidth, trim=0 100px 0 0, clip]{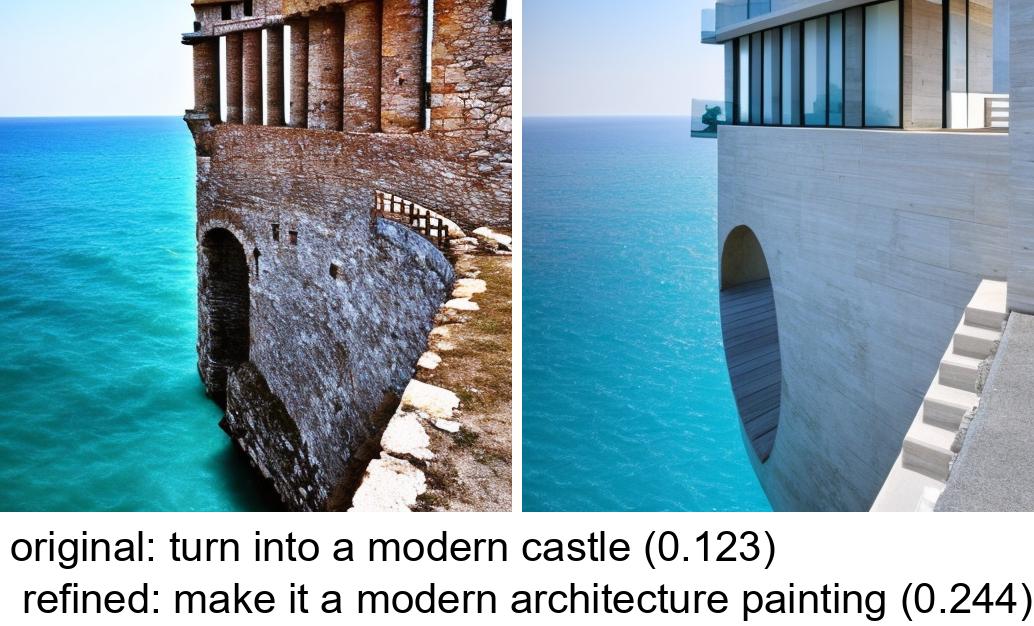}
        \parbox{\linewidth}{
        {\sc Original:} ``turn into a modern castle'' (0.123)\\
        {\sc Refined:} ``make it a modern architecture'' (0.224)\\
        }
    \end{subfigure}
    \begin{subfigure}{0.23\linewidth}
        \includegraphics[width=\linewidth, trim=0 100px 0 0, clip]{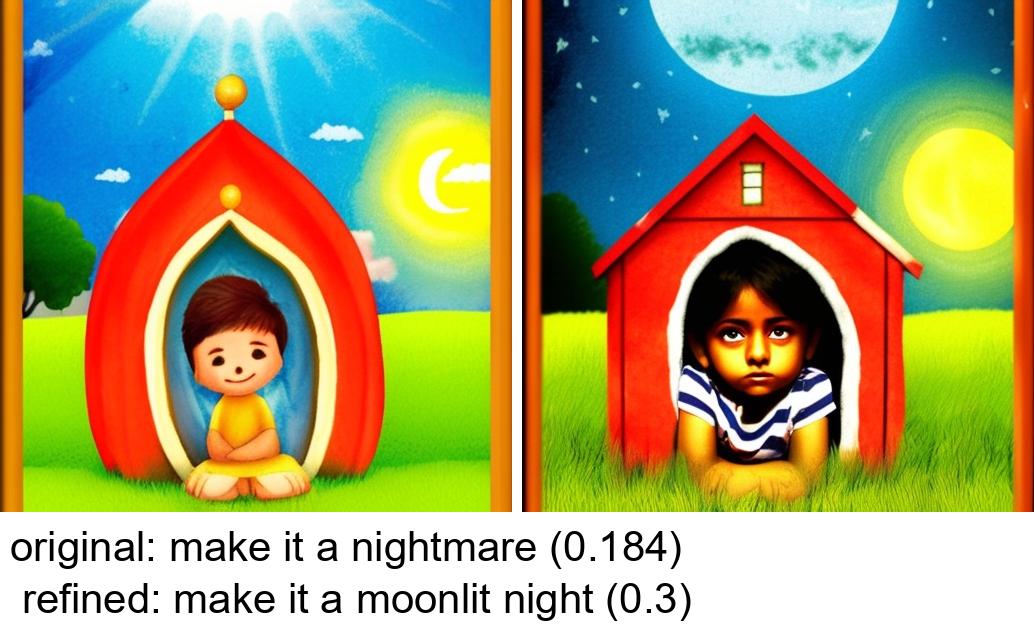}
        \parbox{\linewidth}{
        {\sc Original:} ``make it a nightmare'' (0.184)\\
        {\sc Refined:} ``make it a moonlit night'' (0.300)\\
        }
    \end{subfigure}
    \begin{subfigure}{0.23\linewidth}
        \includegraphics[width=\linewidth, trim=0 100px 0 0, clip]{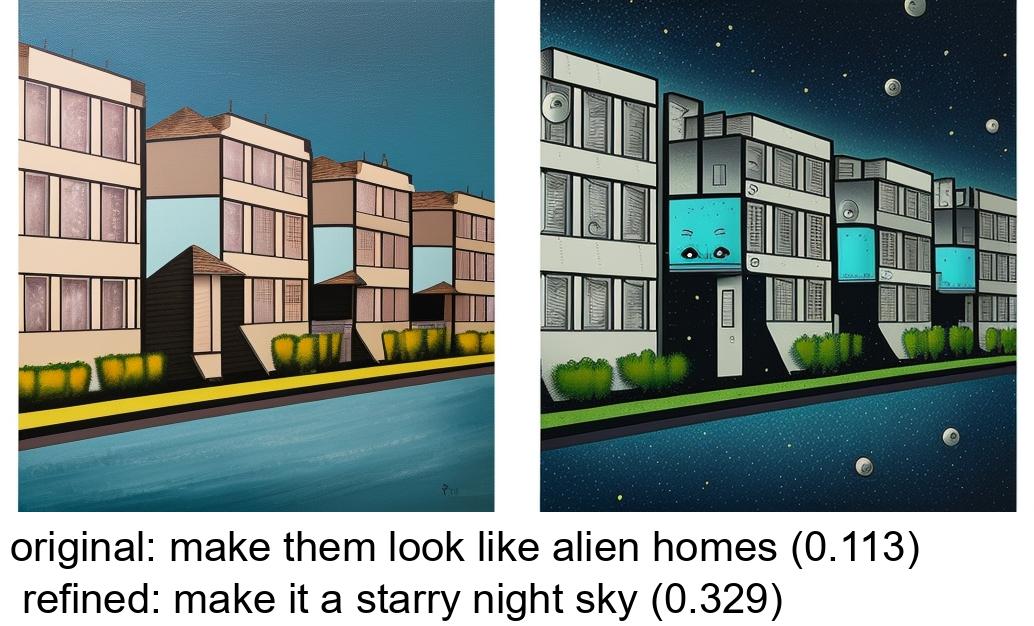}
        \parbox{\linewidth}{
        {\sc Original:} ``make them look like alien homes'' (0.113)\\
        {\sc Refined:} ``make it a starry night sky'' (0.329)\\
        }
    \end{subfigure}

    \begin{subfigure}{0.23\linewidth}
        \includegraphics[width=\linewidth, trim=0 100px 0 0, clip]{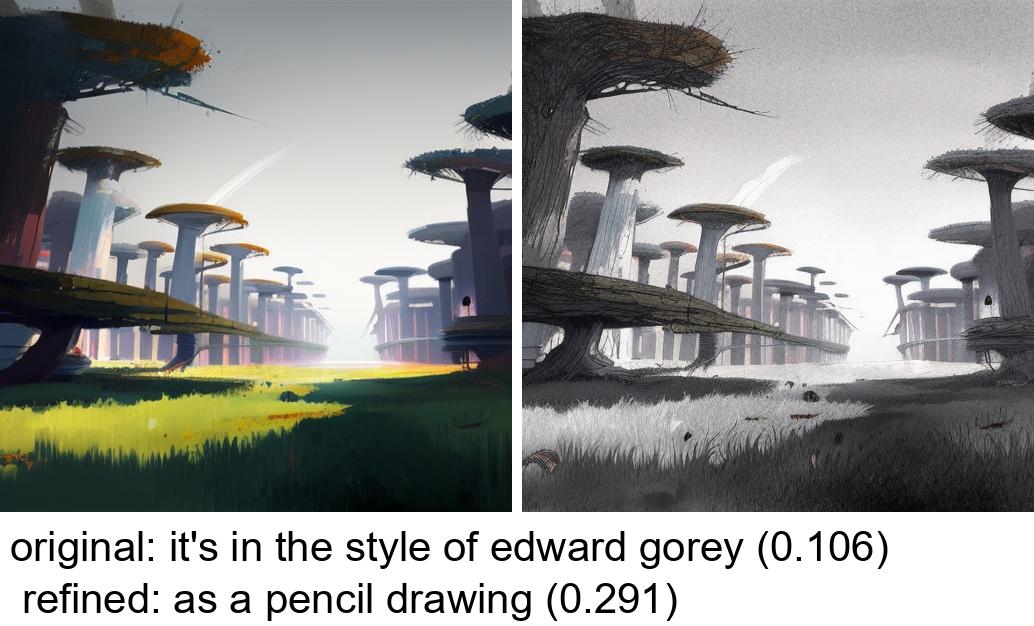}
        \parbox{\linewidth}{
        {\sc Original:} ``it's in the style of edward gorey'' (0.106)\\
        {\sc Refined:} ``as a pencil drawing'' (0.291)\\
        }
    \end{subfigure}
    \begin{subfigure}{0.23\linewidth}
        \includegraphics[width=\linewidth, trim=0 100px 0 0, clip]{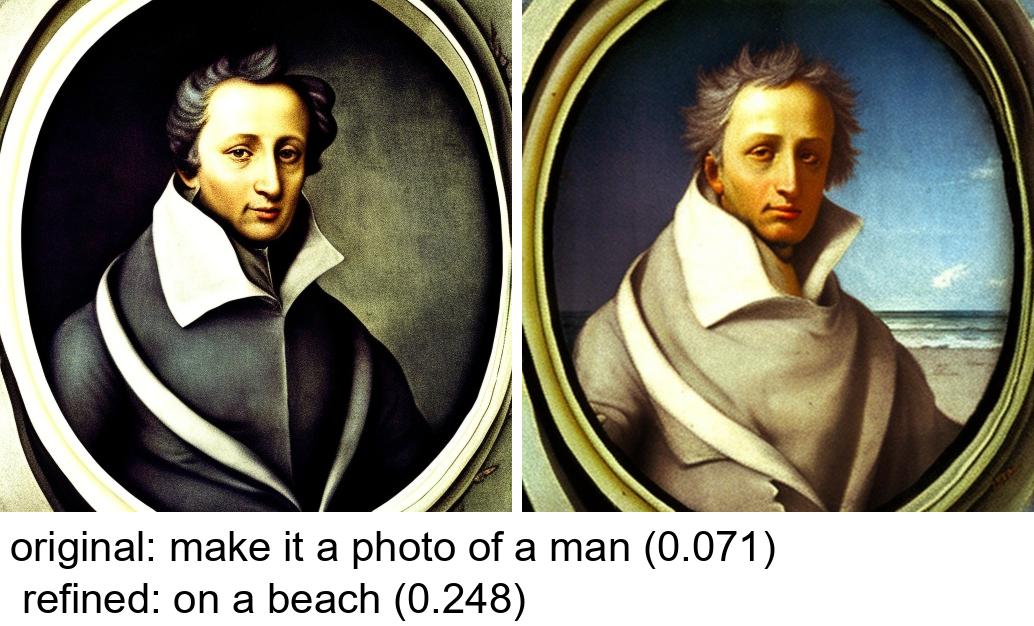}
        \parbox{\linewidth}{
        {\sc Original:} ``make it a photo of a man'' (0.071)\\
        {\sc Refined:} ``on a beach'' (0.248)\\
        }
    \end{subfigure}
    \begin{subfigure}{0.23\linewidth}
        \includegraphics[width=\linewidth, trim=0 100px 0 0, clip]{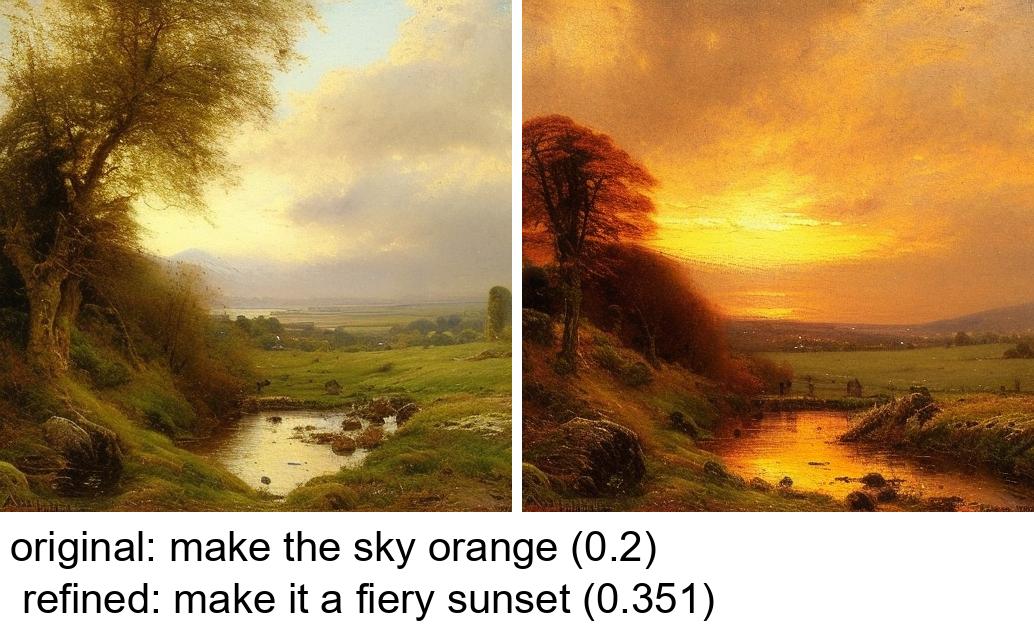}
        \parbox{\linewidth}{
        {\sc Original:} ``make the sky orange'' (0.200)\\
        {\sc Refined:} ``make it a fiery sunset'' (0.351)\\
        }
    \end{subfigure}
    \begin{subfigure}{0.23\linewidth}
        \includegraphics[width=\linewidth, trim=0 100px 0 0, clip]{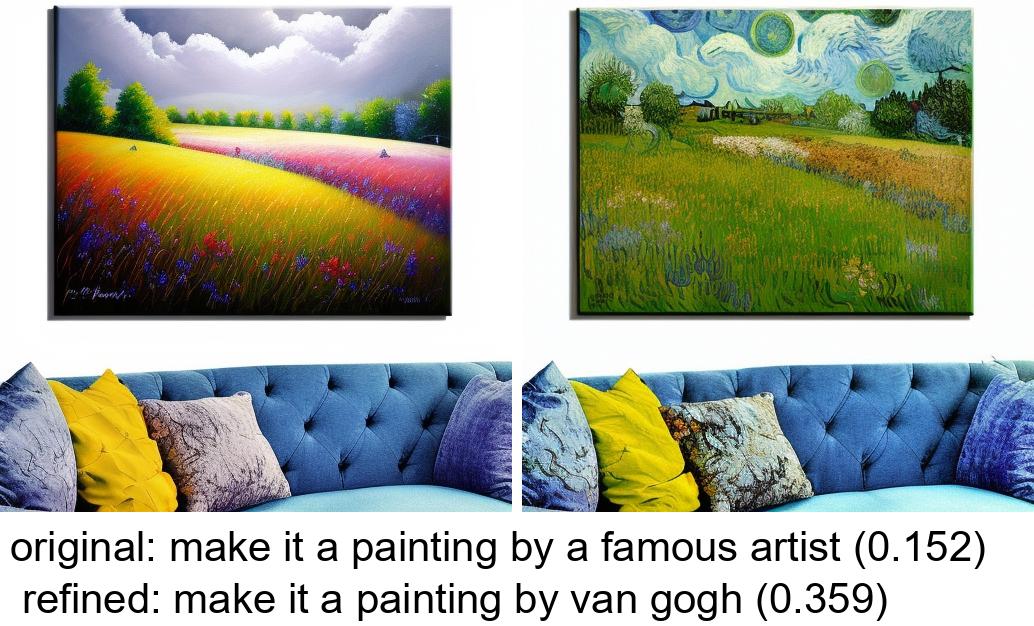}
        \parbox{\linewidth}{
        {\sc Original:} ``make it a painting by a famous artist'' (0.152)\\
        {\sc Refined:} ``make it a painting by van gogh'' (0.359)\\
        }
    \end{subfigure}

    \begin{subfigure}{0.23\linewidth}
        \includegraphics[width=\linewidth, trim=0 100px 0 0, clip]{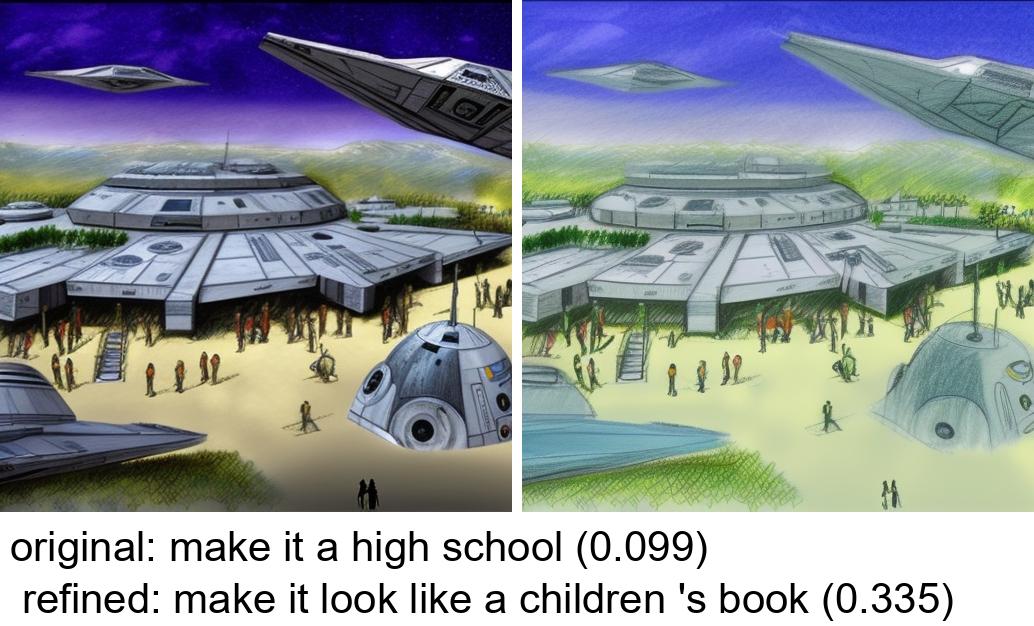}
        \parbox{\linewidth}{
        {\sc Original:} ``make it a high school'' (0.099)\\
        {\sc Refined:} ``make it look like a children's book'' (0.335)\\
        }
    \end{subfigure}
    \begin{subfigure}{0.23\linewidth}
        \includegraphics[width=\linewidth, trim=0 100px 0 0, clip]{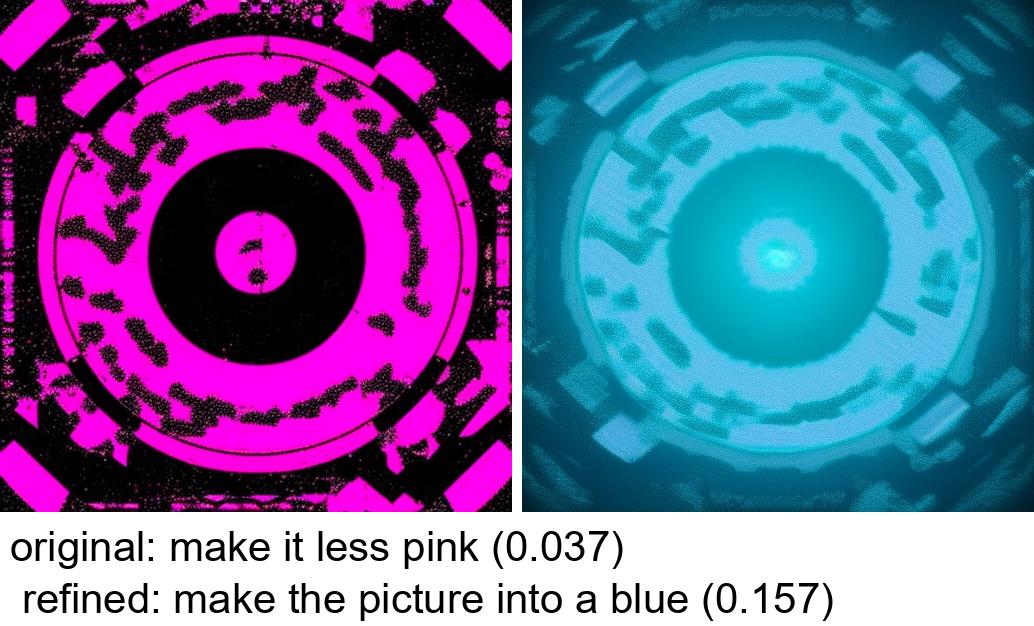}
        \parbox{\linewidth}{
        {\sc Original:} ``make it less pink'' (0.037)\\
        {\sc Refined:} ``make the picture into a blue'' (0.157)\\
        }
    \end{subfigure}
    \begin{subfigure}{0.23\linewidth}
        \includegraphics[width=\linewidth, trim=0 100px 0 0, clip]{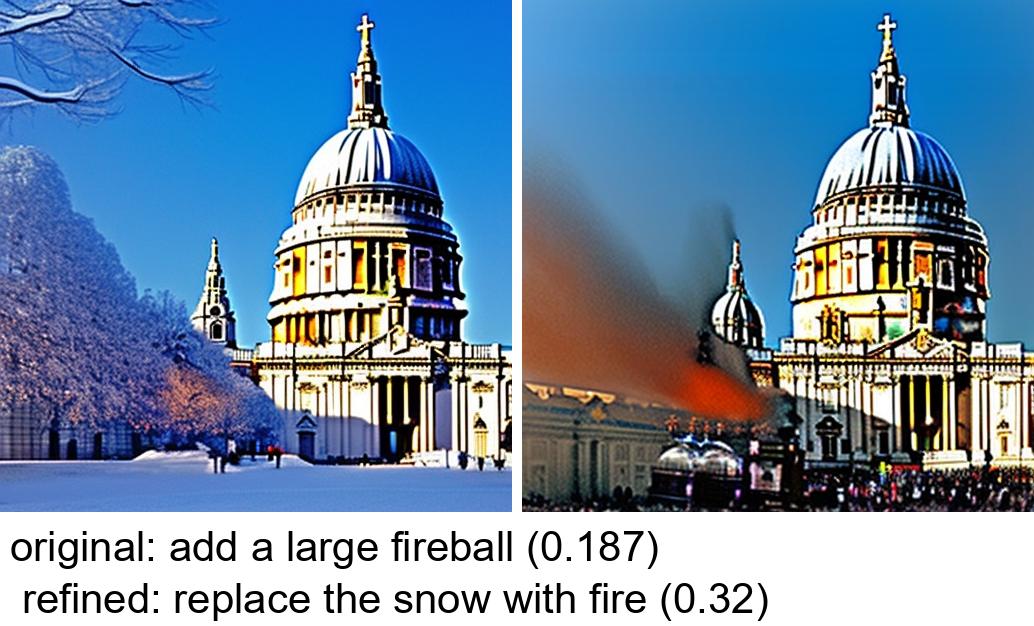}
        \parbox{\linewidth}{
        {\sc Original:} ``add a large fireball'' (0.187)\\
        {\sc Refined:} ``replace the snow with fire'' (0.320)\\
        }
    \end{subfigure}
    \begin{subfigure}{0.23\linewidth}
        \includegraphics[width=\linewidth, trim=0 100px 0 0, clip]{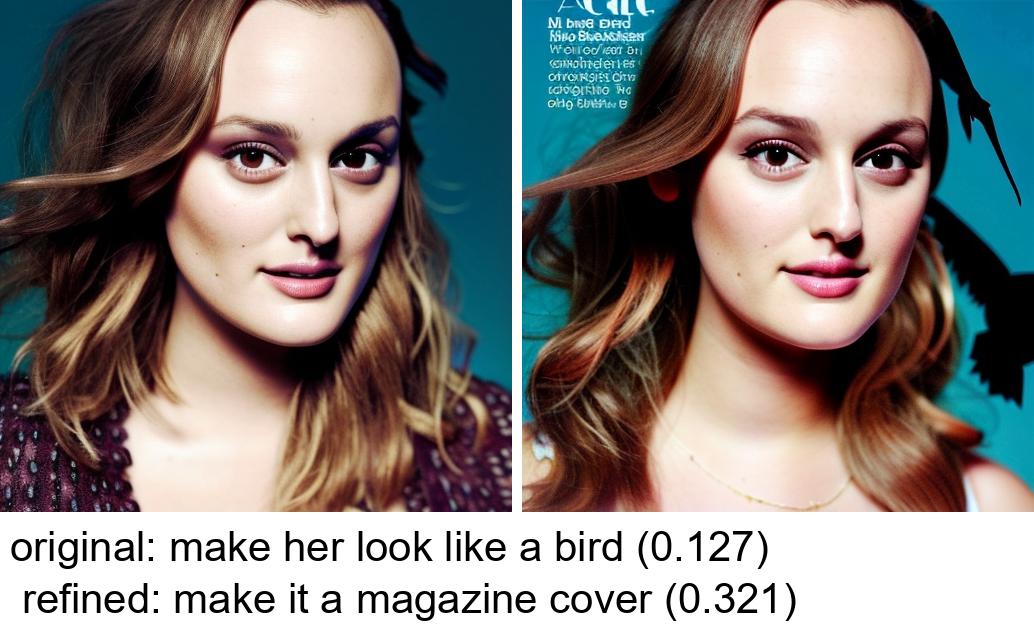}
        \parbox{\linewidth}{
        {\sc Original:} ``make her look like a bird'' (0.127)\\
        {\sc Refined:} ``make it a magazine cover'' (0.321)\\
        }
    \end{subfigure}

    \begin{subfigure}{0.23\linewidth}
        \includegraphics[width=\linewidth, trim=0 100px 0 0, clip]{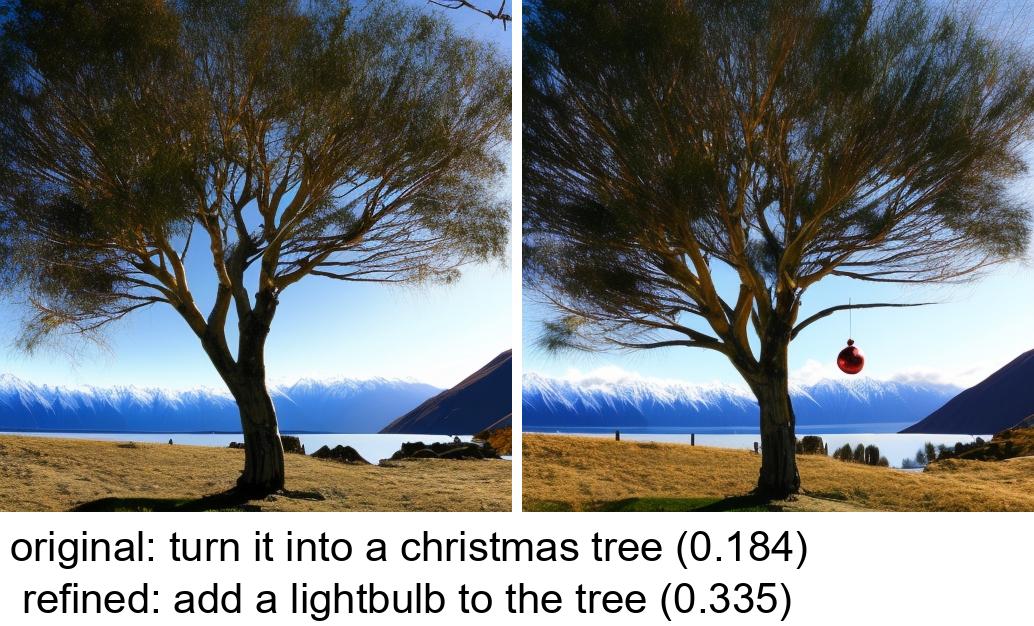}
        \parbox{\linewidth}{
        {\sc Original:} ``turn it into a christmas tree'' (0.184)\\
        {\sc Refined:} ``add a lightbulb to the tree'' (0.335)\\
        }
    \end{subfigure}
    \begin{subfigure}{0.23\linewidth}
        \includegraphics[width=\linewidth, trim=0 100px 0 0, clip]{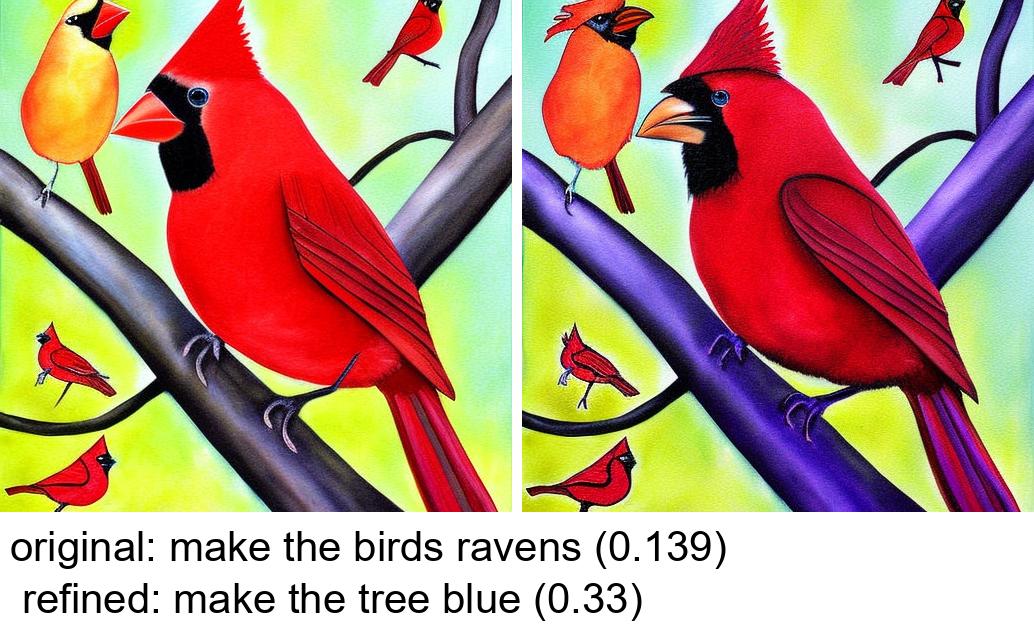}
        \parbox{\linewidth}{
        {\sc Original:} ``make the birds ravens'' (0.127)\\
        {\sc Refined:} ``make the tree blue'' (0.321)\\
        }
    \end{subfigure}
    \begin{subfigure}{0.23\linewidth}
        \includegraphics[width=\linewidth, trim=0 100px 0 0, clip]{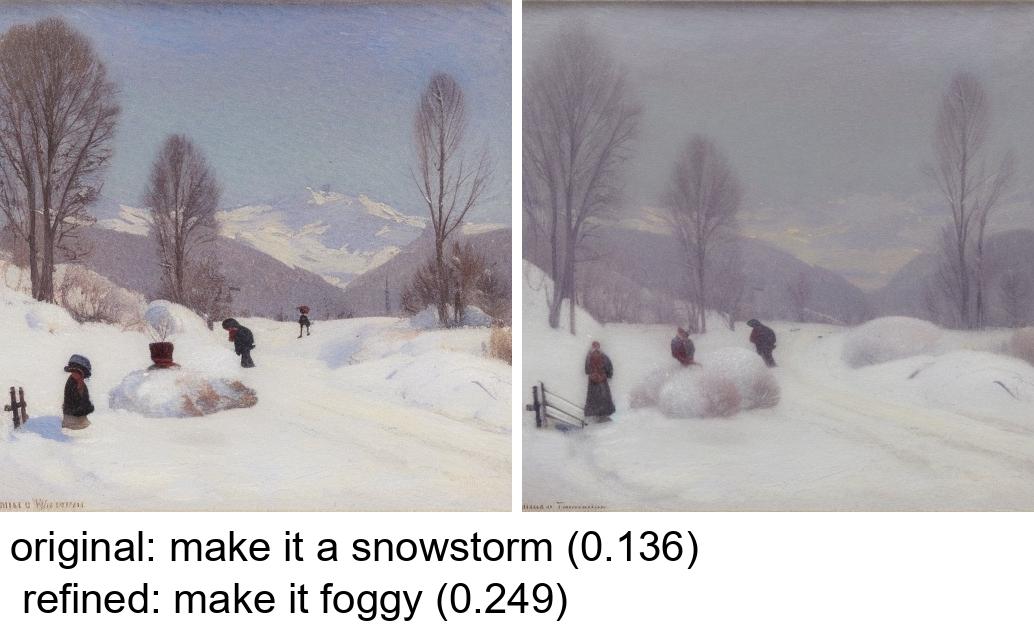}
        \parbox{\linewidth}{
        {\sc Original:} ``make it a snowstorm'' (0.248)\\
        {\sc Refined:} ``make it foggy'' (0.379)\\
        }
    \end{subfigure}
    \begin{subfigure}{0.23\linewidth}
        \includegraphics[width=\linewidth, trim=0 100px 0 0, clip]{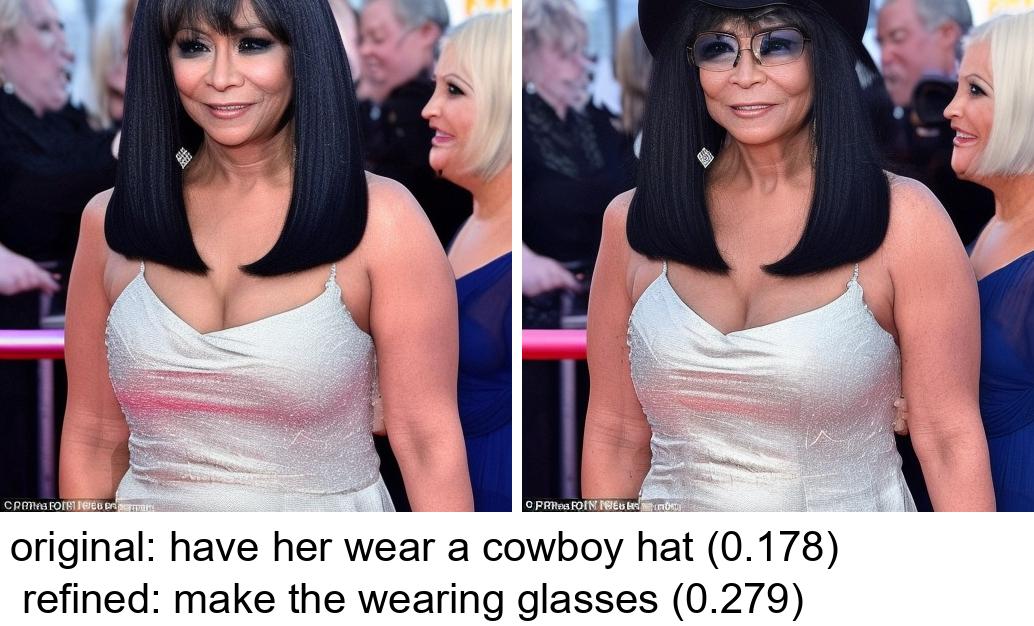}
        \parbox{\linewidth}{
        {\sc Original:} ``have her wear a cowboy hat'' (0.178)\\
        {\sc Refined:} ``make the wearing glasses'' (0.279)\\
        }
    \end{subfigure}

    \begin{subfigure}{0.23\linewidth}
        \includegraphics[width=\linewidth, trim=0 100px 0 0, clip]{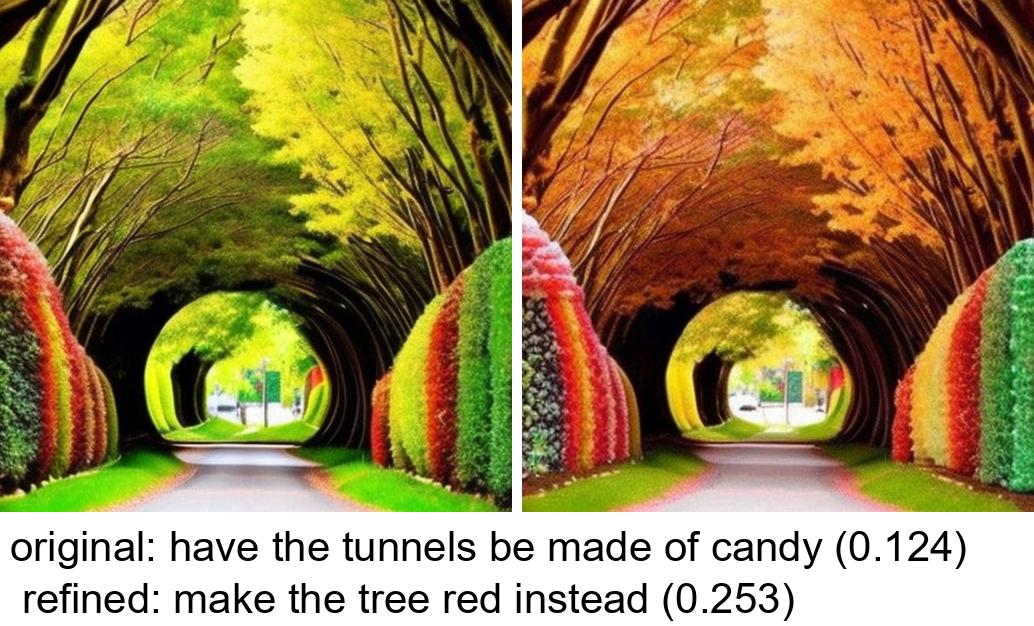}
        \parbox{\linewidth}{
        {\sc Original:} ``have the tunnels be made of candy'' (0.124)\\
        {\sc Refined:} ``make the tree red instead'' (0.253)\\
        }
    \end{subfigure}
    \begin{subfigure}{0.23\linewidth}
        \includegraphics[width=\linewidth, trim=0 100px 0 0, clip]{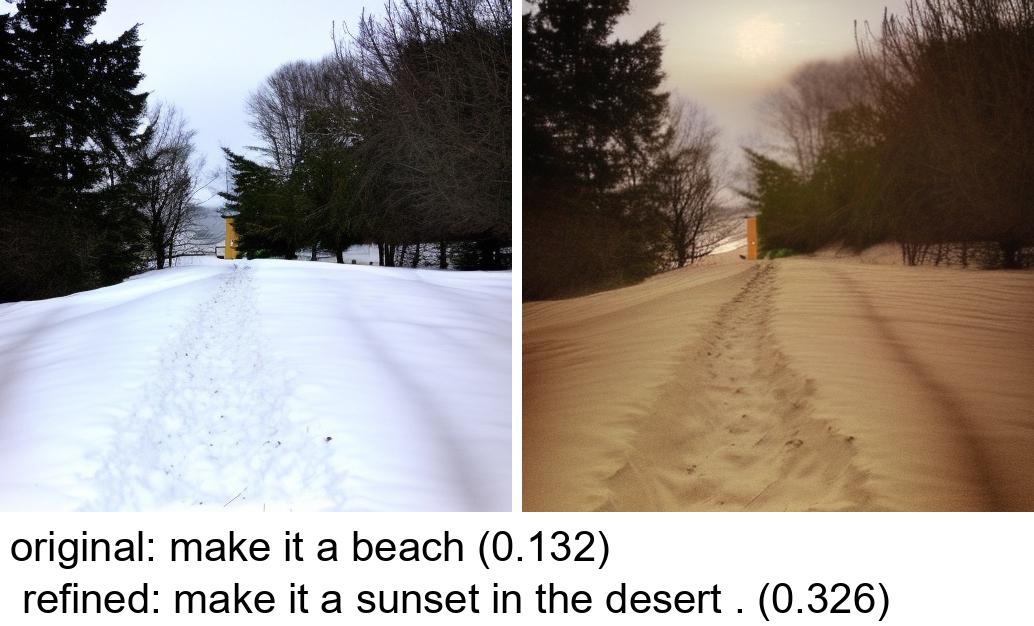}
        \parbox{\linewidth}{
        {\sc Original:} ``make it a beach'' (0.132)\\
        {\sc Refined:} ``make it a sunset in the desert'' (0.326)\\
        }
    \end{subfigure}
    \begin{subfigure}{0.23\linewidth}
        \includegraphics[width=\linewidth, trim=0 100px 0 0, clip]{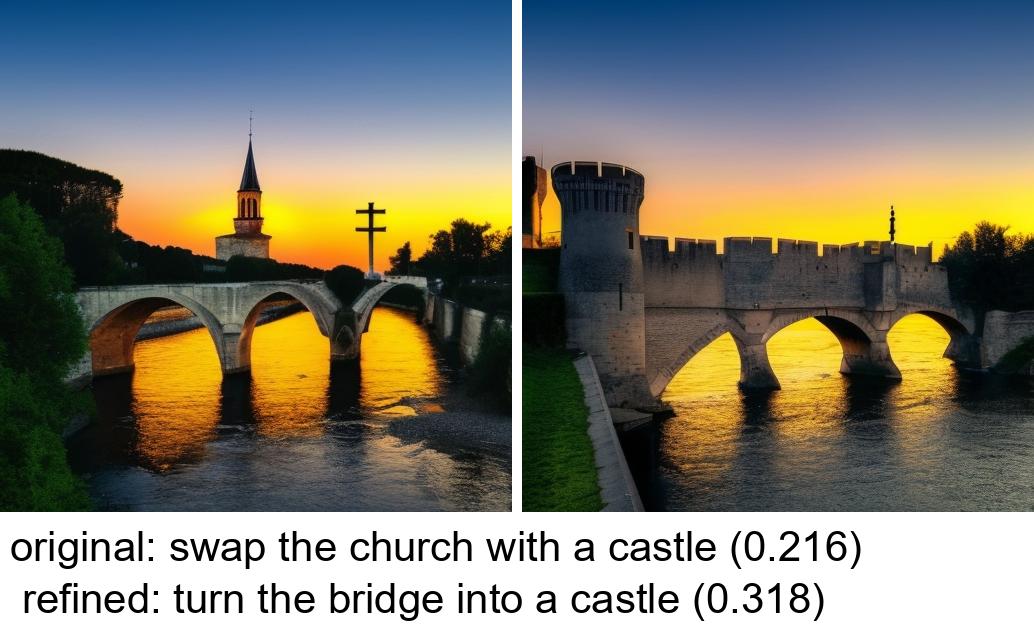}
        \parbox{\linewidth}{
        {\sc Original:} ``swap the church with a castle'' (0.216)\\
        {\sc Refined:} ``turn the bridge into a castle'' (0.318)\\
        }
    \end{subfigure}
    \begin{subfigure}{0.23\linewidth}
        \includegraphics[width=\linewidth, trim=0 100px 0 0, clip]{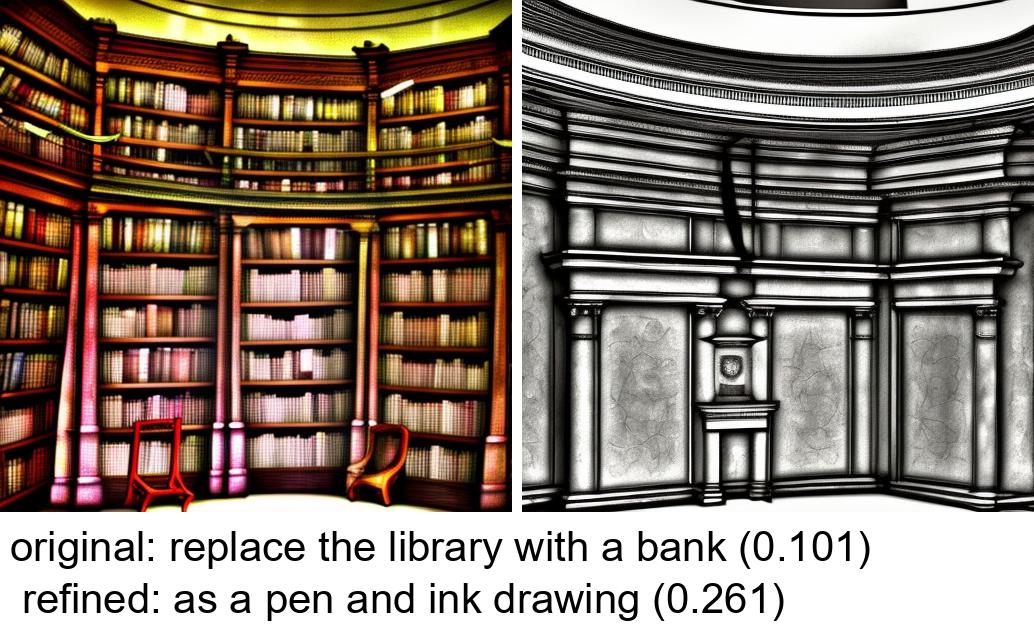}
        \parbox{\linewidth}{
        {\sc Original:} ``replace the library with a bank'' (0.101)\\
        {\sc Refined:} ``as a pen and ink drawing'' (0.261)\\
        }
    \end{subfigure}

    \begin{subfigure}{0.23\linewidth}
        \includegraphics[width=\linewidth, trim=0 100px 0 0, clip]{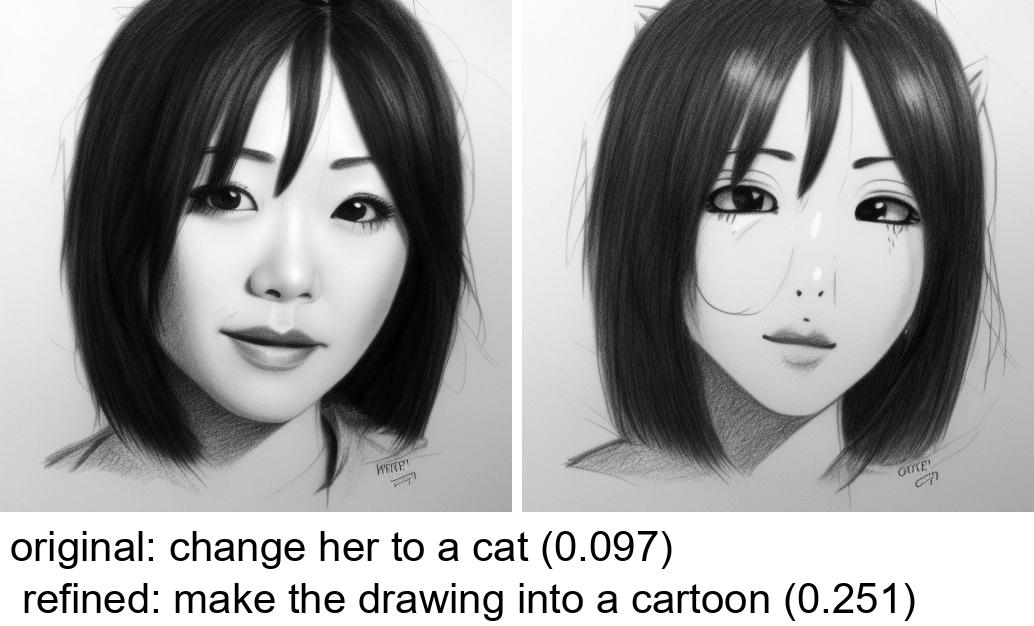}
        \parbox{\linewidth}{
        {\sc Original:} ``change her to a cat'' (0.097)\\
        {\sc Refined:} ``make the drawing into a cartoon'' (0.251)\\
        }
    \end{subfigure}
    \begin{subfigure}{0.23\linewidth}
        \includegraphics[width=\linewidth, trim=0 100px 0 0, clip]{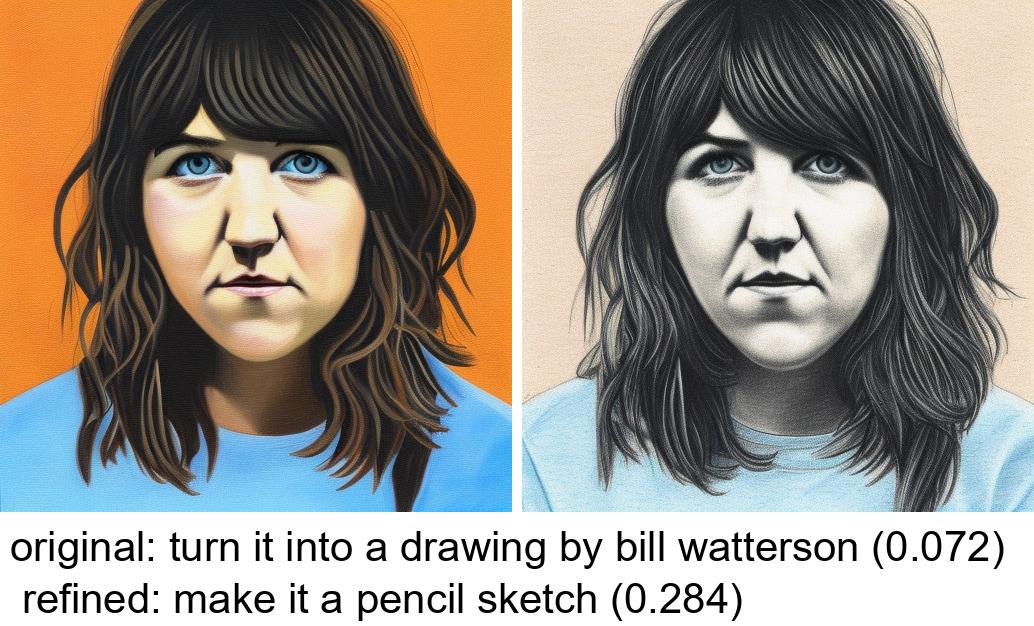}
        \parbox{\linewidth}{
        {\sc Original:} ``turn into drawing by bill watterson'' (0.072)\\
        {\sc Refined:} ``make it a pencil sketch'' (0.284)\\
        }
    \end{subfigure}
    \begin{subfigure}{0.23\linewidth}
        \includegraphics[width=\linewidth, trim=0 100px 0 0, clip]{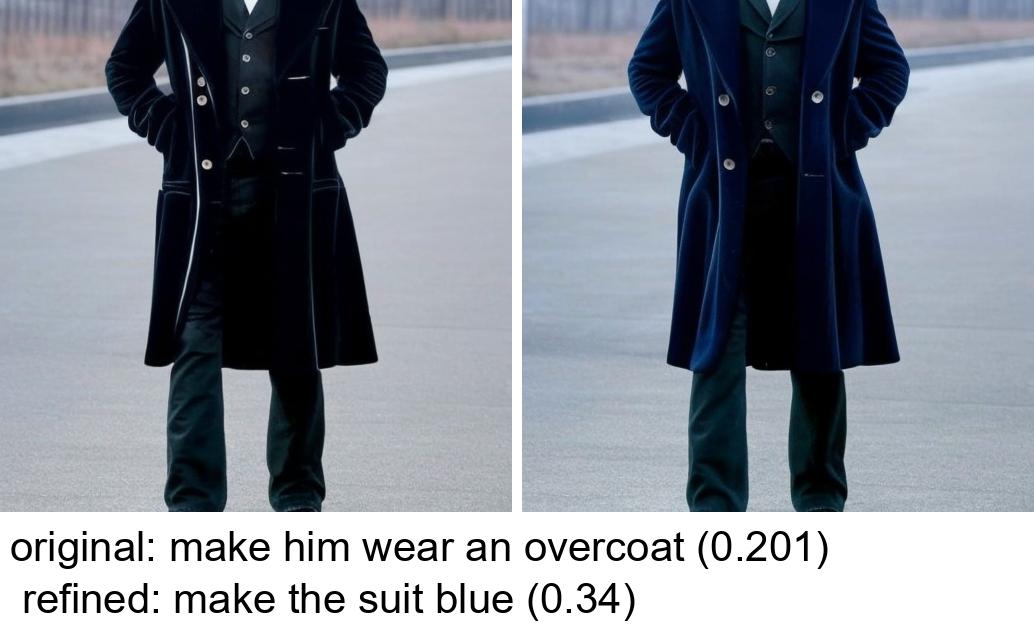}
        \parbox{\linewidth}{
        {\sc Original:} ``make him wear an overcoat'' (0.201)\\
        {\sc Refined:} ``make the suit blue'' (0.340)\\
        }
    \end{subfigure}
    \begin{subfigure}{0.23\linewidth}
        \includegraphics[width=\linewidth, trim=0 100px 0 0, clip]{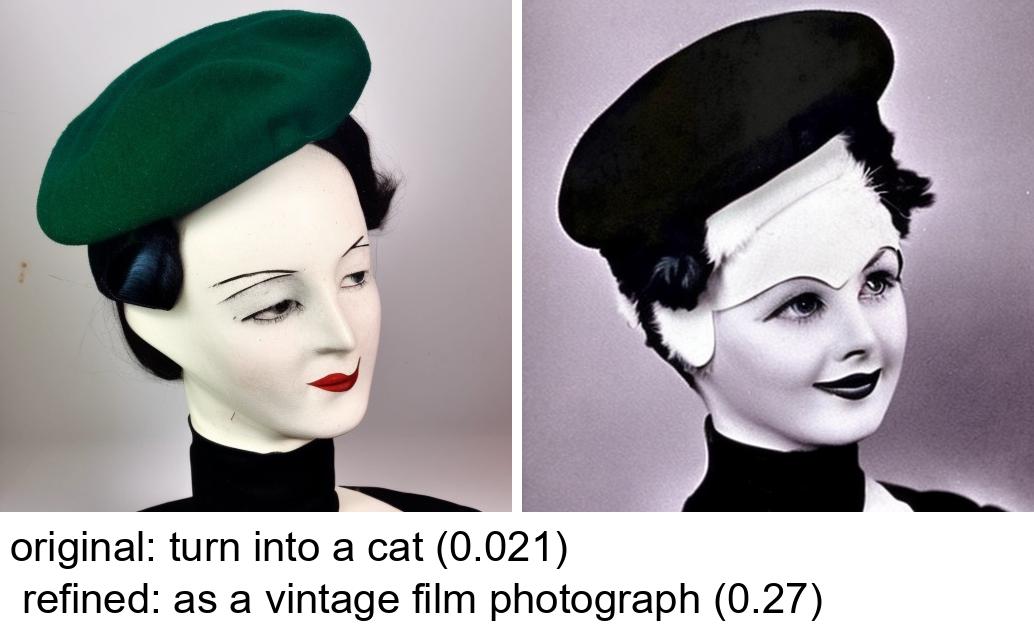}
        \parbox{\linewidth}{
        {\sc Original:} ``turn into a cat'' (0.021)\\
        {\sc Refined:} ``as a vintage film photograph'' (0.270)\\
        }
    \end{subfigure}

    \begin{subfigure}{0.23\linewidth}
        \includegraphics[width=\linewidth, trim=0 100px 0 0, clip]{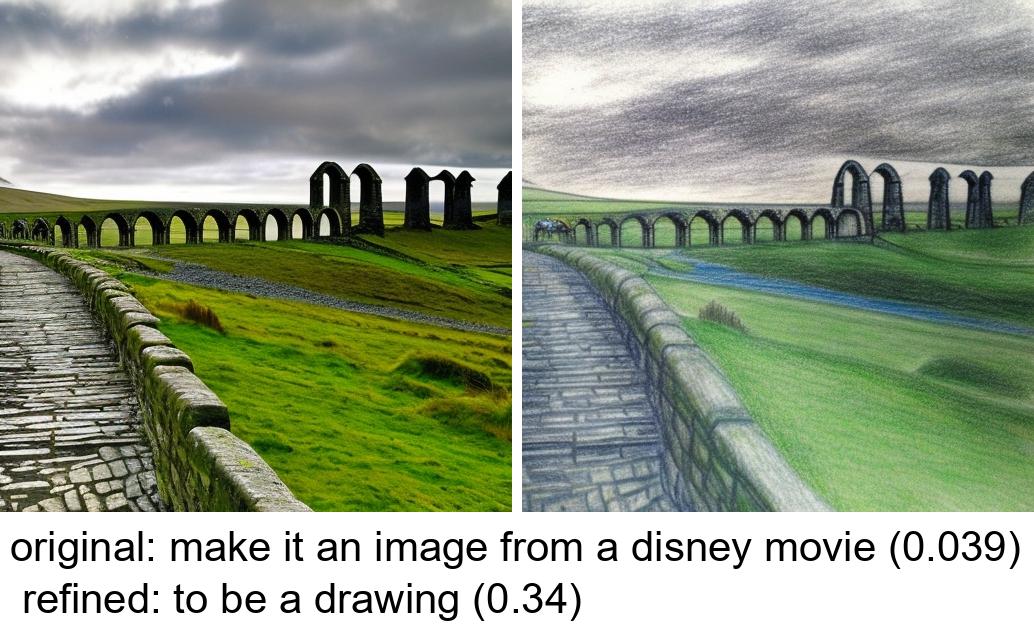}
        \parbox{\linewidth}{
        {\sc Original:} ``make it an image from a disney movie'' (0.039)\\
        {\sc Refined:} ``to be a drawing'' (0.340)\\
        }
    \end{subfigure}
    \begin{subfigure}{0.23\linewidth}
        \includegraphics[width=\linewidth, trim=0 100px 0 0, clip]{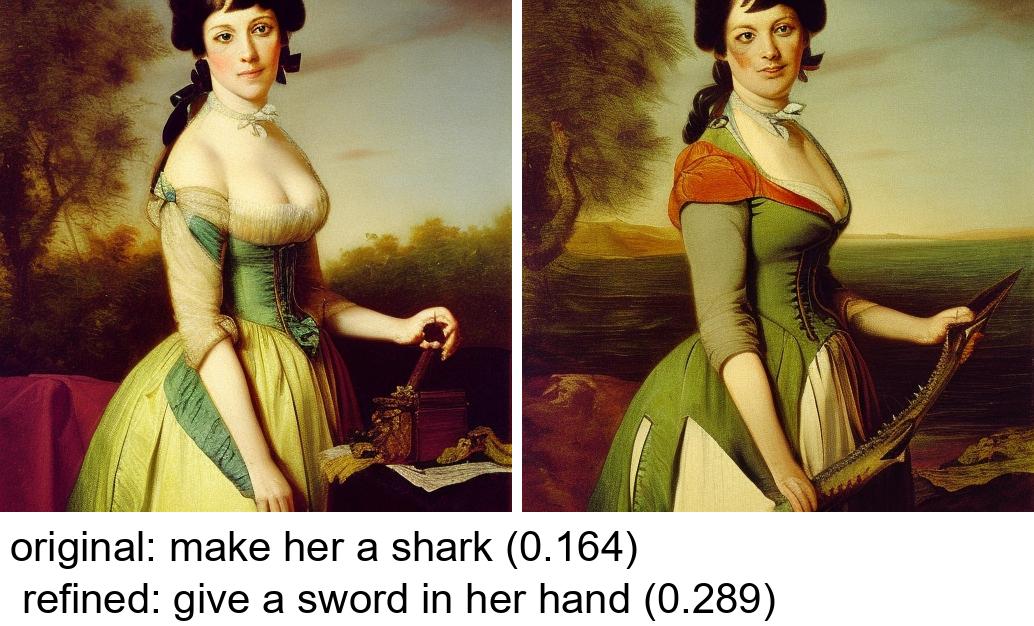}
        \parbox{\linewidth}{
        {\sc Original:} ``make her a shark'' (0.164)\\
        {\sc Refined:} ``give a sword in her hand'' (0.289)\\
        }
    \end{subfigure}
    \begin{subfigure}{0.23\linewidth}
        \includegraphics[width=\linewidth, trim=0 100px 0 0, clip]{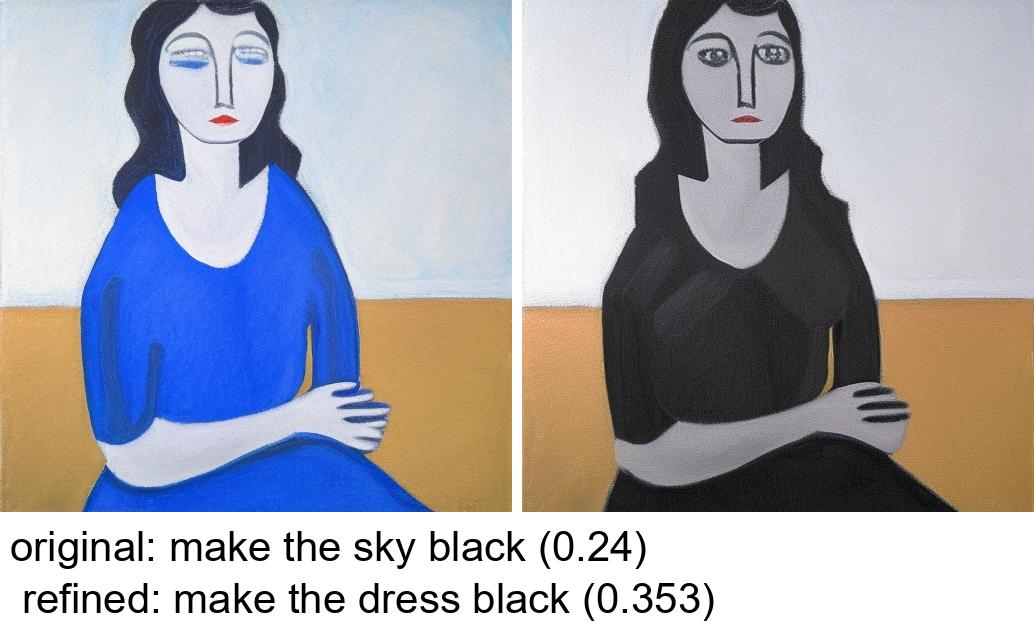}
        \parbox{\linewidth}{
        {\sc Original:} ``make the sky black'' (0.240)\\
        {\sc Refined:} ``make the dress black'' (0.353)\\
        }
    \end{subfigure}
    \begin{subfigure}{0.23\linewidth}
        \includegraphics[width=\linewidth, trim=0 100px 0 0, clip]{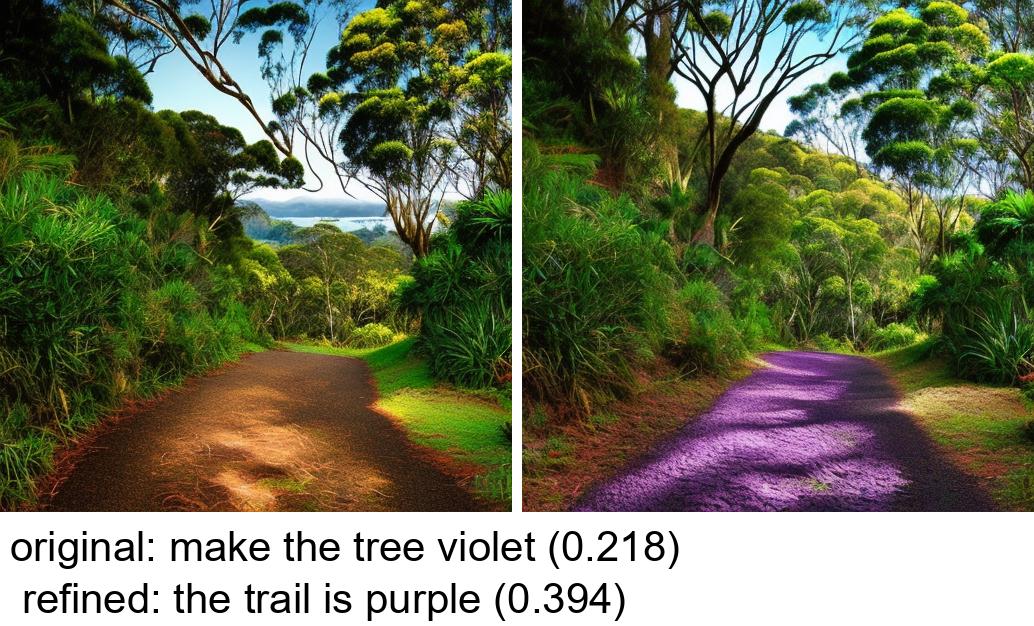}
        \parbox{\linewidth}{
        {\sc Original:} ``make the tree violet'' (0.218)\\
        {\sc Refined:} ``the trail is purple'' (0.394)\\
        }
    \end{subfigure}

    \begin{subfigure}{0.23\linewidth}
        \includegraphics[width=\linewidth, trim=0 100px 0 0, clip]{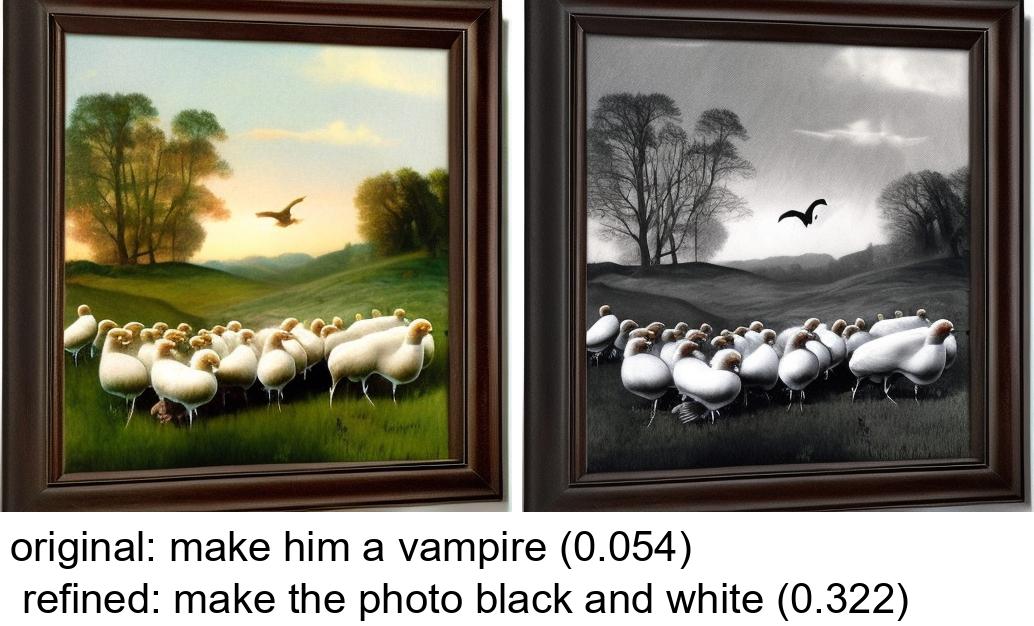}
        \parbox{\linewidth}{
        {\sc Original:} ``make him a vampire'' (0.054)\\
        {\sc Refined:} ``make the photo black and white'' (0.332)\\
        }
    \end{subfigure}
    \begin{subfigure}{0.23\linewidth}
        \includegraphics[width=\linewidth, trim=0 100px 0 0, clip]{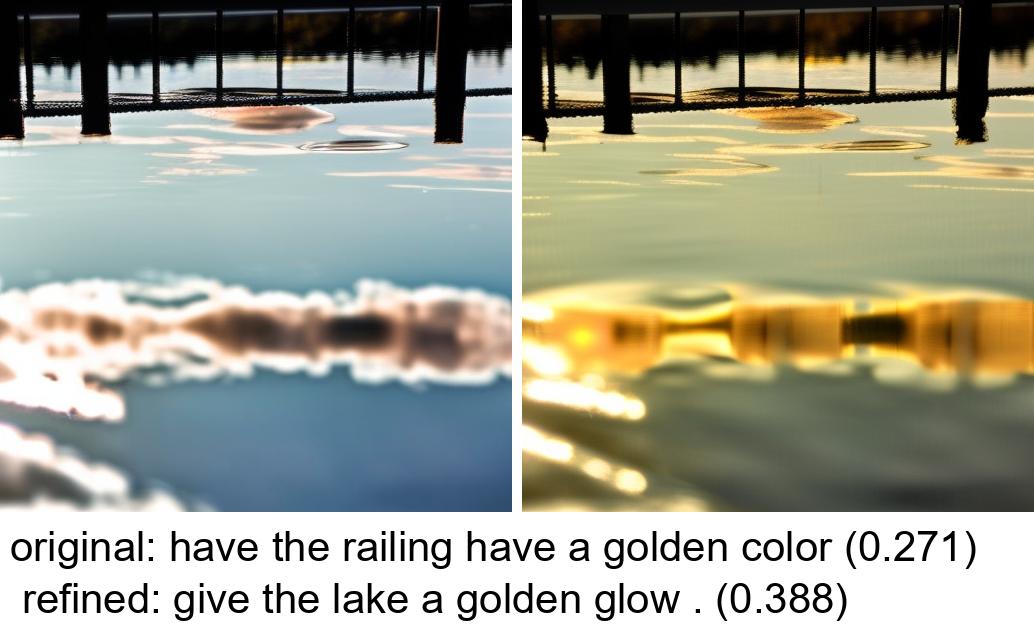}
        \parbox{\linewidth}{
        {\sc Original:} ``have the railing have a gold color'' (0.271)\\
        {\sc Refined:} ``give the lake a golden glow'' (0.388)\\
        }
    \end{subfigure}
    \begin{subfigure}{0.23\linewidth}
        \includegraphics[width=\linewidth, trim=0 100px 0 0, clip]{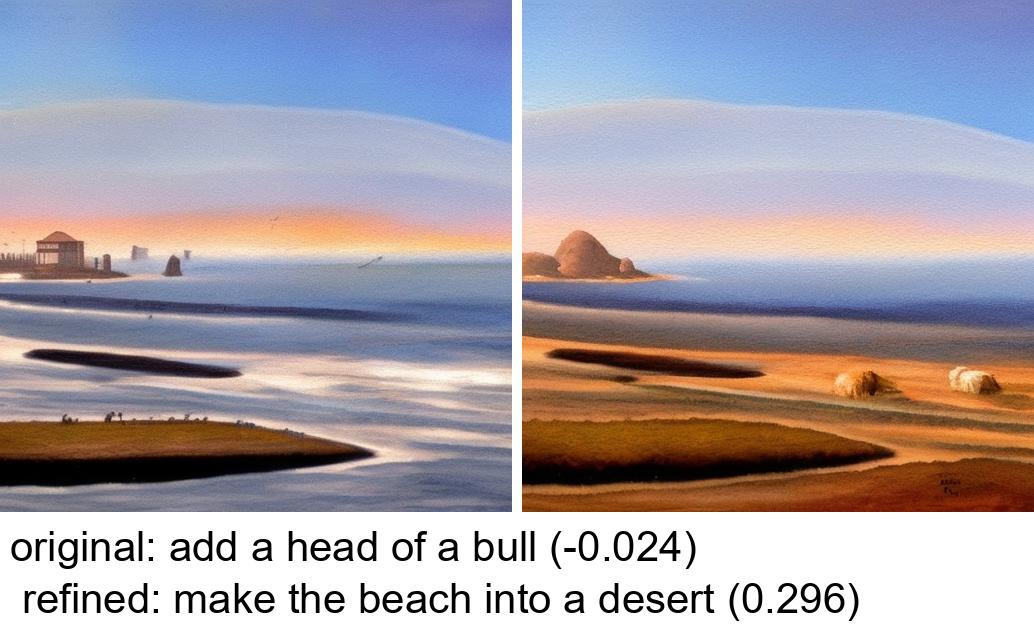}
        \parbox{\linewidth}{
        {\sc Original:} ``add a head of bull'' (-0.024)\\
        {\sc Refined:} ``make the beach into a desert'' (0.296)\\
        }
    \end{subfigure}
    \begin{subfigure}{0.23\linewidth}
        \includegraphics[width=\linewidth, trim=0 100px 0 0, clip]{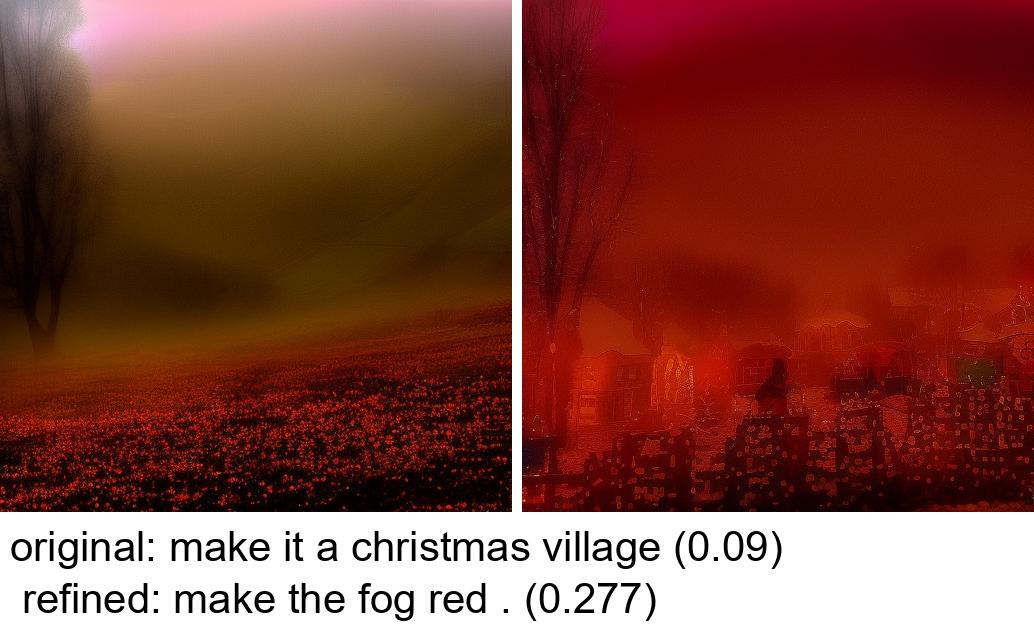}
        \parbox{\linewidth}{
        {\sc Original:} ``make it a christmas village'' (0.090)\\
        {\sc Refined:} ``make the fog red'' (0.277)\\
        }
    \end{subfigure}
    
    \caption{Additional refined instruction from our dataset (Part 1/2)}
    \label{fig:additional_dataset_samples_1}
\end{figure*}

\begin{figure*}
    \centering
    \begin{tabbing}
        \hspace{2.7em}\= \hspace{5.9em} \= \hspace{5.5em}\= \hspace{5.9em} \= \hspace{5.7em}\= \hspace{5.7em} \= \hspace{5.7em}\= \hspace{5.7em} \= \kill
        \> Original \> Edited \> Original \> Edited  \> Original \> Edited  \> Original \> Edited \\
    \end{tabbing} 
    \vspace{-0.3in}

    \tiny
    \begin{subfigure}{0.23\linewidth}
        \includegraphics[width=\linewidth, trim=0 100px 0 0, clip]{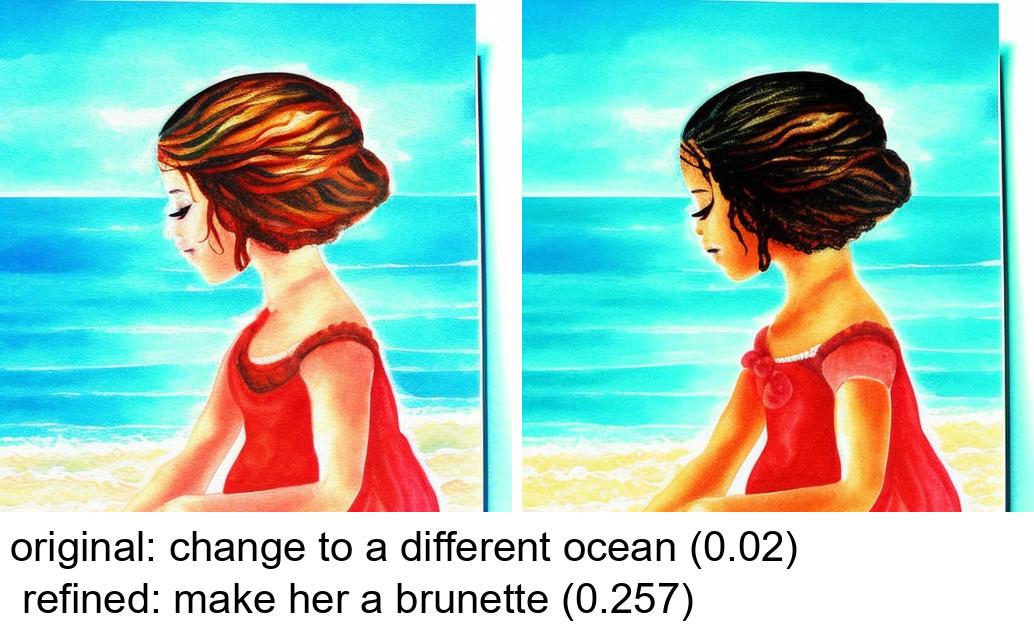}
        \parbox{\linewidth}{
        {\sc Original:} ``change to a different ocean'' (0.020)\\
        {\sc Refined:} ``make her a brunette'' (0.257)\\
        }
    \end{subfigure}
    \begin{subfigure}{0.23\linewidth}
        \includegraphics[width=\linewidth, trim=0 100px 0 0, clip]{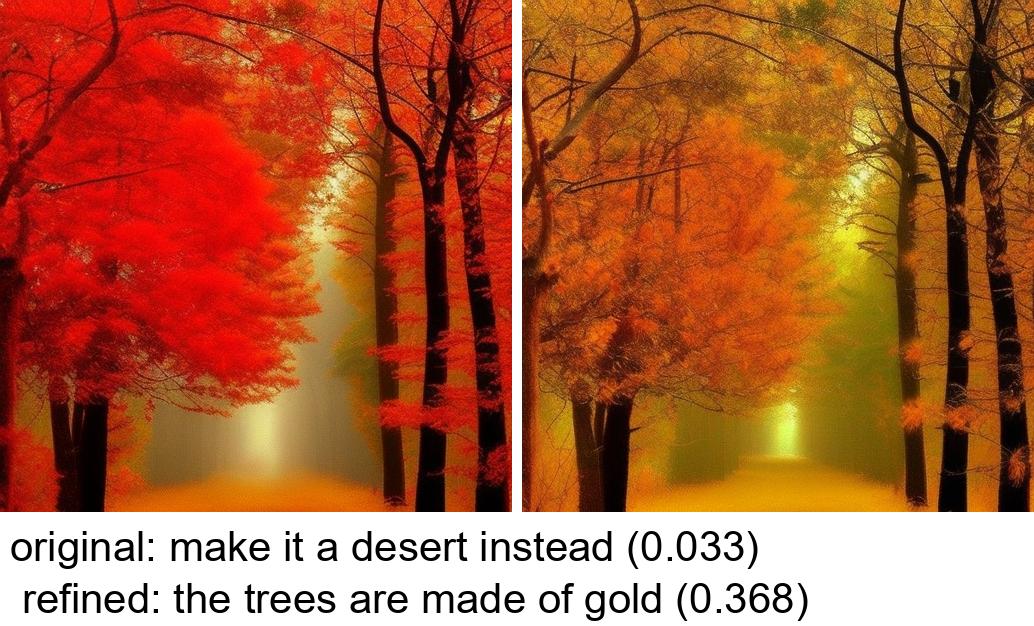}
        \parbox{\linewidth}{
        {\sc Original:} ``make it a desert instead'' (0.033)\\
        {\sc Refined:} ``the trees are made of gold'' (0.368)\\
        }
    \end{subfigure}
    \begin{subfigure}{0.23\linewidth}
        \includegraphics[width=\linewidth, trim=0 100px 0 0, clip]{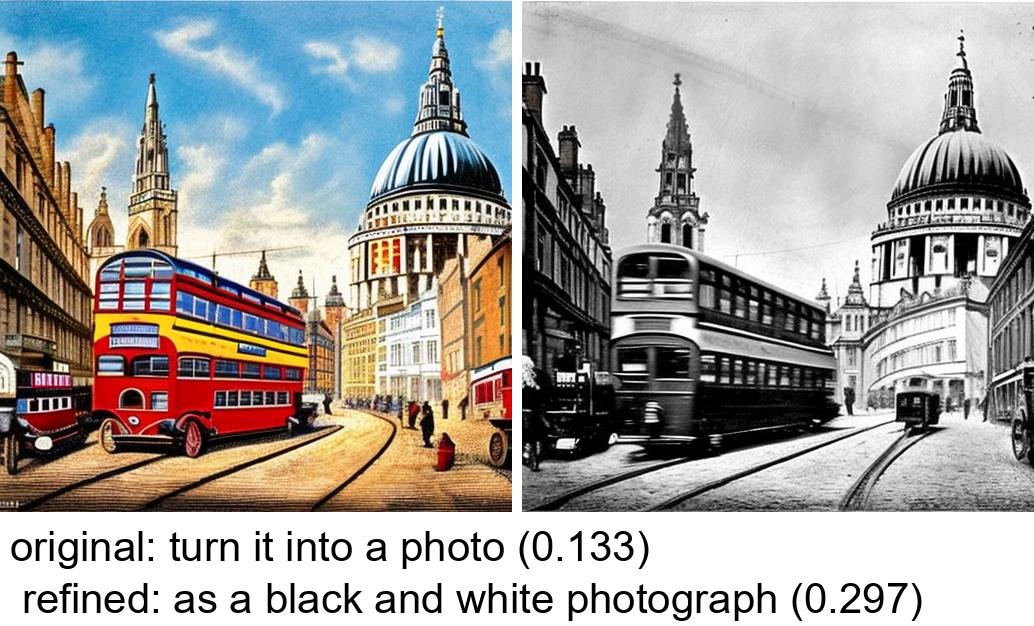}
        \parbox{\linewidth}{
        {\sc Original:} ``turn into a photo'' (0.133)\\
        {\sc Refined:} ``as a black and white photograph'' (0.297)\\
        }
    \end{subfigure}
    \begin{subfigure}{0.23\linewidth}
        \includegraphics[width=\linewidth, trim=0 100px 0 0, clip]{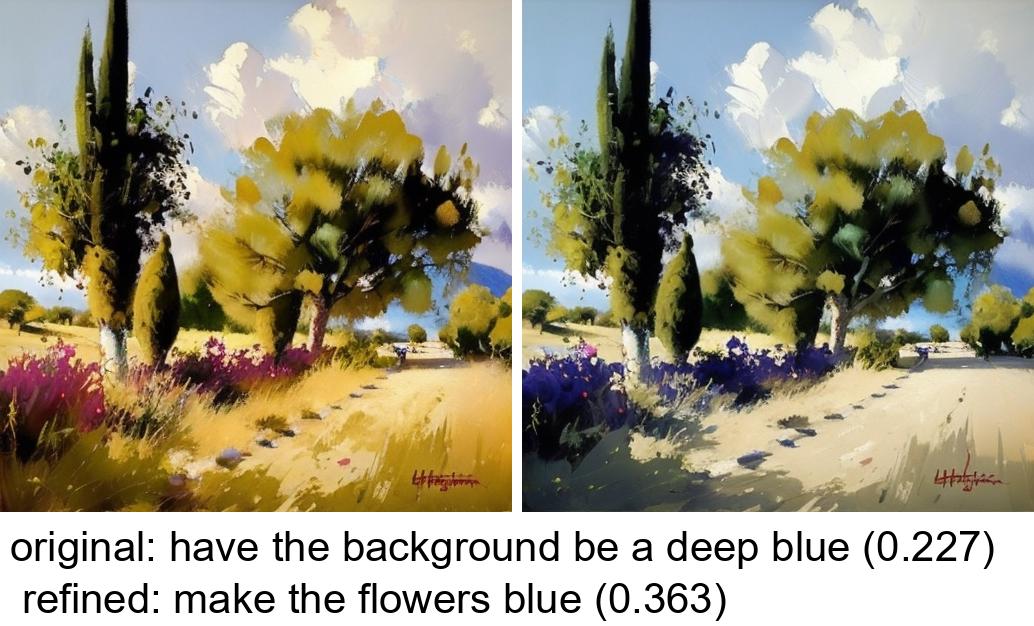}
        \parbox{\linewidth}{
        {\sc Original:} ``have the background be deep blue'' (0.227)\\
        {\sc Refined:} ``make the flowers blue'' (0.363)\\
        }
    \end{subfigure}

    \begin{subfigure}{0.23\linewidth}
        \includegraphics[width=\linewidth, trim=0 100px 0 0, clip]{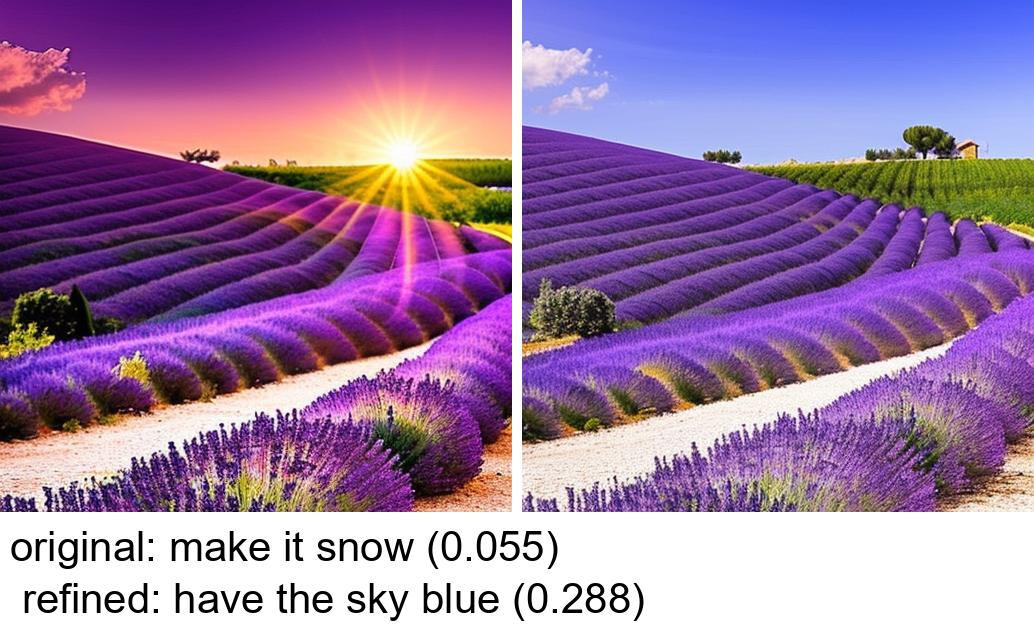}
        \parbox{\linewidth}{
        {\sc Original:} ``make it snow'' (0.055)\\
        {\sc Refined:} ``have the sky blue'' (0.288)\\
        }
    \end{subfigure}
    \begin{subfigure}{0.23\linewidth}
        \includegraphics[width=\linewidth, trim=0 100px 0 0, clip]{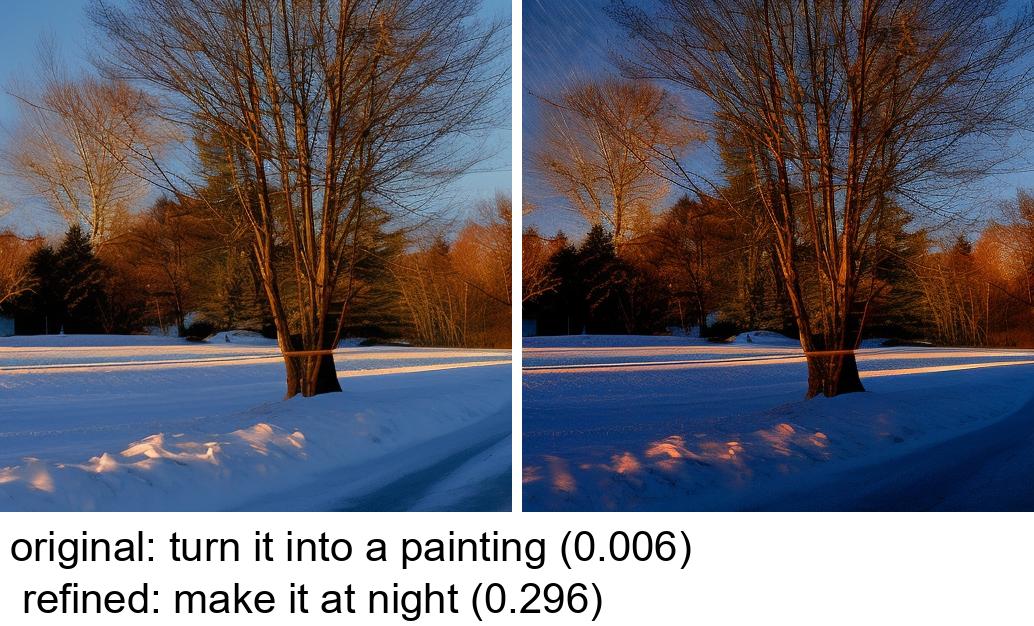}
        \parbox{\linewidth}{
        {\sc Original:} ``turn into a painting'' (0.006)\\
        {\sc Refined:} ``make it at night'' (0.296)\\
        }
    \end{subfigure}
    \begin{subfigure}{0.23\linewidth}
        \includegraphics[width=\linewidth, trim=0 100px 0 0, clip]{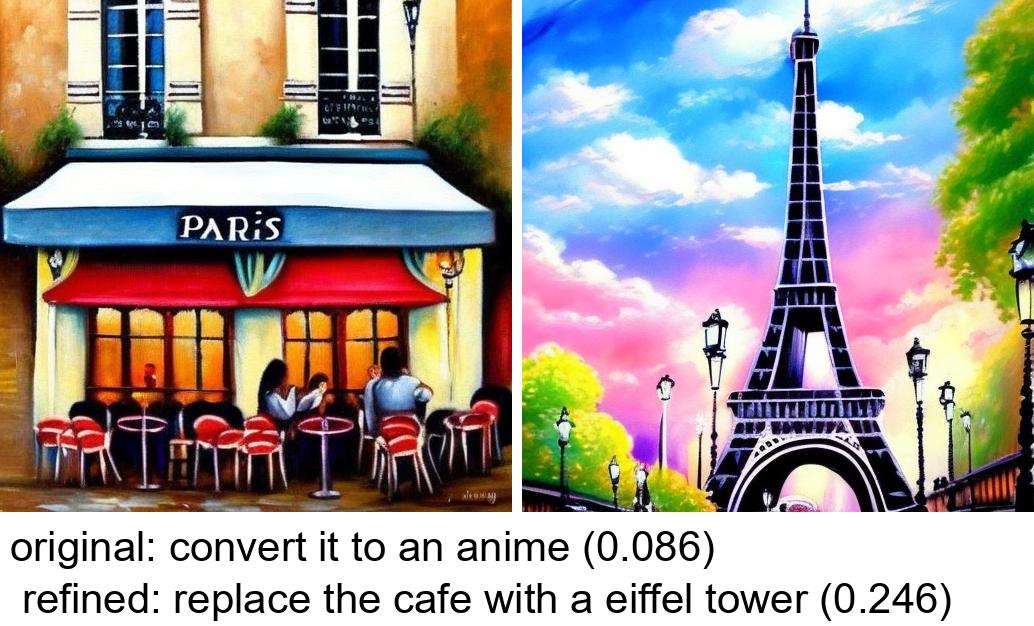}
        \parbox{\linewidth}{
        {\sc Original:} ``convert into an anime'' (0.086)\\
        {\sc Refined:} ``replace the cafe with a eiffel tower'' (0.246)\\
        }
    \end{subfigure}
    \begin{subfigure}{0.23\linewidth}
        \includegraphics[width=\linewidth, trim=0 100px 0 0, clip]{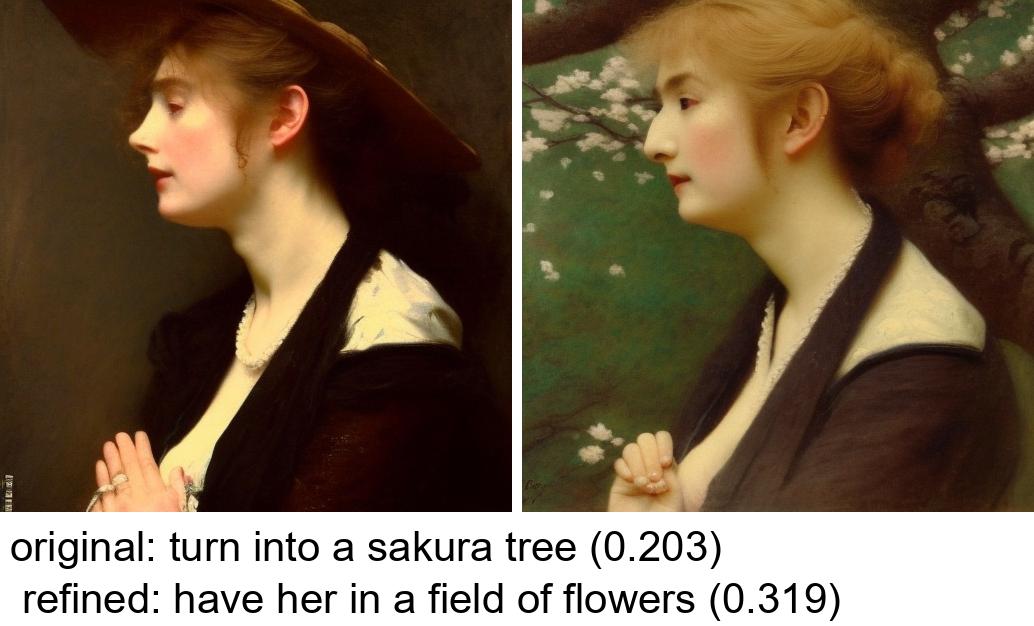}
        \parbox{\linewidth}{
        {\sc Original:} ``turn into a sakura tree'' (0.203)\\
        {\sc Refined:} ``have her in a field of flowers'' (0.319)\\
        }
    \end{subfigure}

    \begin{subfigure}{0.23\linewidth}
        \includegraphics[width=\linewidth, trim=0 100px 0 0, clip]{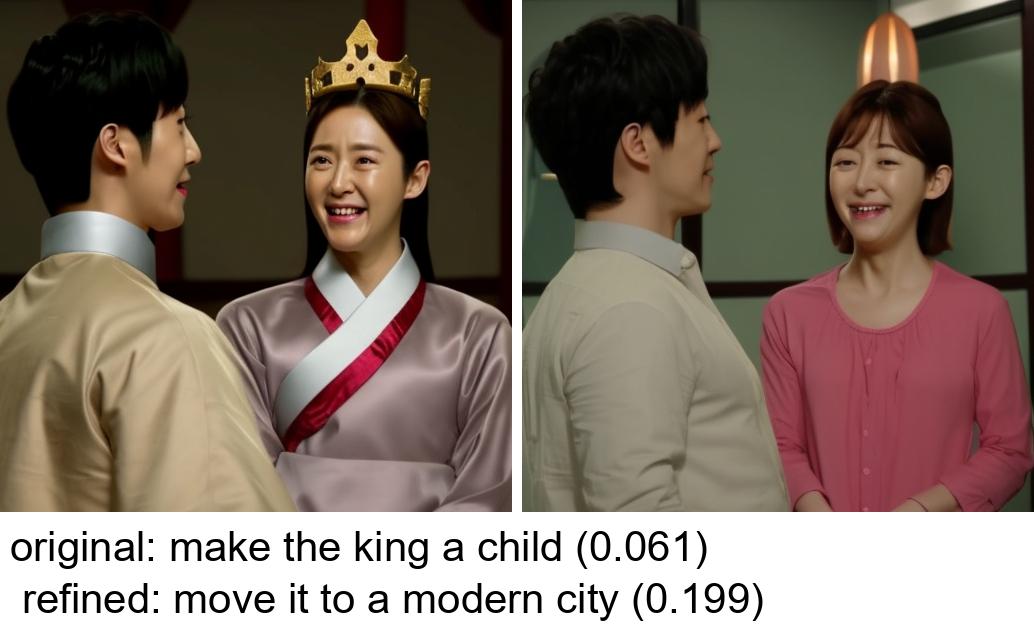}
        \parbox{\linewidth}{
        {\sc Original:} ``make the king a child'' (0.061)\\
        {\sc Refined:} ``move it to a modern city'' (0.199)\\
        }
    \end{subfigure}
    \begin{subfigure}{0.23\linewidth}
        \includegraphics[width=\linewidth, trim=0 100px 0 0, clip]{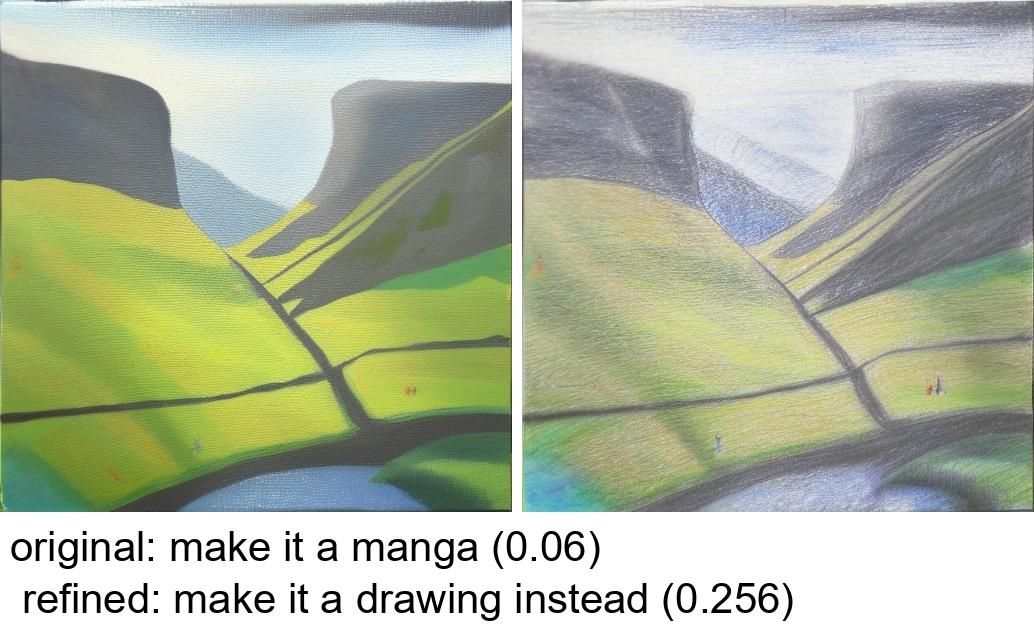}
        \parbox{\linewidth}{
        {\sc Original:} ``make it a mango'' (0.060)\\
        {\sc Refined:} ``make it a drawing instead'' (0.256)\\
        }
    \end{subfigure}
    \begin{subfigure}{0.23\linewidth}
        \includegraphics[width=\linewidth, trim=0 100px 0 0, clip]{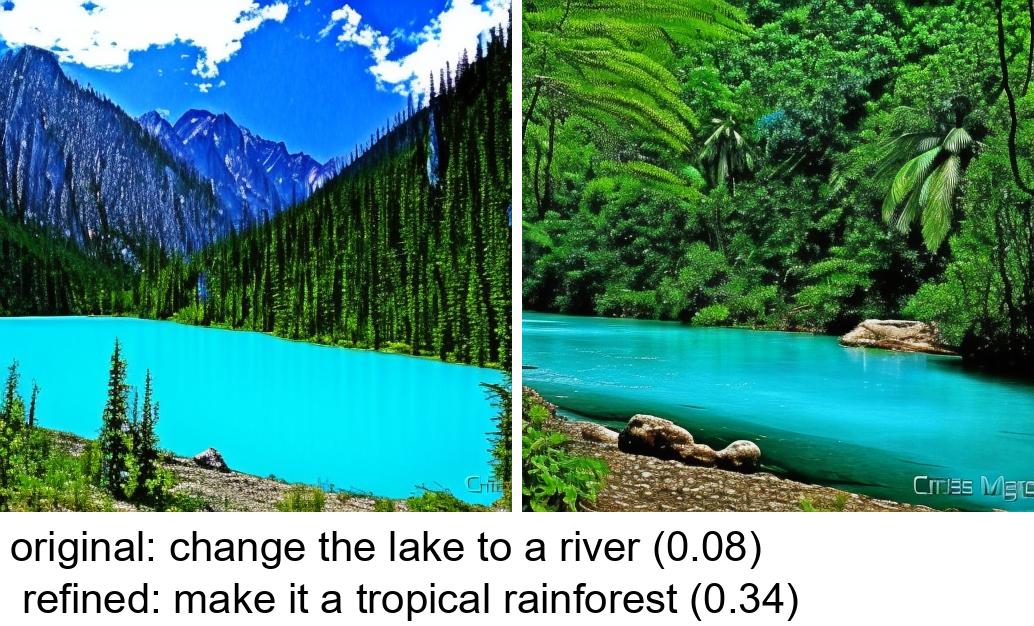}
        \parbox{\linewidth}{
        {\sc Original:} ``change the lake to a river'' (0.080)\\
        {\sc Refined:} ``make it a tropical rainforest'' (0.340)\\
        }
    \end{subfigure}
    \begin{subfigure}{0.23\linewidth}
        \includegraphics[width=\linewidth, trim=0 100px 0 0, clip]{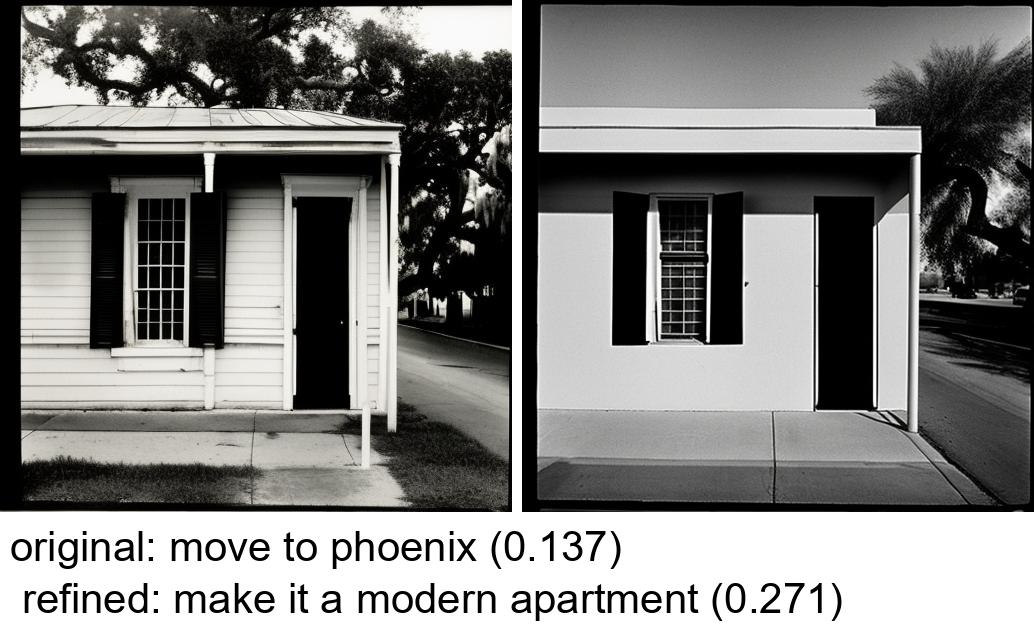}
        \parbox{\linewidth}{
        {\sc Original:} ``move to phoenix'' (0.137)\\
        {\sc Refined:} `` make it a modern apartment'' (0.271)\\
        }
    \end{subfigure}
    
    \begin{subfigure}{0.23\linewidth}
        \includegraphics[width=\linewidth, trim=0 100px 0 0, clip]{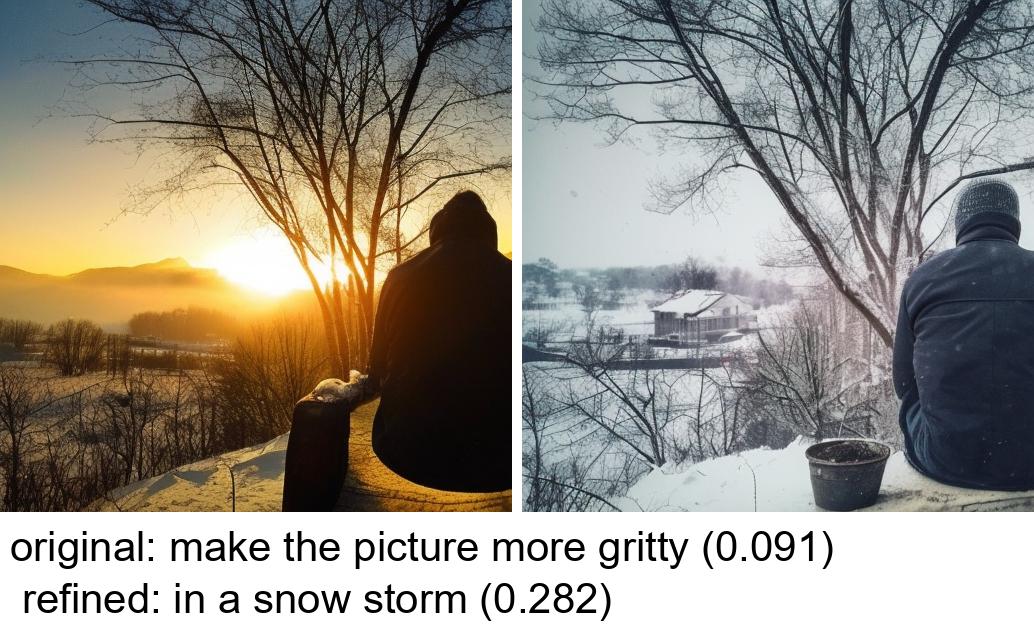}
        \parbox{\linewidth}{
        {\sc Original:} ``make the picture more gritty'' (0.091)\\
        {\sc Refined:} ``in a snow storm'' (0.282)\\
        }
    \end{subfigure}
    \begin{subfigure}{0.23\linewidth}
        \includegraphics[width=\linewidth, trim=0 100px 0 0, clip]{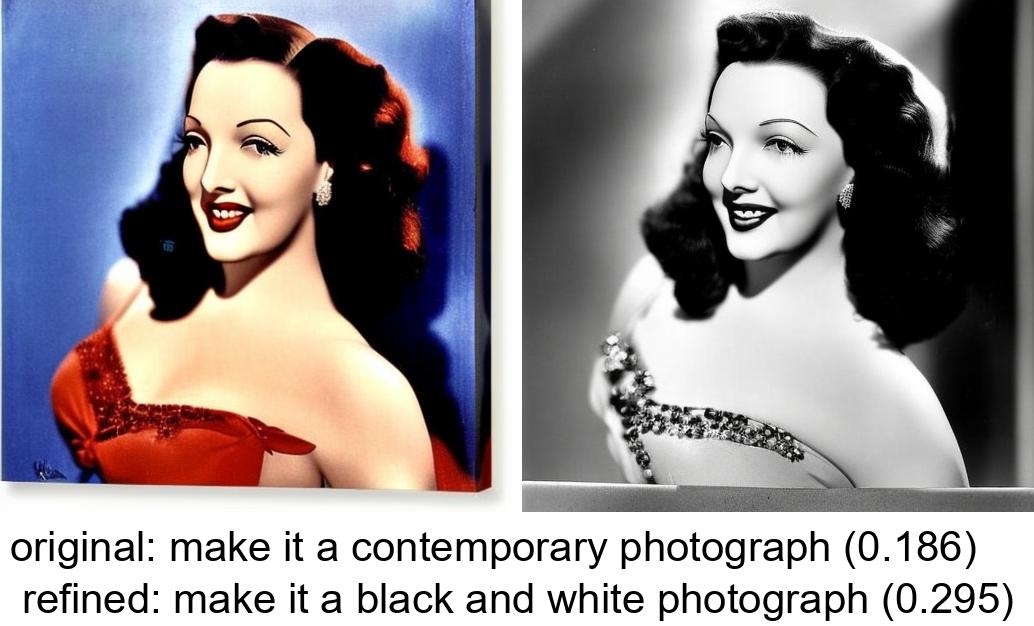}
        \parbox{\linewidth}{
        {\sc Original:} ``make it contemporary photograph'' (0.186)\\
        {\sc Refined:} ``make it black and white photograph'' (0.295)\\
        }
    \end{subfigure}
    \begin{subfigure}{0.23\linewidth}
        \includegraphics[width=\linewidth, trim=0 100px 0 0, clip]{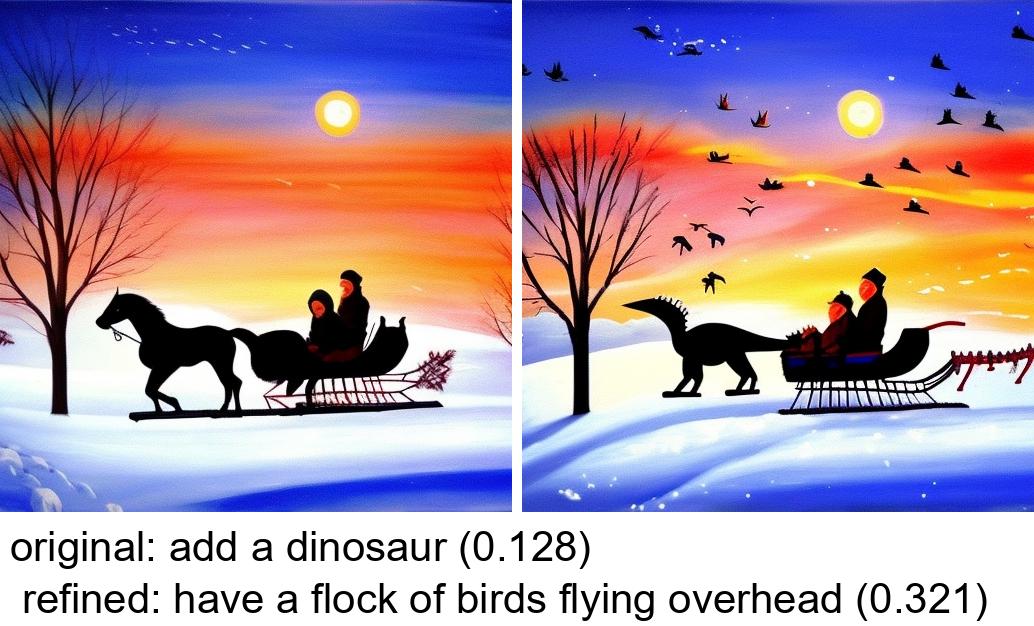}
        \parbox{\linewidth}{
        {\sc Original:} ``add a dinosaur'' (0.128)\\
        {\sc Refined:} ``have a flock of birds flying overhead'' (0.321)\\
        }
    \end{subfigure}
    \begin{subfigure}{0.23\linewidth}
        \includegraphics[width=\linewidth, trim=0 100px 0 0, clip]{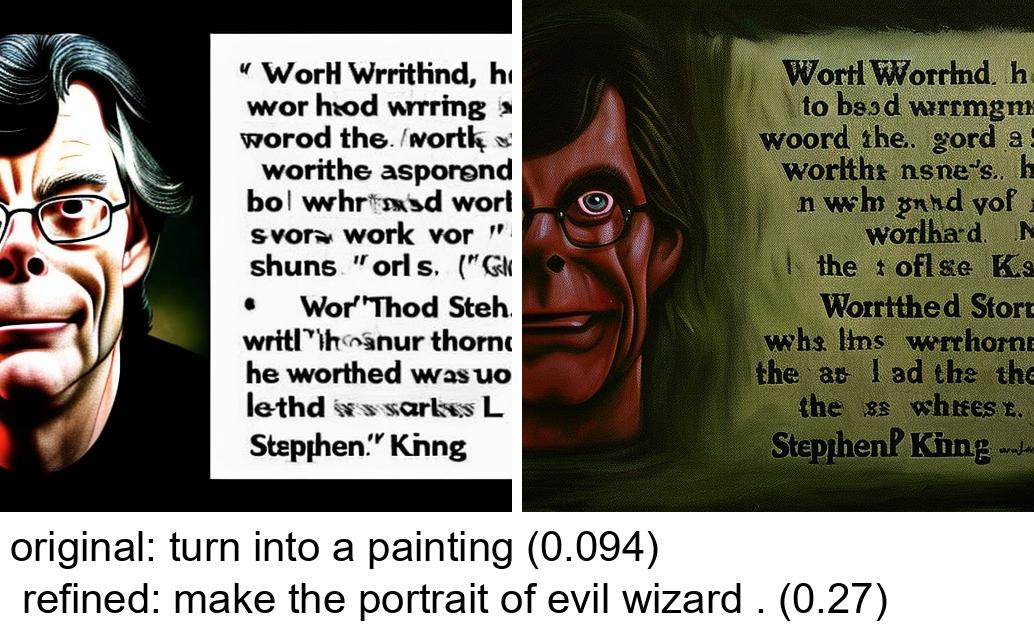}
        \parbox{\linewidth}{
        {\sc Original:} ``turn into a painting'' (0.094)\\
        {\sc Refined:} ``make the portrait of evil wizard'' (0.270)\\
        }
    \end{subfigure}

    \begin{subfigure}{0.23\linewidth}
        \includegraphics[width=\linewidth, trim=0 100px 0 0, clip]{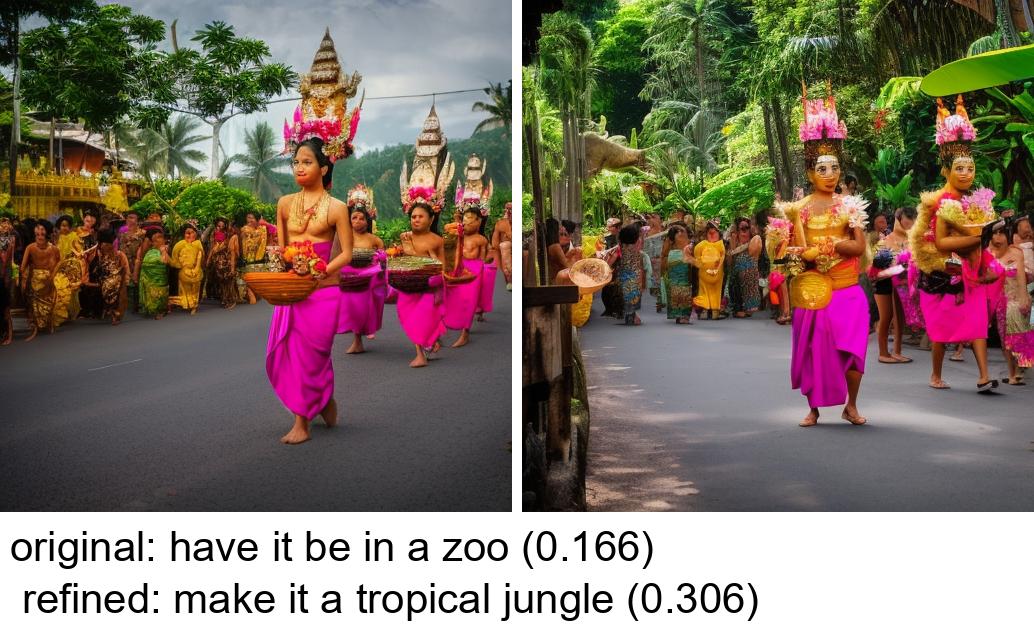}
        \parbox{\linewidth}{
        {\sc Original:} ``have it be in a zoo'' (0.166)\\
        {\sc Refined:} ``make it a tropical jungle'' (0.306)\\
        }
    \end{subfigure}
    \begin{subfigure}{0.23\linewidth}
        \includegraphics[width=\linewidth, trim=0 100px 0 0, clip]{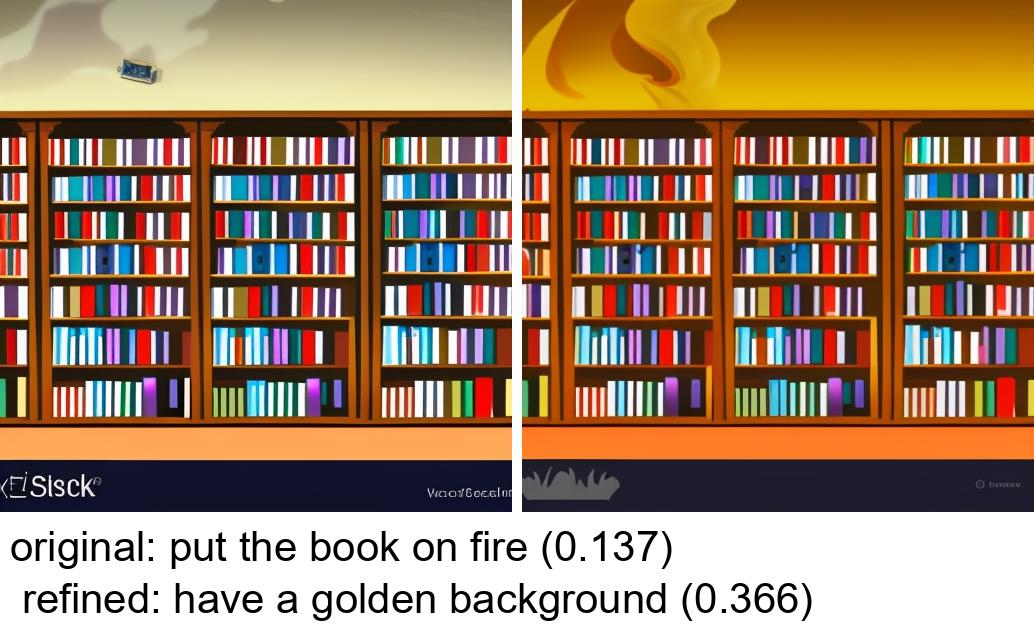}
        \parbox{\linewidth}{
        {\sc Original:} ``put the book on fire'' (0.137)\\
        {\sc Refined:} ``have a golden background'' (0.366)\\
        }
    \end{subfigure}
    \begin{subfigure}{0.23\linewidth}
        \includegraphics[width=\linewidth, trim=0 100px 0 0, clip]{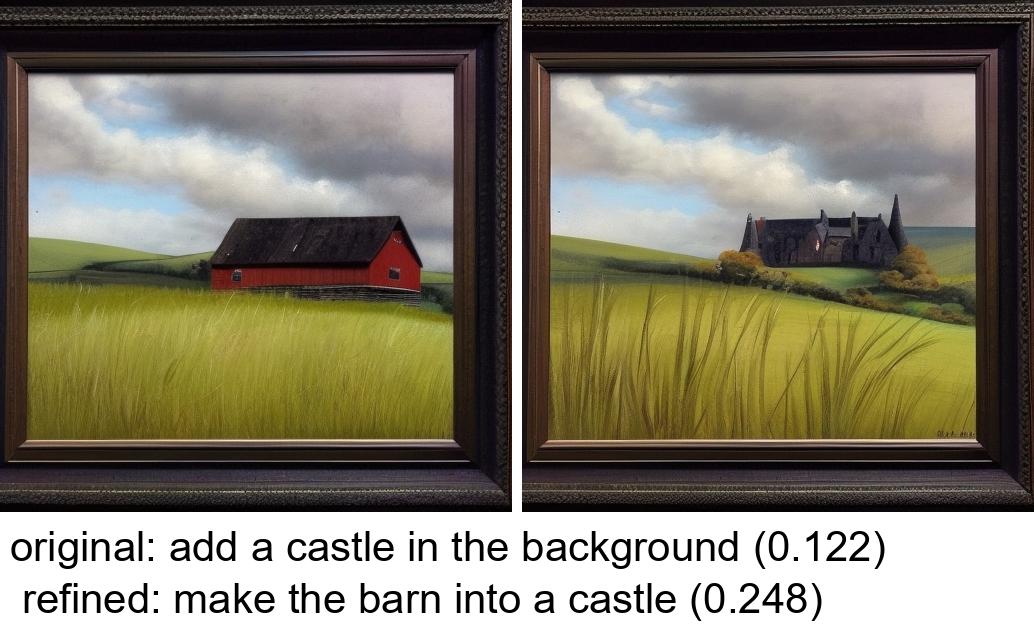}
        \parbox{\linewidth}{
        {\sc Original:} ``add a castle in the background'' (0.122)\\
        {\sc Refined:} ``make the barn into a castle'' (0.248)\\
        }
    \end{subfigure}
    \begin{subfigure}{0.23\linewidth}
        \includegraphics[width=\linewidth, trim=0 100px 0 0, clip]{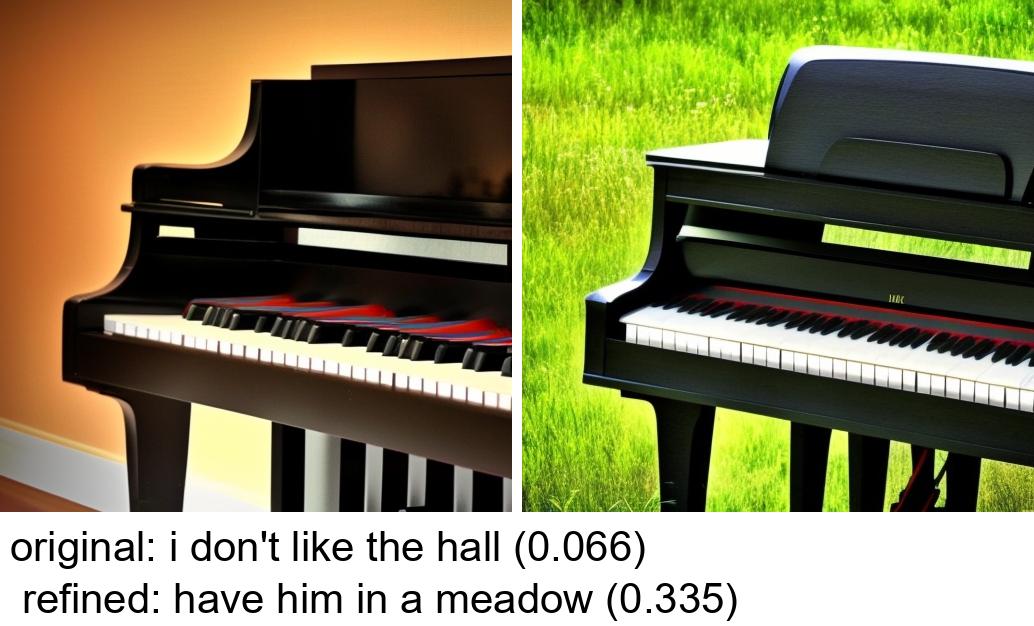}
        \parbox{\linewidth}{
        {\sc Original:} ``i don't like the hall'' (0.066)\\
        {\sc Refined:} ``have him in a meadow'' (0.335)\\
        }
    \end{subfigure}

    \begin{subfigure}{0.23\linewidth}
        \includegraphics[width=\linewidth, trim=0 100px 0 0, clip]{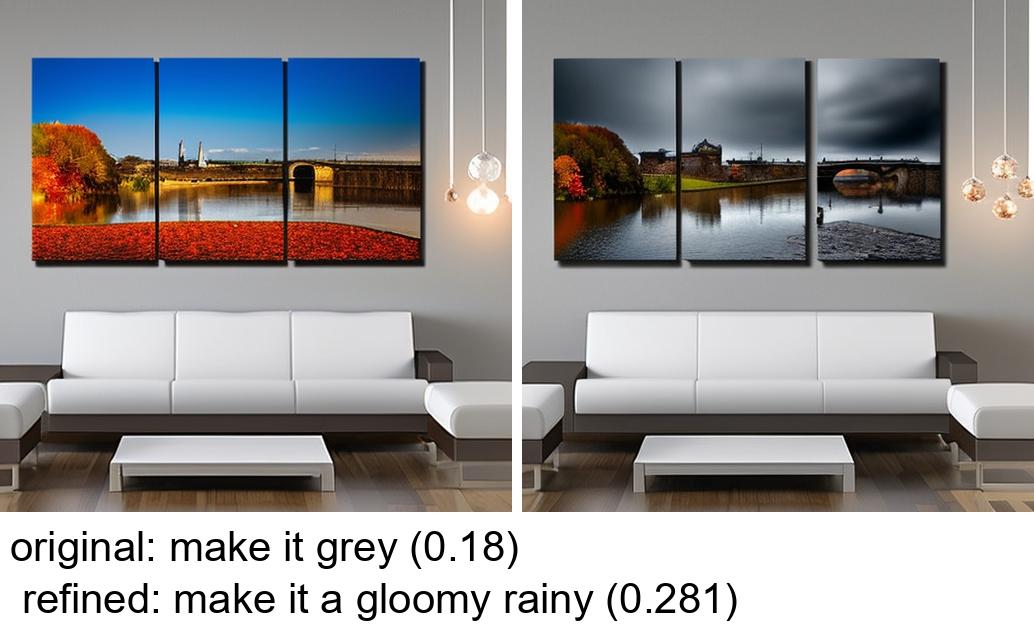}
        \parbox{\linewidth}{
        {\sc Original:} ``make it grey'' (0.180)\\
        {\sc Refined:} ``make it a gloomy rainy'' (0.281)\\
        }
    \end{subfigure}
    \begin{subfigure}{0.23\linewidth}
        \includegraphics[width=\linewidth, trim=0 100px 0 0, clip]{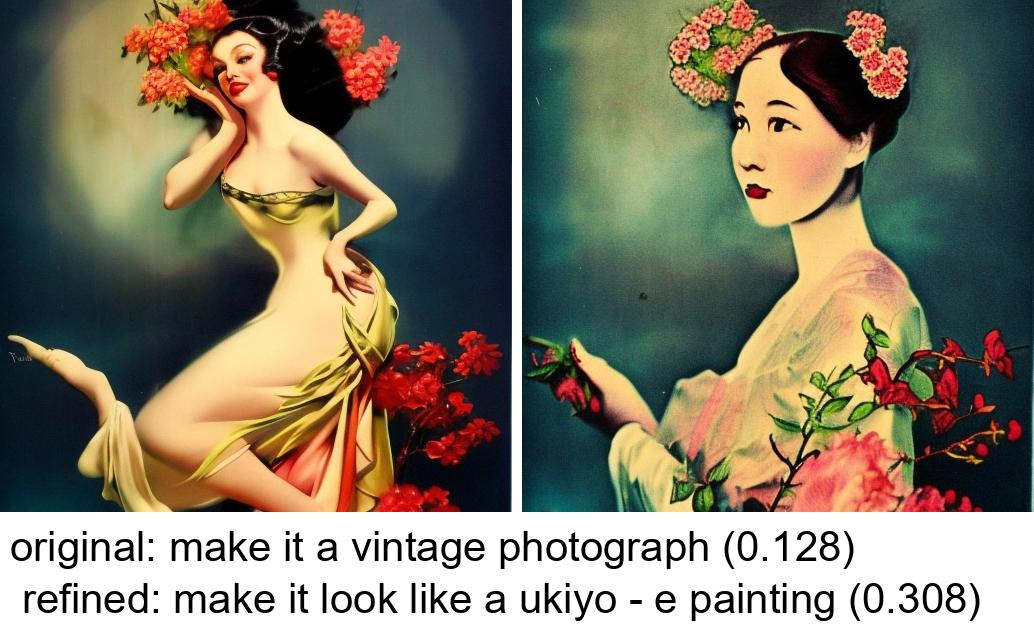}
        \parbox{\linewidth}{
        {\sc Original:} ``make it a vintage photograph'' (0.128)\\
        {\sc Refined:} ``make it look like a ukiyo - e painting'' (0.308)\\
        }
    \end{subfigure}
    \begin{subfigure}{0.23\linewidth}
        \includegraphics[width=\linewidth, trim=0 100px 0 0, clip]{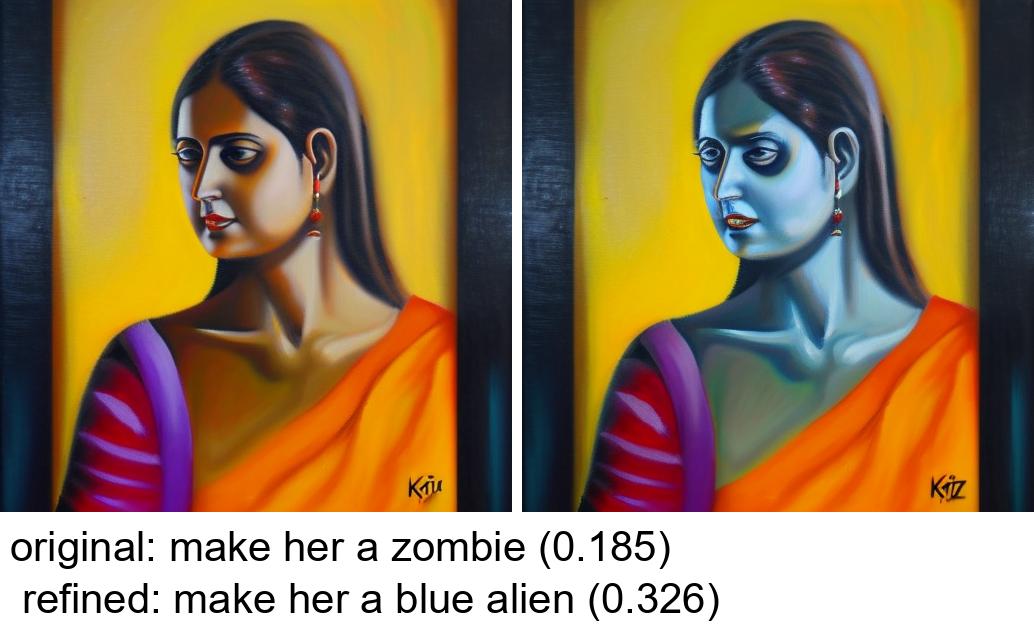}
        \parbox{\linewidth}{
        {\sc Original:} ``make her a zombie'' (0.185)\\
        {\sc Refined:} ``make her a blue alien'' (0.326)\\
        }
    \end{subfigure}
    \begin{subfigure}{0.23\linewidth}
        \includegraphics[width=\linewidth, trim=0 100px 0 0, clip]{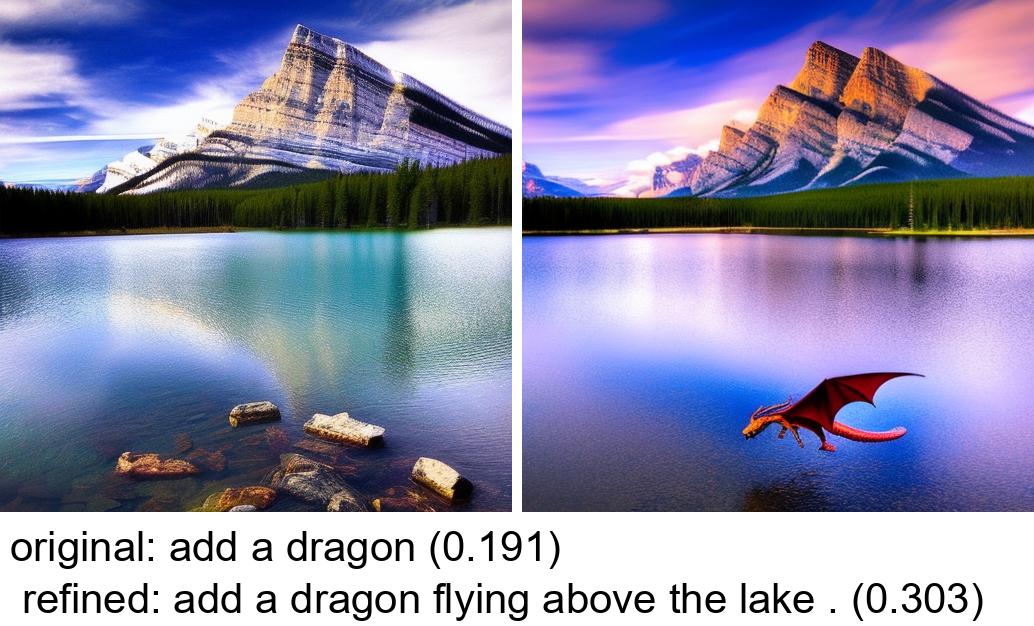}
        \parbox{\linewidth}{
        {\sc Original:} ``add a dragon'' (0.191)\\
        {\sc Refined:} ``add a dragon flying above the lake'' (0.303)\\
        }
    \end{subfigure}

    \begin{subfigure}{0.23\linewidth}
        \includegraphics[width=\linewidth, trim=0 100px 0 0, clip]{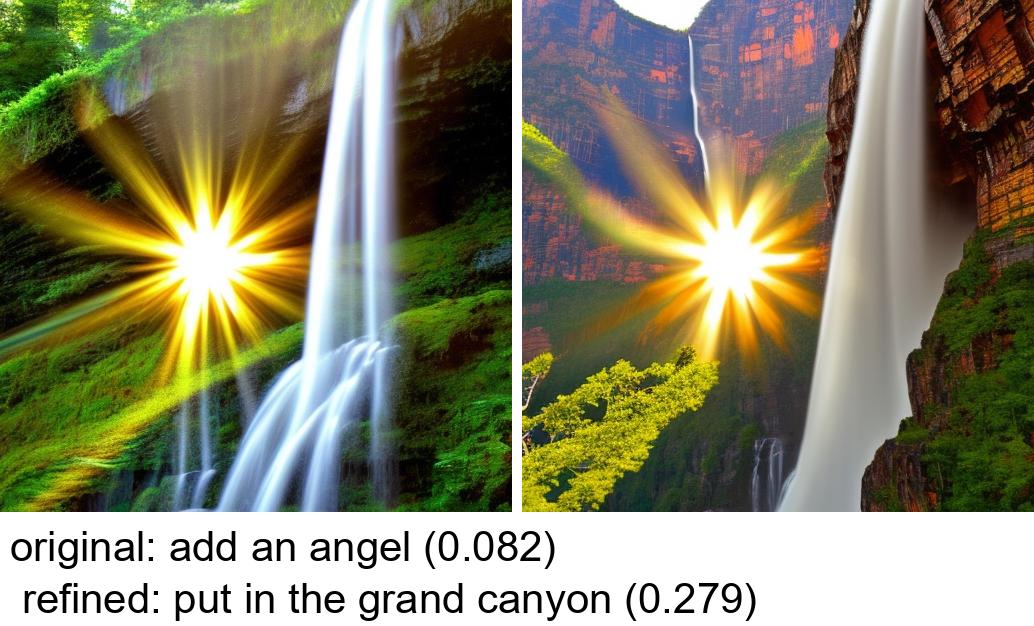}
        \parbox{\linewidth}{
        {\sc Original:} ``add an angel'' (0.082)\\
        {\sc Refined:} ``put in the grand canyon'' (0.279)\\
        }
    \end{subfigure}
    \begin{subfigure}{0.23\linewidth}
        \includegraphics[width=\linewidth, trim=0 100px 0 0, clip]{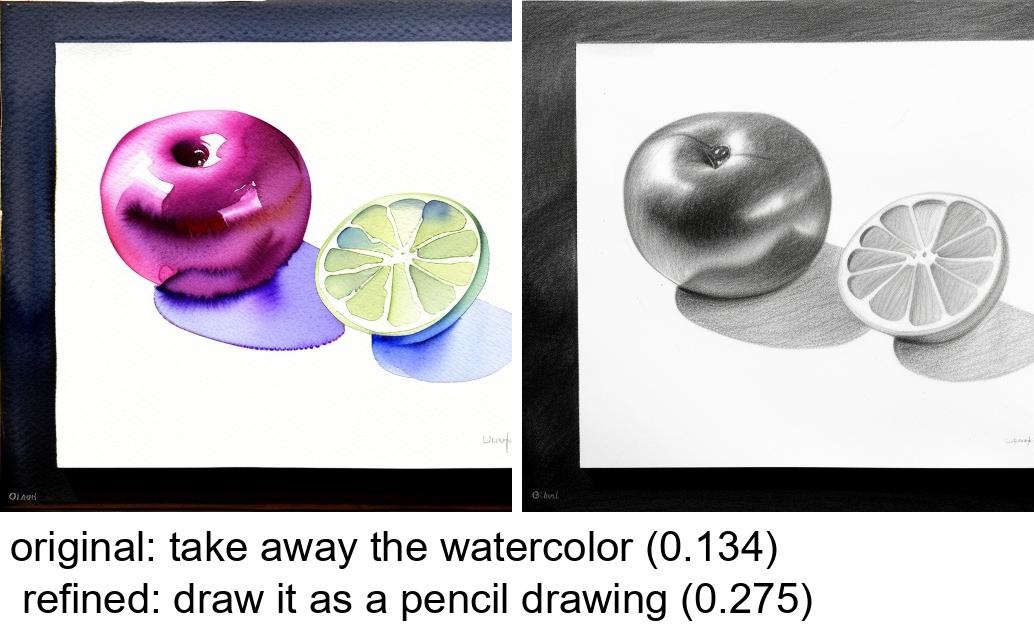}
        \parbox{\linewidth}{
        {\sc Original:} ``take away the watercolor'' (0.134)\\
        {\sc Refined:} ``draw as a pencil drawing'' (0.275)\\
        }
    \end{subfigure}
    \begin{subfigure}{0.23\linewidth}
        \includegraphics[width=\linewidth, trim=0 100px 0 0, clip]{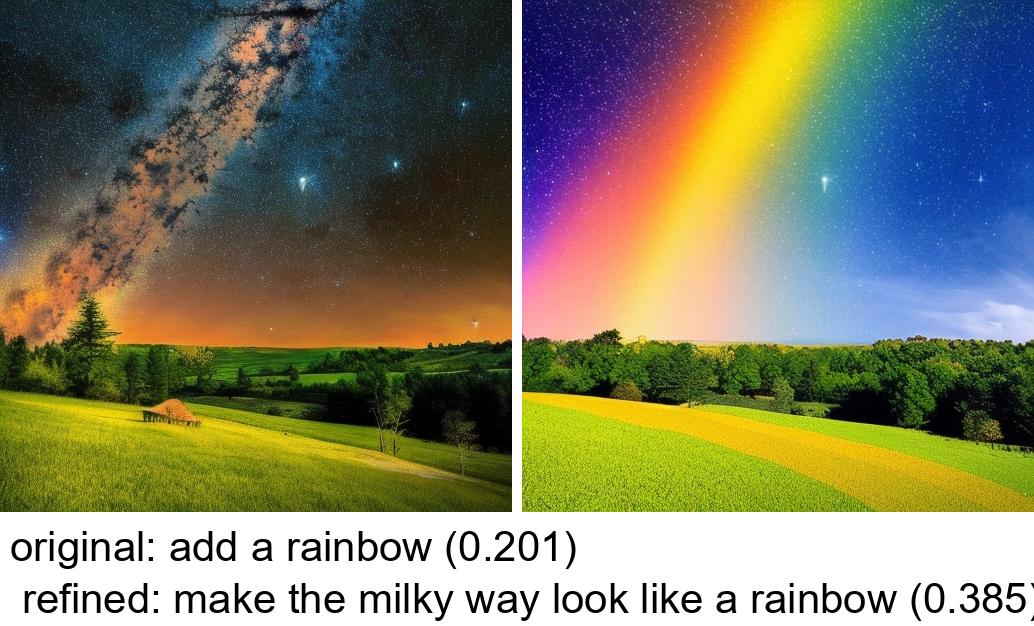}
        \parbox{\linewidth}{
        {\sc Original:} ``add a rainbow'' (0.201)\\
        {\sc Refined:} ``make milky way look like rainbow'' (0.385)\\
        }
    \end{subfigure}
    \begin{subfigure}{0.23\linewidth}
        \includegraphics[width=\linewidth, trim=0 100px 0 0, clip]{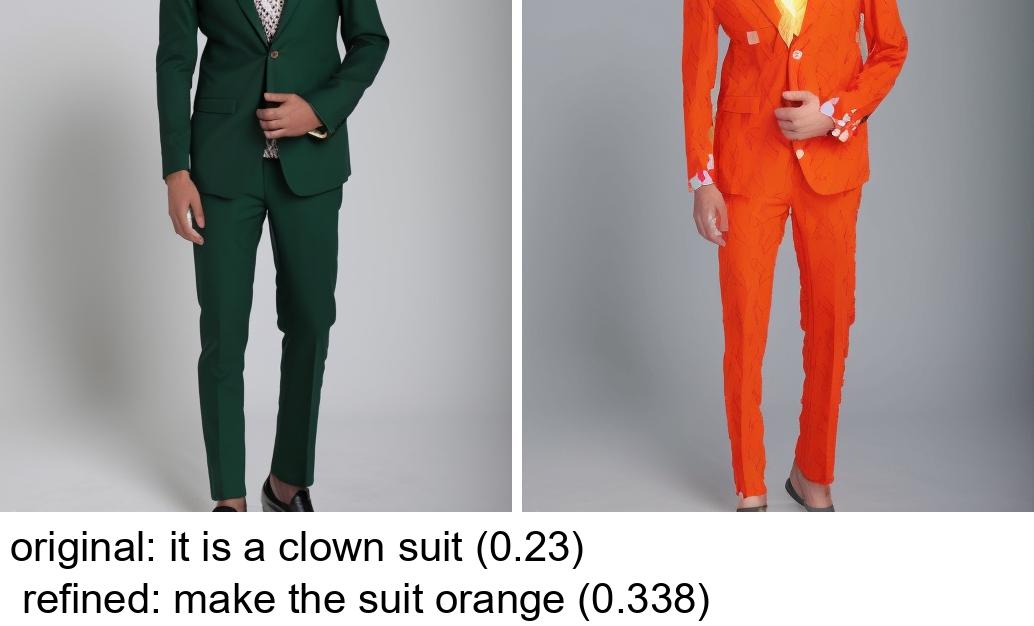}
        \parbox{\linewidth}{
        {\sc Original:} ``it is a clawn suit'' (0.230)\\
        {\sc Refined:} ``make the suit orange'' (0.338)\\
        }
    \end{subfigure}

    \begin{subfigure}{0.23\linewidth}
        \includegraphics[width=\linewidth, trim=0 100px 0 0, clip]{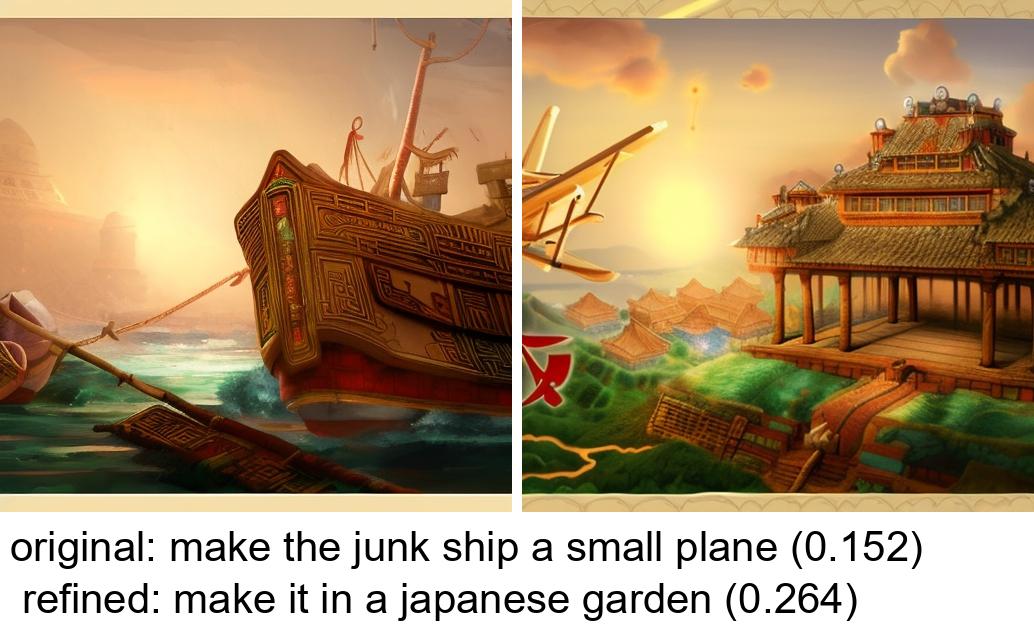}
        \parbox{\linewidth}{
        {\sc Original:} ``make the junk ship a small plane'' (0.152)\\
        {\sc Refined:} `` make it in a japanese garden'' (0.264)\\
        }
    \end{subfigure}
    \begin{subfigure}{0.23\linewidth}
        \includegraphics[width=\linewidth, trim=0 100px 0 0, clip]{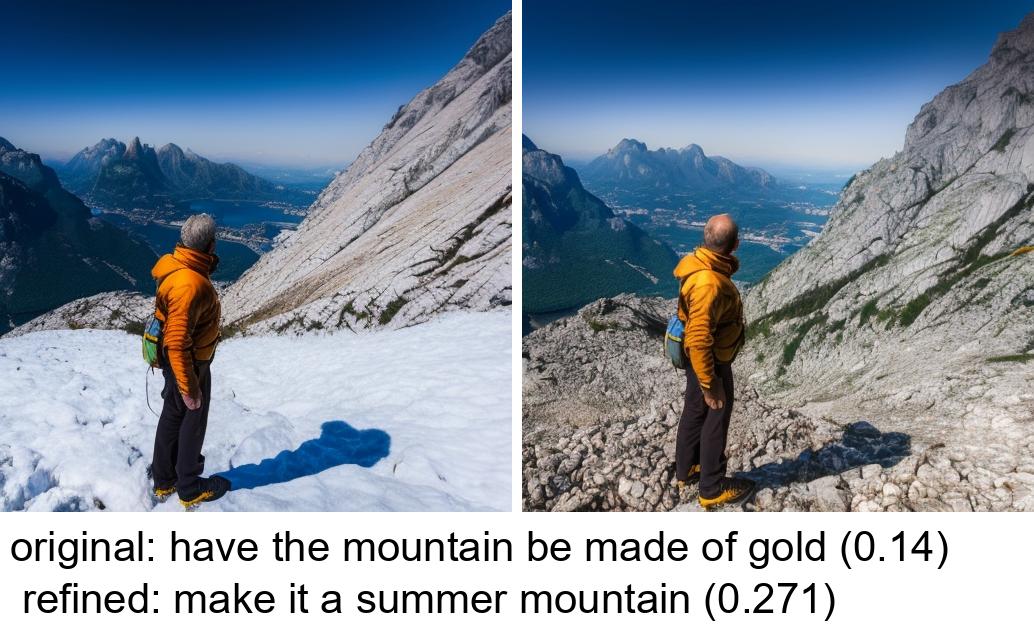}
        \parbox{\linewidth}{
        {\sc Original:} ``have the mountain be made of gold'' (0.140)\\
        {\sc Refined:} ``make it a summer mountain'' (0.271)\\
        }
    \end{subfigure}
    \begin{subfigure}{0.23\linewidth}
        \includegraphics[width=\linewidth, trim=0 100px 0 0, clip]{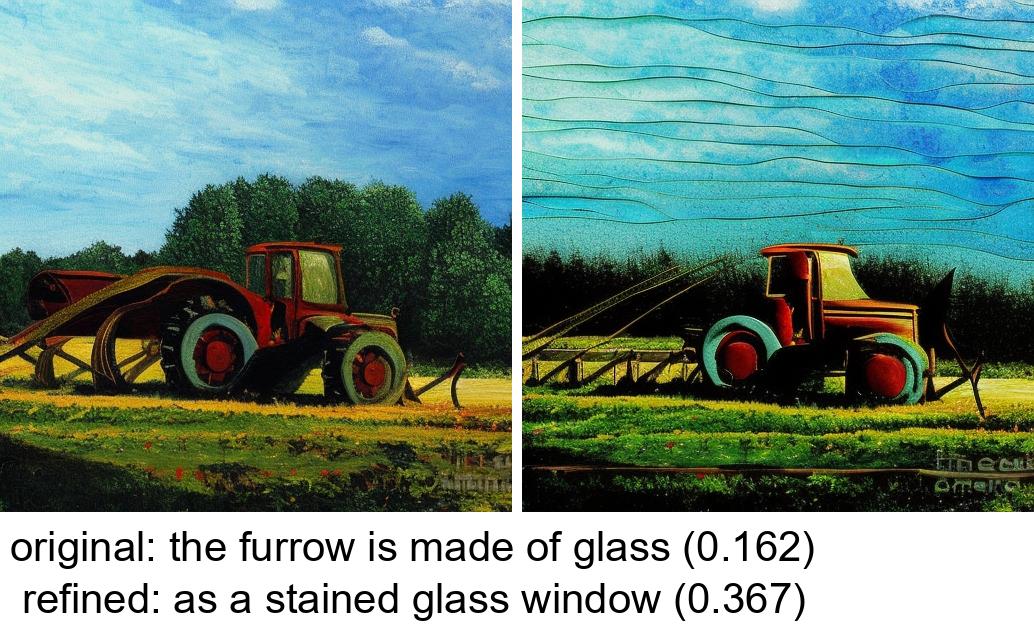}
        \parbox{\linewidth}{
        {\sc Original:} ``the furrow is made of glass'' (0.162)\\
        {\sc Refined:} ``as a stained glass window'' (0.367)\\
        }
    \end{subfigure}
    \begin{subfigure}{0.23\linewidth}
        \includegraphics[width=\linewidth, trim=0 100px 0 0, clip]{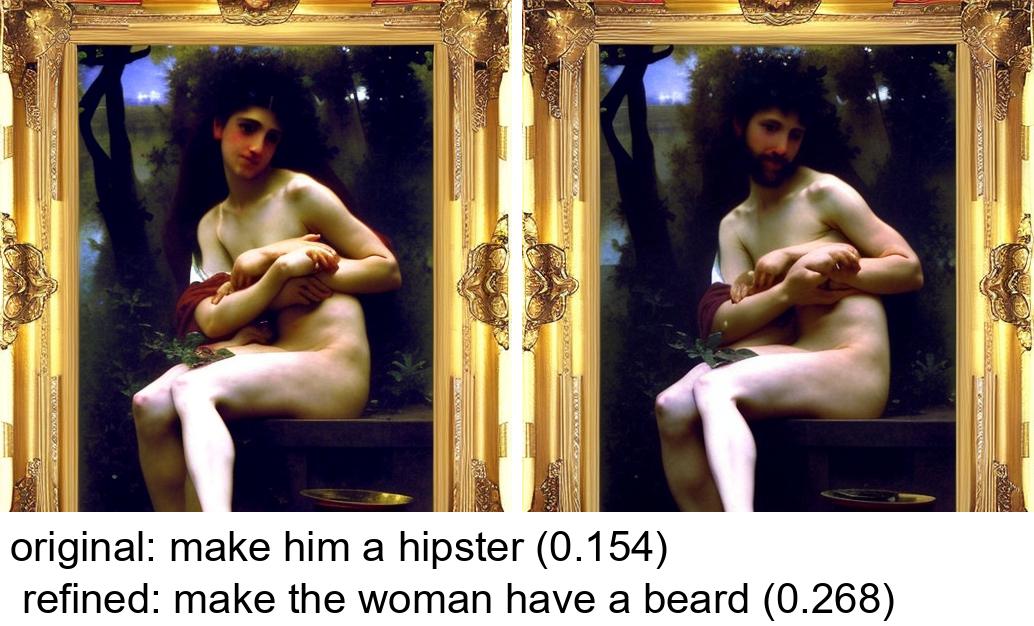}
        \parbox{\linewidth}{
        {\sc Original:} ``make him a hipster'' (0.154)\\
        {\sc Refined:} ``make the woman have a beard'' (0.268)\\
        }
    \end{subfigure}
    
    \caption{Additional refined instruction from our dataset (Part 2/2)}
    \label{fig:additional_dataset_samples_2}
\end{figure*}

\begin{figure*}
    \centering
    \begin{tabbing}
        \hspace{6.3em}\= \hspace{7.0em} \= \hspace{4.8em} \= \hspace{7.9em}\= \hspace{7.0em} \= \hspace{4.7em} \= \kill
        \> Original \> IP2P \> \ourworkabbr{} (Ours) \> Original \> IP2P \> \ourworkabbr{} (Ours)  \\
    \end{tabbing}    
    \vspace{-0.32in}

    \begin{subfigure}{0.40\linewidth}
        \includegraphics[width=\linewidth, trim=0 25px 0 25px, clip]{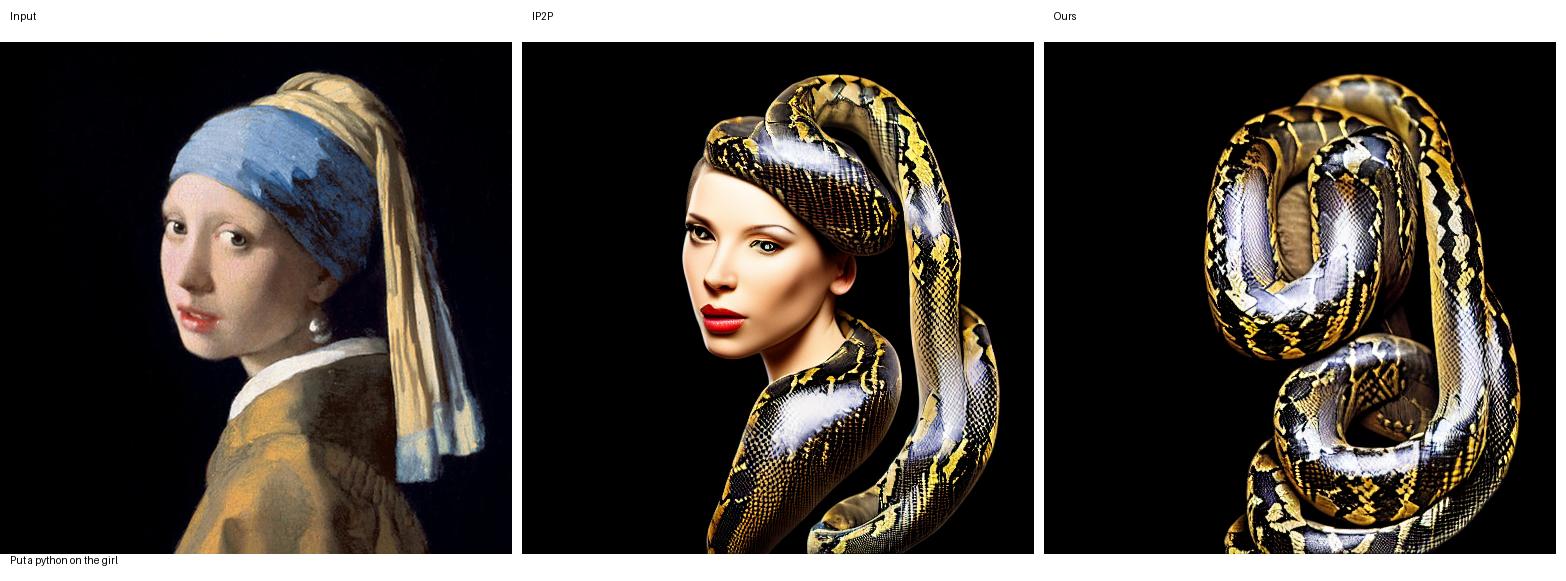}
        \parbox{\linewidth}{\vspace{-0.02in}
        ``Put a python on the girl''
        \vspace{0.05in}}
    \end{subfigure}
    \begin{subfigure}{0.40\linewidth}
        \includegraphics[width=\linewidth, trim=0 25px 0 25px, clip]{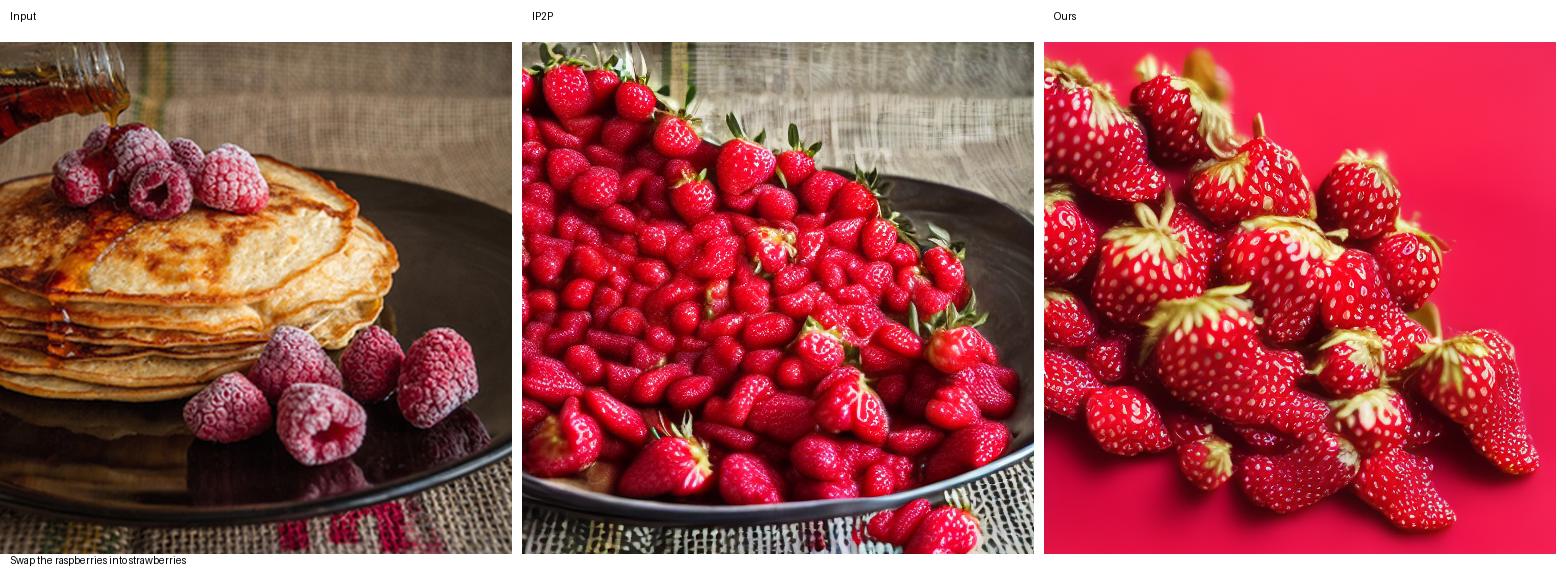}
        \parbox{\linewidth}{\vspace{-0.02in}
        ``Swap the raspberries into strawberries''
        \vspace{0.05in}}
    \end{subfigure}

    \begin{subfigure}{0.40\linewidth}
        \includegraphics[width=\linewidth, trim=0 25px 0 25px, clip]{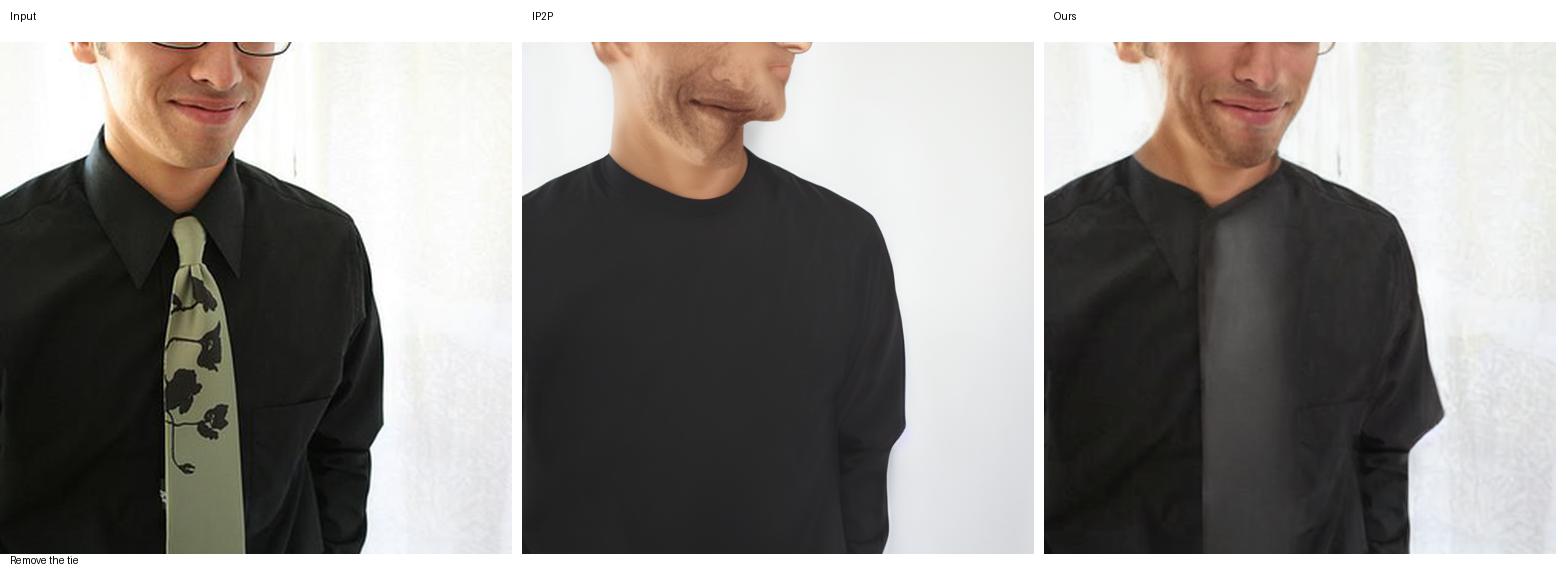}
        \parbox{\linewidth}{\vspace{-0.02in}
        ``Remove the tie''
        \vspace{0.05in}}
    \end{subfigure}
    \begin{subfigure}{0.40\linewidth}
        \includegraphics[width=\linewidth, trim=0 25px 0 25px, clip]{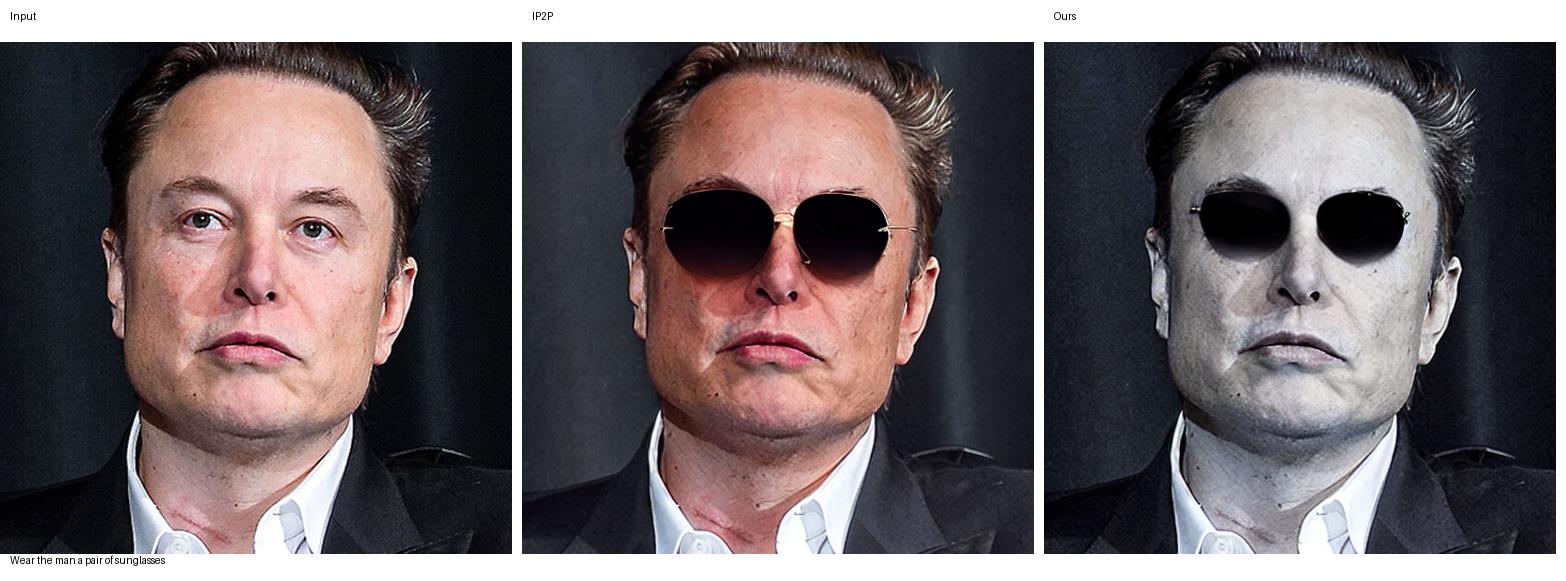}
        \parbox{\linewidth}{\vspace{-0.02in}
        ``Wear the man a pair of sunglasses''
        \vspace{0.05in}}
    \end{subfigure}

    \begin{subfigure}{0.40\linewidth}
        \includegraphics[width=\linewidth, trim=0 25px 0 25px, clip]{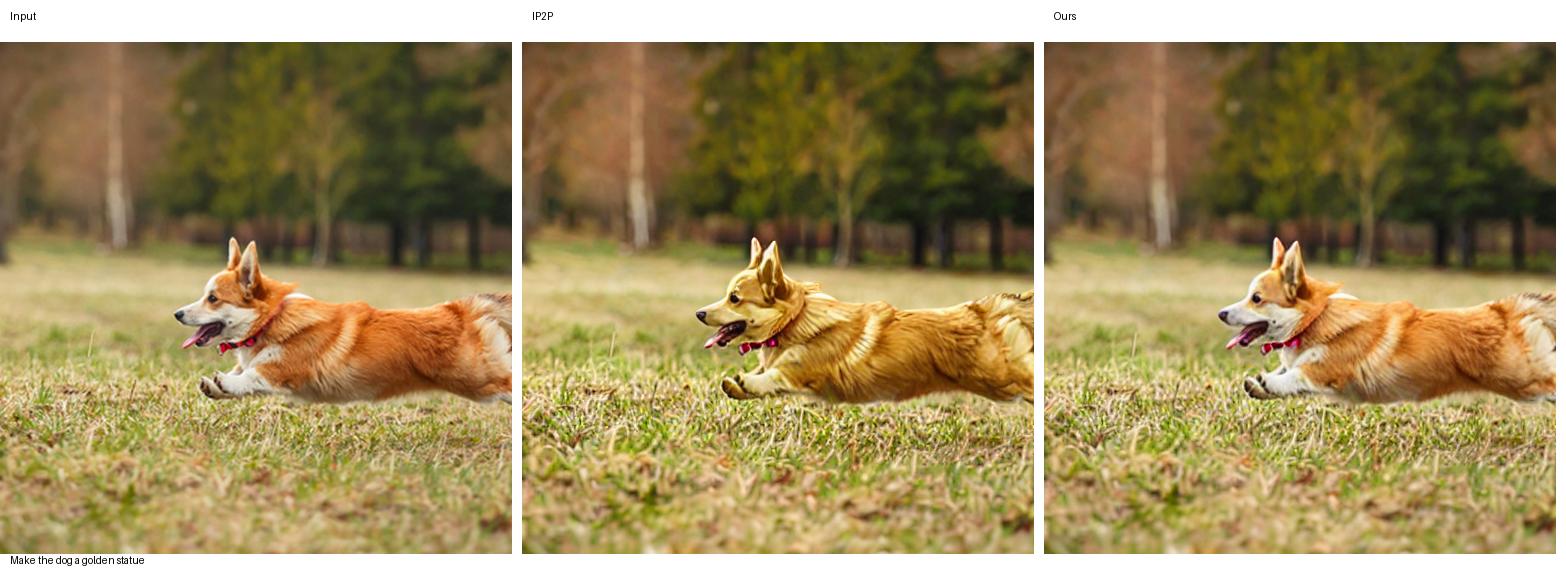}
        \parbox{\linewidth}{\vspace{-0.02in}
        ``Make the dog a golden statue''
        \vspace{0.05in}}
    \end{subfigure}
    \begin{subfigure}{0.40\linewidth}
        \includegraphics[width=\linewidth, trim=0 25px 0 25px, clip]{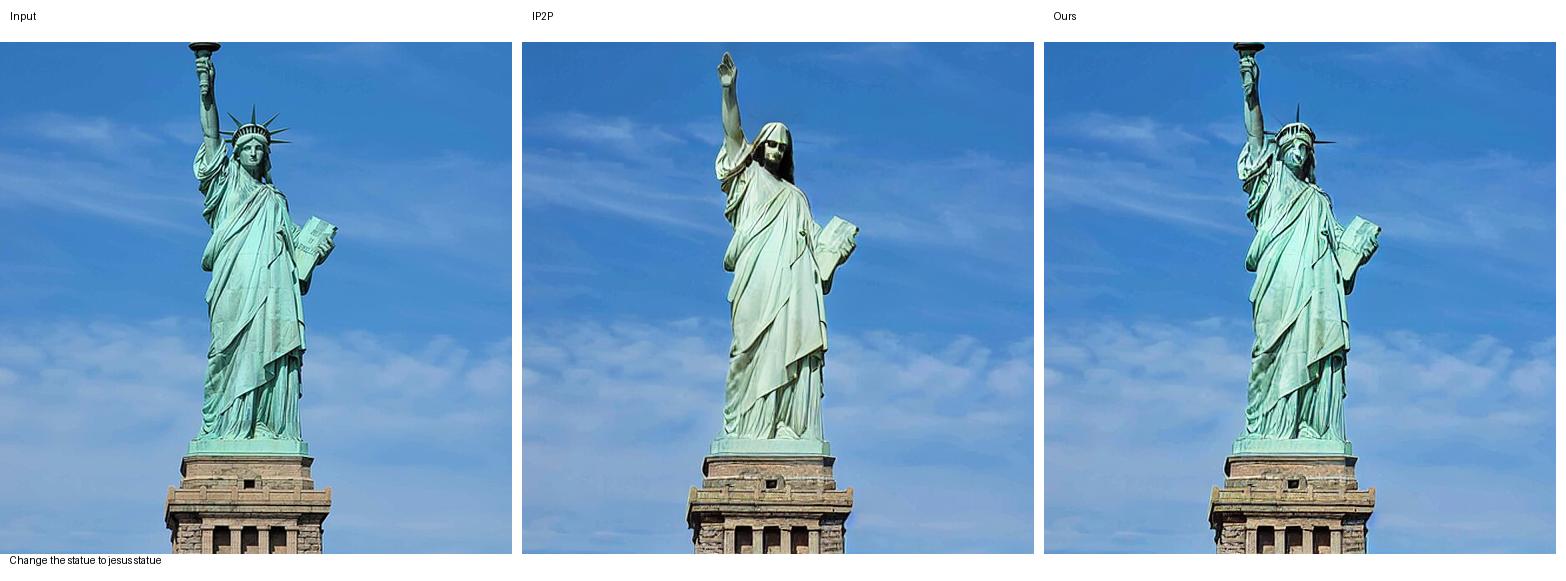}
        \parbox{\linewidth}{\vspace{-0.02in}
        ``Change the statue to jesus status''
        \vspace{0.05in}}
    \end{subfigure}

    \begin{subfigure}{0.40\linewidth}
        \includegraphics[width=\linewidth, trim=0 25px 0 25px, clip]{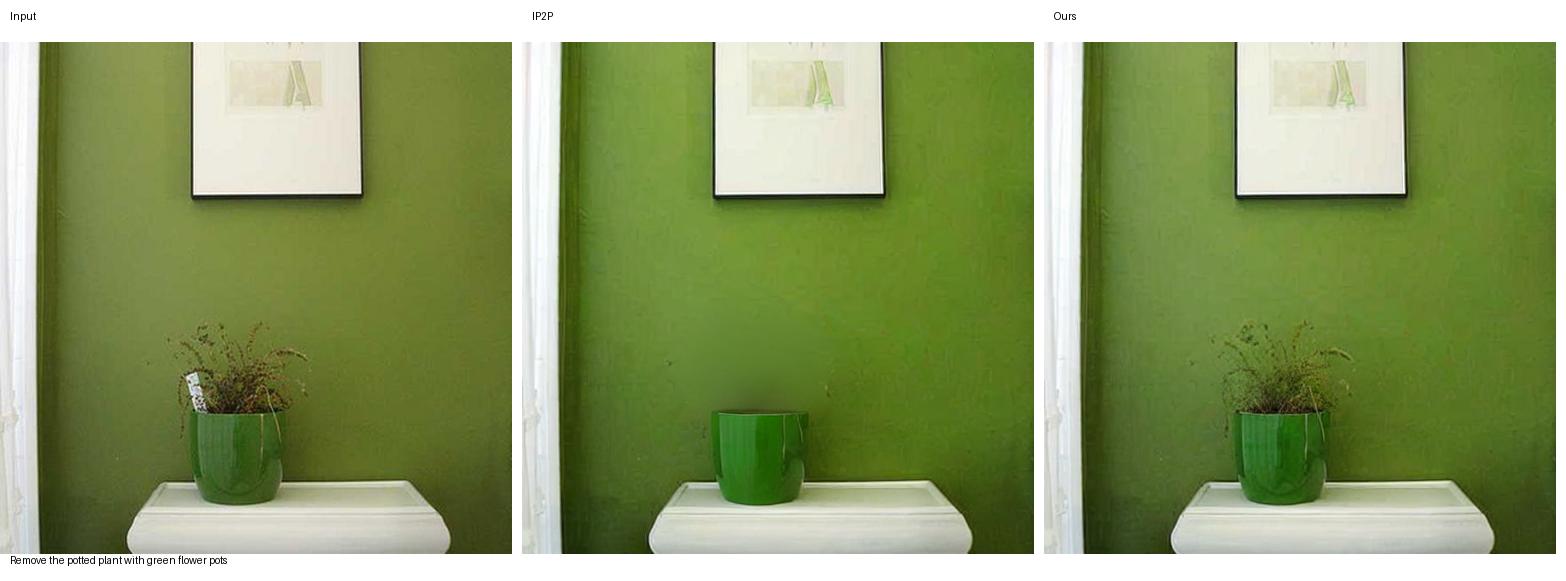}
        \parbox{\linewidth}{\vspace{-0.02in}
        ``Remove the potted plant with green flower pots''
        \vspace{0.05in}}
    \end{subfigure}
    \begin{subfigure}{0.40\linewidth}
        \includegraphics[width=\linewidth, trim=0 25px 0 25px, clip]{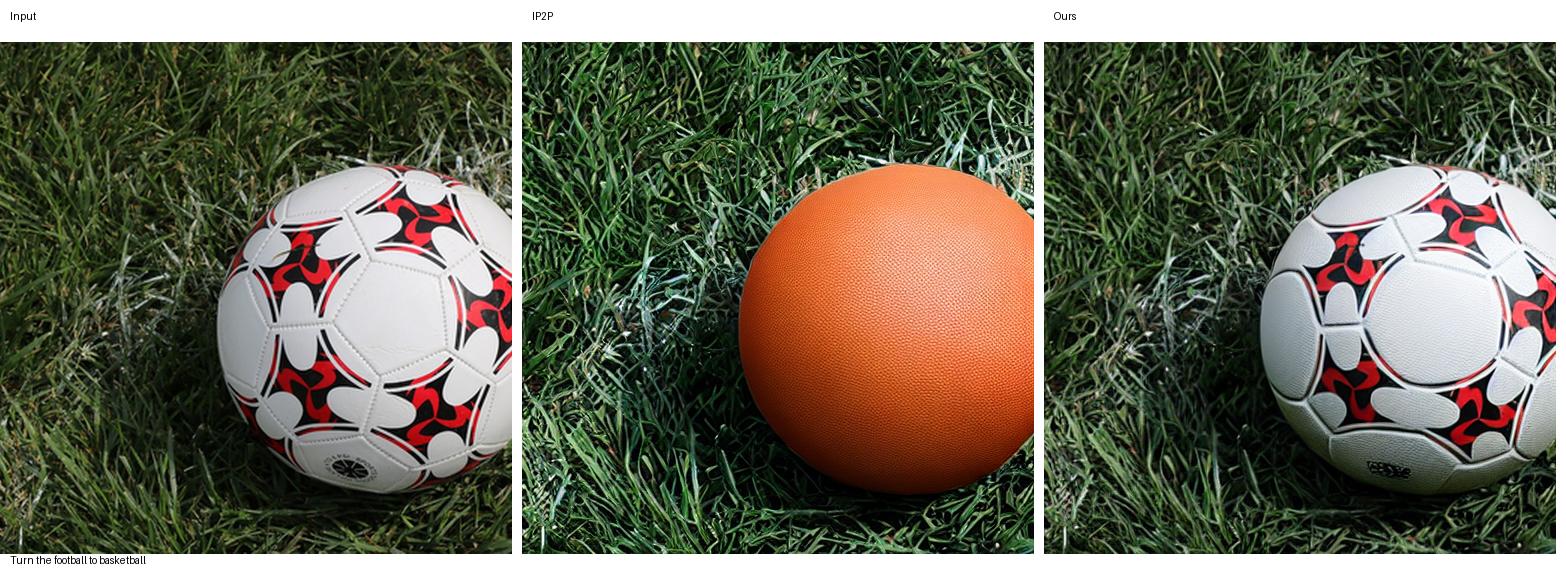}
        \parbox{\linewidth}{\vspace{-0.02in}
        ``Turn the football to basketball''
        \vspace{0.05in}}
    \end{subfigure}

    \begin{subfigure}{0.40\linewidth}
        \includegraphics[width=\linewidth, trim=0 25px 0 25px, clip]{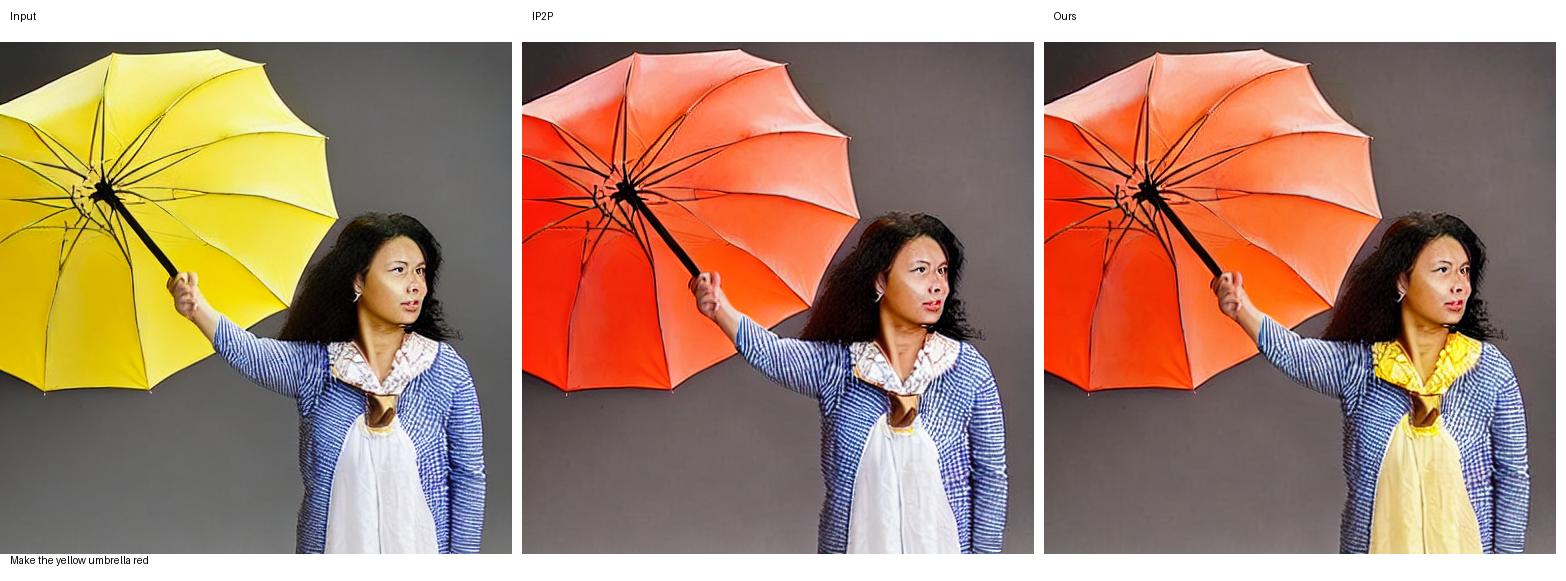}
        \parbox{\linewidth}{\vspace{-0.02in}
        ``Make the yellow umbrella red''
        \vspace{0.05in}}
    \end{subfigure}
    \begin{subfigure}{0.40\linewidth}
        \includegraphics[width=\linewidth, trim=0 25px 0 25px, clip]{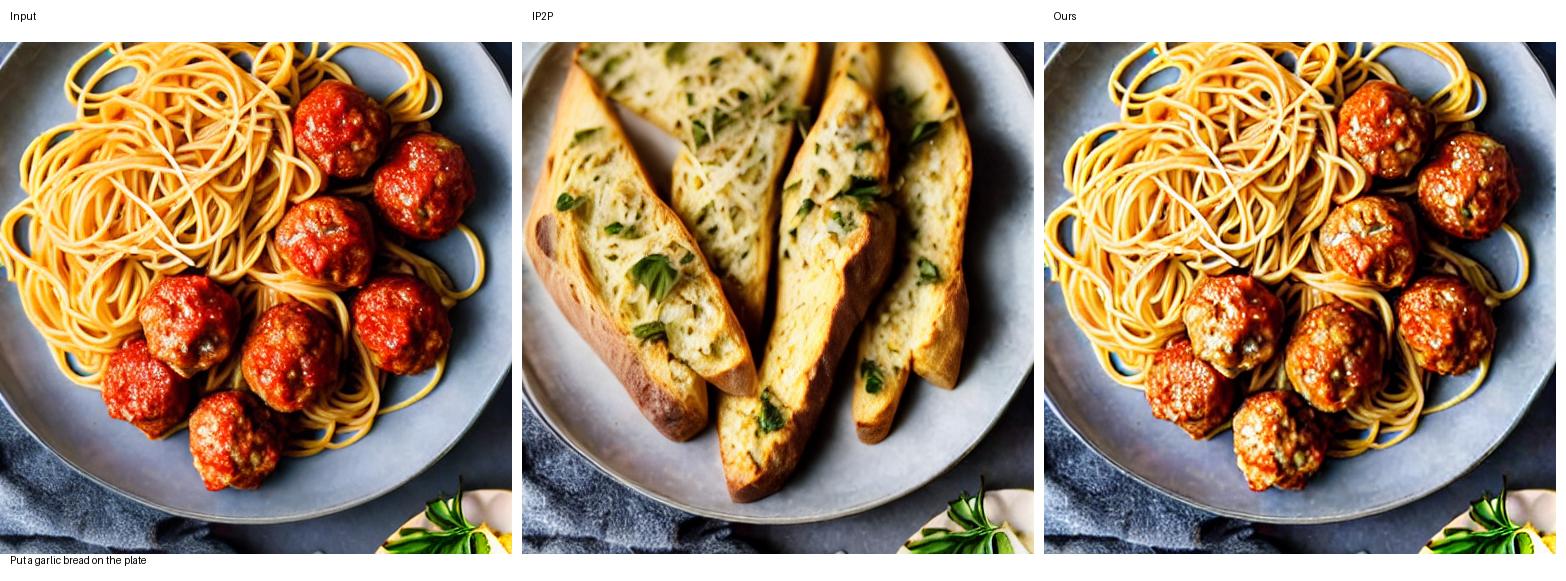}
        \parbox{\linewidth}{\vspace{-0.02in}
        ``Put a garlic bread on the plate''
        \vspace{0.05in}}
    \end{subfigure}

    \begin{subfigure}{0.40\linewidth}
        \includegraphics[width=\linewidth, trim=0 25px 0 25px, clip]{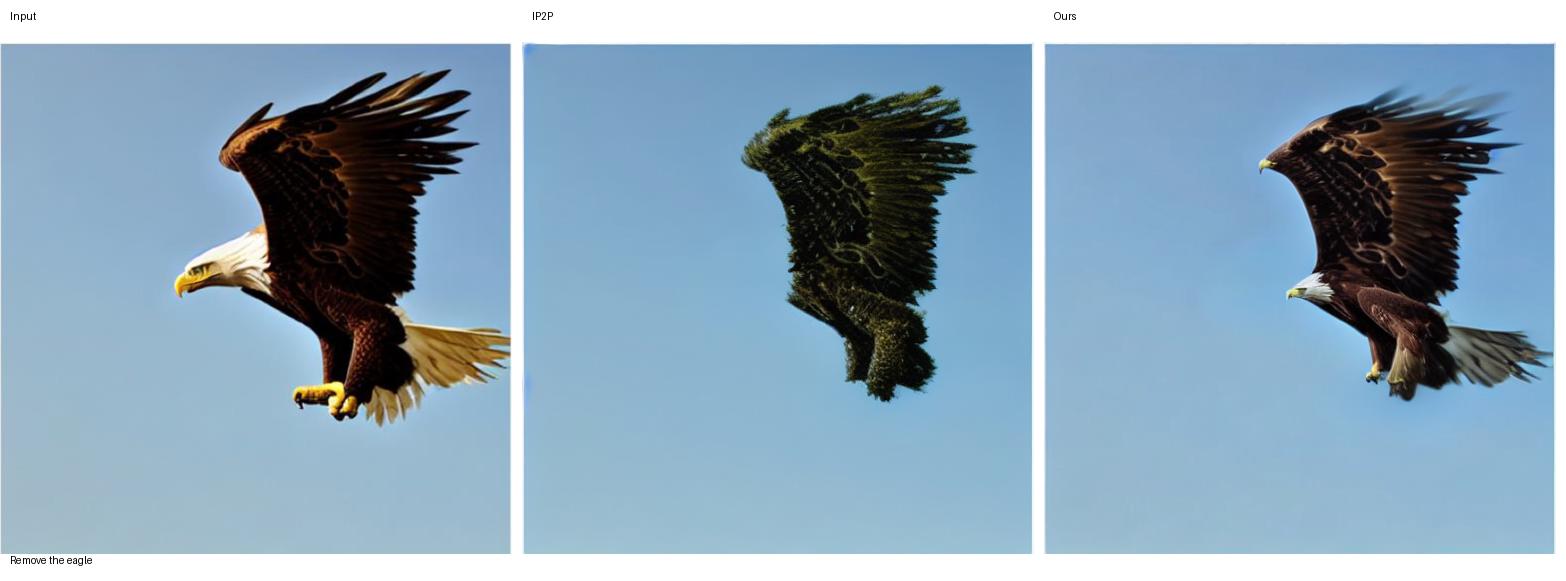}
        \parbox{\linewidth}{\vspace{-0.02in}
        ``Remove the eagel''
        \vspace{0.05in}}
    \end{subfigure}
    \begin{subfigure}{0.40\linewidth}
        \includegraphics[width=\linewidth, trim=0 25px 0 25px, clip]{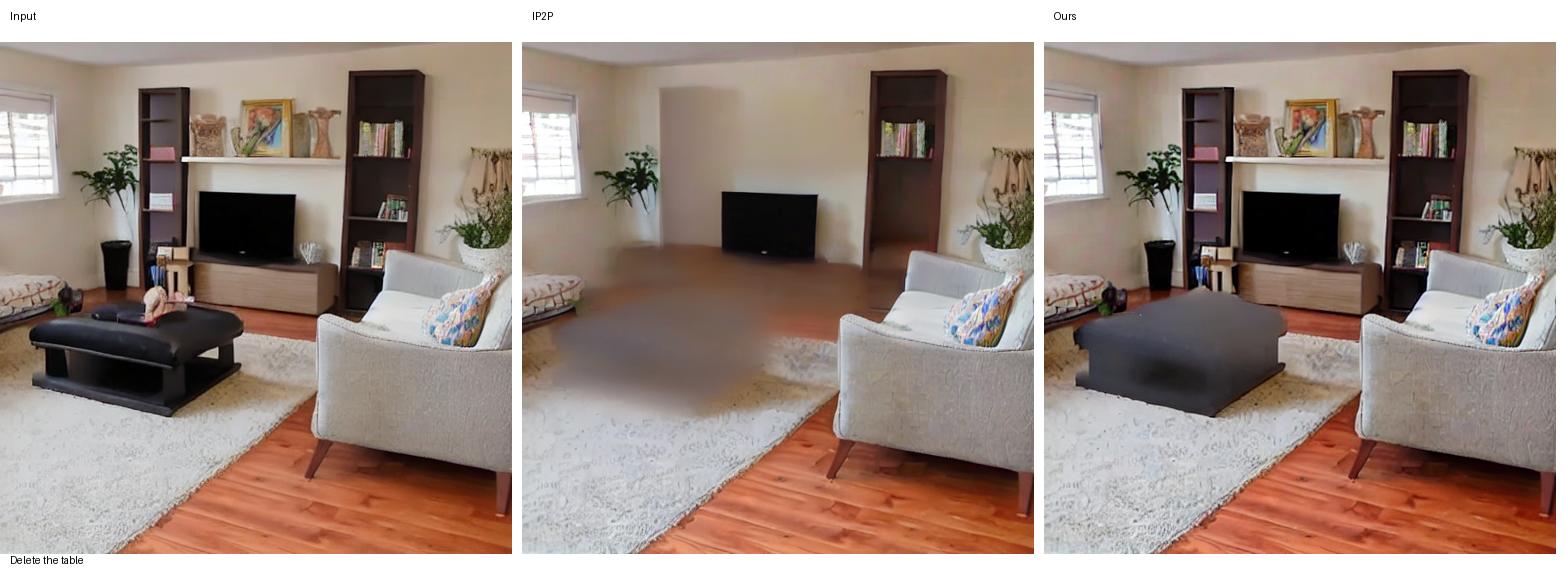}
        \parbox{\linewidth}{\vspace{-0.02in}
        ``Delete the table''
        \vspace{0.05in}}
    \end{subfigure}

    \begin{subfigure}{0.40\linewidth}
        \includegraphics[width=\linewidth, trim=0 25px 0 25px, clip]{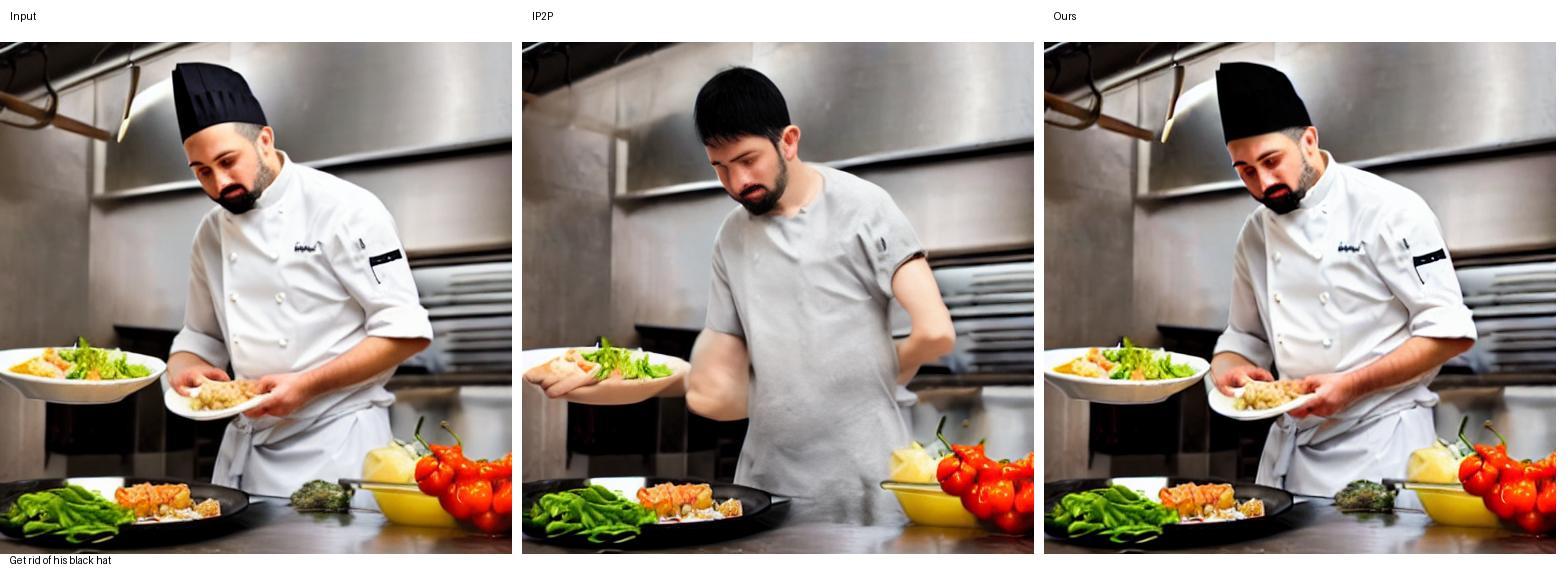}
        \parbox{\linewidth}{\vspace{-0.02in}
        ``Gid rid of his black hat''
        \vspace{0.05in}}
    \end{subfigure}
    \begin{subfigure}{0.40\linewidth}
        \includegraphics[width=\linewidth, trim=0 25px 0 25px, clip]{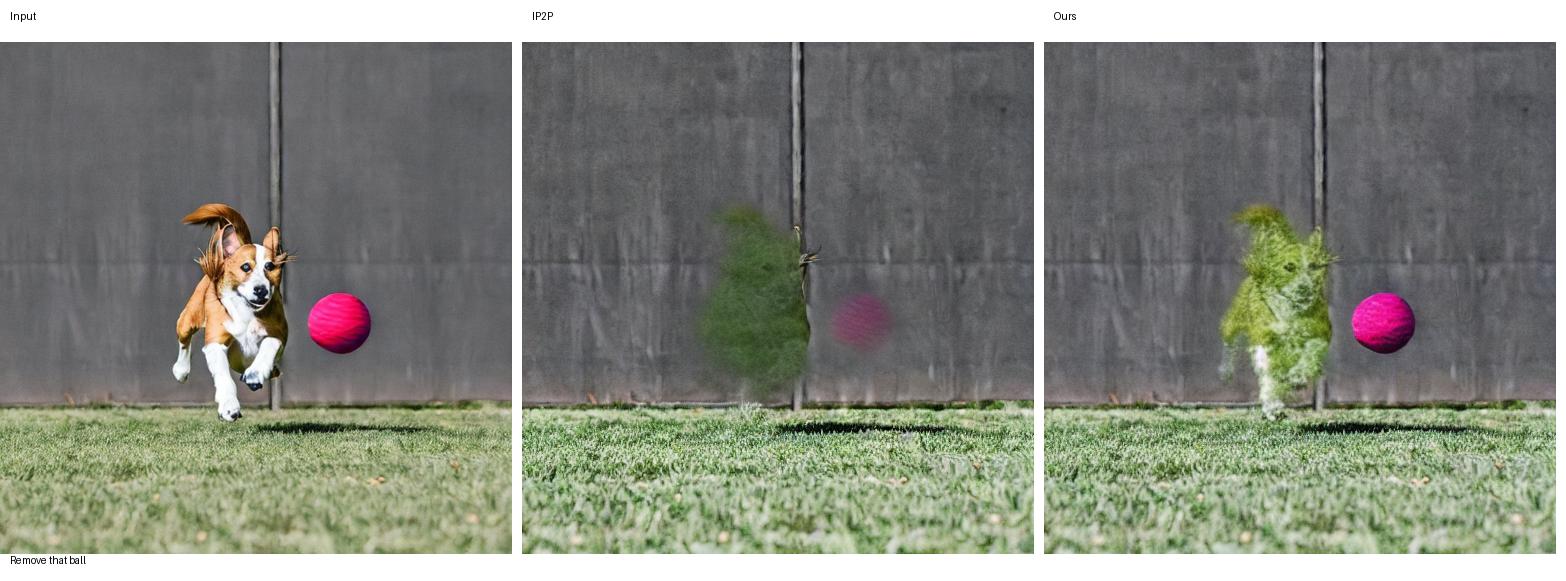}
        \parbox{\linewidth}{\vspace{-0.02in}
        ``Remove the ball''
        \vspace{0.05in}}
    \end{subfigure}


    \caption{Additional failure cases from our \ourwork{} image editing method on benchmarks~\cite{zhang2024magicbrush,zhang2023hive,zhao2024instructbrush,li2024zone})}
    \label{fig:additional_failure_cases}
\end{figure*}

We include the multi-turn edit example (Fig.~\ref{fig:multiturn}) mentioned in the paper. Additionally, we provide more instruction-guided image editing results in Figs.~\ref{fig:additional_editing_results_1} and~\ref{fig:additional_editing_results_2}, as well as samples from our dataset with refined instructions in Figs.~\ref{fig:additional_dataset_samples_1} and~\ref{fig:additional_dataset_samples_2}. Lastly, we present additional failure cases in Fig.~\ref{fig:additional_failure_cases}.

\end{document}